\documentclass{article}

\PassOptionsToPackage{numbers, compress}{natbib}
\usepackage[final]{neurips_2021}

\usepackage[utf8]{inputenc} % allow utf-8 input
\usepackage[T1]{fontenc}    % use 8-bit T1 fonts
\usepackage{hyperref}       % hyperlinks
\usepackage{url}            % simple URL typesetting
\usepackage{booktabs}       % professional-quality tables
\usepackage{amsfonts}       % blackboard math symbols
\usepackage{nicefrac}       % compact symbols for 1/2, etc.
\usepackage{microtype}      % microtypography
\usepackage{xcolor}         % colors
\usepackage{amsmath}
\usepackage{amssymb}
\usepackage{graphicx}
\usepackage{subcaption}
\usepackage{colortbl}
\usepackage{algorithm}
\usepackage{subcaption}
\usepackage{tabularx}
\usepackage{amsmath,amssymb,amsthm,bbm}
\usepackage{multirow}
\usepackage{adjustbox}
\usepackage{algorithm}
\usepackage{algpseudocode}
\usepackage{amsmath}
\usepackage[noEnd=false]{algpseudocodex}
\usepackage[toc,page]{appendix} % For appendix and TOC in the appendix
\usepackage{tocloft} % For customizing TOC
\usepackage{titlesec}
\usepackage{booktabs}
\usepackage{threeparttable}
\usepackage{enumitem}
\setlist[enumerate]{left=10pt}
\usepackage[font=small]{caption}
\usepackage{wrapfig}
\usepackage{listings}
\usepackage{booktabs}
\usepackage{mdframed}

\newtheorem{proposition}{Proposition}
\newtheorem{lemma}{Lemma}

\definecolor{lightgreen}{rgb}{0.88,1,0.88} % Define a custom color for light green

\newcounter{todocounter}
\title{Gumbel-Softmax Flow Matching \\with Straight-Through Guidance for Controllable\\Biological Sequence Generation}

\author{
  Sophia Tang$^{1, 2}$, Yinuo Zhang$^{1, 3}$, Alexander Tong$^{4,5}$,  
  \textbf{Pranam Chatterjee}$^{1, 6, 7, \dag}$ \\\\
  $^{1}$Department of Biomedical Engineering, Duke University \\
  $^{2}$Management and Technology Program, University of Pennsylvania \\
  $^{3}$Center of Computational Biology, Duke-NUS Medical School \\
  $^{4}$Mila, Quebec AI Institute, $^{5}$Université de Montréal \\
  $^{6}$Department of Computer Science, Duke University \\
  $^{7}$Department of Biostatistics and Bioinformatics, Duke University
  \\\\
  $^{\dag}$Corresponding author: 
  \href{mailto:pranam.chatterjee@duke.edu}{pranam.chatterjee@duke.edu}
}

\begin{document}

\maketitle

\begin{abstract}
\looseness=-1
Flow matching in the continuous simplex has emerged as a promising strategy for DNA sequence design, but struggles to scale to higher simplex dimensions required for peptide and protein generation. We introduce \textbf{Gumbel-Softmax Flow and Score Matching}, a generative framework on the simplex based on a novel Gumbel-Softmax interpolant with a time-dependent temperature. Using this interpolant, we introduce Gumbel-Softmax Flow Matching by deriving a parameterized velocity field that transports from smooth categorical distributions to distributions concentrated at a single vertex of the simplex. We alternatively present Gumbel-Softmax Score Matching which learns to regress the gradient of the probability density. Our framework enables high-quality, diverse generation and scales efficiently to higher-dimensional simplices. To enable training-free guidance, we propose \textbf{Straight-Through Guided Flows (STGFlow)}, a classifier-based guidance method that leverages straight-through estimators to steer the unconditional velocity field toward optimal vertices of the simplex. STGFlow enables efficient inference-time guidance using classifiers pre-trained on clean sequences, and can be used with any discrete flow method. Together, these components form a robust framework for controllable \textit{de novo} sequence generation. We demonstrate state-of-the-art performance in conditional DNA promoter design, sequence-only protein generation, and target-binding peptide design for rare disease treatment.
\end{abstract}

%%%%%%%%%%%%%%%%%%%%%%%%%%%%%%%%%%%%%%%%%%%%%%%%%%%%%%%%%%%%
%%%%%%%%%%%%%%%%%%%%%%%%%%%%%%%%%%%%%%%%%%%%%%%%%%%%%%%%%%%%
\section{Introduction}
Generative modeling has transformed the design of biological sequences, enabling \textit{de novo} protein design \citep{Madani2023, Ferruz2022, Nisinoff2024}, DNA regulatory elements \citep{Stark2024, Nisinoff2024}, and peptides \citep{Bhat2025, Chen2024pepmlm, Tang2024}. However, generating structured sequences in discrete spaces remains an open challenge due to the inherent non-differentiability of categorical variables. Traditional autoregressive models, such as ProtGPT2 \citep{Ferruz2022} and ProGen2 \citep{Madani2023}, learn sequence distributions by iteratively predicting tokens, but suffer from compounding errors, bias accumulation, and limited global coherence. To address these issues, generative models based on diffusion \citep{Austin2021, Wang2024, Shi2024, Sahoo2024} and flow matching \citep{Gat2024, Stark2024, Nisinoff2024, Davis2024} have been developed to enable non-autoregressive sampling of sequences. 

Discrete diffusion \citep{Shi2024, Sahoo2024, Austin2021} and flow-matching \citep{Gat2024, Nisinoff2024} models, iteratively reconstruct sequences by modeling forward and reverse noise processes in a Markovian framework. These approaches have demonstrated success in DNA sequence design \citep{Stark2024, Nisinoff2024}, protein generation \citep{Wang2024, Goel2024}, and recently, multi-objective generation of therapeutic peptides \citep{Tang2024}. However, these methods operate in the fully \textit{discrete} state space, which means that the noisy sequence at each time step is a fully discrete sequence of one-hot vectors sampled from continuous categorical distributions. This can result in discretization errors during sampling when abruptly restricting continuous distributions to a single token. This presents the question: \textit{Can we generate discrete sequences by iteratively fine-tuning continuous probability distributions?} This is the motivation behind discrete flow matching models on the simplex \cite{Stark2024, Davis2024}, which defines a smooth interpolation from a uniform prior over the simplex to a unitary distribution concentrated at a single vertex.

Despite these advances, previous discrete simplex-based flow-matching methods have yet to be applied to \textit{de novo} design tasks like protein and target-specific peptide design that require learning diverse flow trajectories that scale to higher simplex dimensions. Furthermore, there remains a lack of \textit{controllability} at inference time due to strictly deterministic paths and the absence of modular training-free guidance methods. To address these gaps, we introduce \textbf{Gumbel-Softmax Flow Matching} (Gumbel-Softmax FM), a generative framework that transforms noisy to clean data on the interior of the simplex by defining a novel Gumbel-Softmax interpolant with a time-dependent temperature parameter. By applying Gumbel noise during training, Gumbel-Softmax FM avoids overfitting to the training data, increasing the exploration of diverse flow trajectories. We also introduce STGFlow, a training-free classifier-based guidance strategy that enables training-free classifier-based guidance for target-binding peptide generation. 

Our key contributions are as follows:
\begin{enumerate}
    \item \textbf{Gumbel-Softmax Flow Matching}. We introduce Gumbel-Softmax FM, a generative framework that leverages temperature-controlled Gumbel-softmax interpolants for smooth transport from noisy to clean distributions on the simplex. We define a new velocity field that follows a mixture of learned interpolations between categorical distributions that converge to high-quality sequences (Section \ref{section:Gumbel-Softmax Flow Matching}). 
    \item \textbf{Gumbel-Softmax Score Matching}. As an alternative generative framework using the same Gumbel-softmax interpolant, we propose Gumbel-Softmax SM that estimates the gradient of probability density at varying temperatures to enable sampling from high-density regions on the simplex (Section \ref{section:Gumbel-Softmax Score Matching}). 
    \item \textbf{Straight-Through Guided Flow Matching (STGFlow).} Given the lack of post-training guidance methods for discrete flow matching, we introduce Straight-Through Guided Flow Matching, a novel training-free classifier-based guidance algorithm that leverages straight-through gradients to guide the flow trajectory towards high-scoring sequences (Section \ref{section:Straight-Through Guided Flows}). We apply this method to generate high-affinity peptide binders to target proteins (Section \ref{section:Peptide Binder Design}). 
    \item \textbf{Biological Sequence Generation}. We apply our framework to conditional DNA promoter design, \textit{de novo} protein sequence generation, and target-binding peptide design, demonstrating competitive performance compared to autoregressive and discrete diffusion-based baselines (Section \ref{section:Experiments}).
\end{enumerate}

Our framework offers several theoretical and empirical advantages over autoregressive and discrete diffusion models, and we believe it will serve as a foundation for controllable flow matching for discrete sequence generation. 

\section{Preliminaries}
We consider a noisy uniform distribution over the $(V-1)$-dimensional simplex $p_0(\mathbf{x}_0)$ and a clean distribution $p_1(\mathbf{x}_1)$ over discrete samples $\mathbf{x}_1\sim \mathcal{D}$ from a dataset $\mathcal{D}$. The challenge of generative modeling over the simplex consists of defining a time-dependent flow $\psi_t$ that smoothly interpolates between $p_0$ and $p_1$. Then, we can generate samples from $p_1$ by first sampling from $p_0$ the applying a learned velocity field that transports distributions from $p_0$ to $p_1$. 

\subsection{The Gumbel-Softmax Distribution}
The Gumbel-Softmax distribution or Concrete distribution \citep{Jang2016, Maddison2016} is a relaxation of discrete random variables onto the interior of the simplex $\Delta^{V-1}=\{\mathbf{x}\in \mathbb{R}^V |x_i \in [0, 1], \sum_{j=1}^Vx_j=1\}$. This continuous relaxation is achieved by adding i.i.d. sampled Gumbel noise $g_i= -\log(-\log \mathcal{U}_i))$, where $\mathcal{U}_i\sim \text{Uniform}(0, 1)$, scaling down by the temperature parameter $\tau>0$, and applying the differentiable \texttt{softmax} function across the distribution such that the elements sum to 1. Given parameters $\pi_i\in (\epsilon, \infty)$ representing the original logits of each category where $\epsilon$ is a small constant to avoid undefined logarithms, the Gumbel-Softmax random variable is given by
\begin{align}
    x_i=\text{SM}\left(\frac{\log \pi_i+g_i}{\tau}\right)=\frac{\exp\left(\frac{\log \pi_i+g_i}{\tau}\right)}{\sum_{j=1}^V \exp\left(\frac{\log \pi_j+g_j}{\tau}\right)}
\end{align}
where $\text{SM}(\cdot)$ denotes the \texttt{softmax} function. We observe that as $\tau\to 0$, the distribution converges to a one-hot vector where $x_k\to 1$ and $x_j\to 0$ for $j\neq k$ given that $k=\arg \max _k\left(\log \pi_k+g_k\right)$. Conversely, as $\tau\to \infty$, the distribution approaches a uniform distribution where $x_j\to \frac{1}{V}$ for all $j\in [1, V]$. 

\subsection{Discrete Flow Matching} 
Flow matching~\cite{peluchetti2022nondenoising,liu_rectified_2022,albergo_stochastic_2023} is a \textit{simulation-free} generative framework that aims to transform noisy samples $\mathbf{x}_0\sim p_0$ from a source distribution $p_0$ to clean samples $\mathbf{x}_1\sim p_1$ from the data distribution $p_1$ by learning to predict the marginal velocity field $u_t(\mathbf{x}_t)$ that transports $p_0$ to $p_1$ as a mixture of conditional velocity fields $u_t^{\theta}(\mathbf{x}_t|\mathbf{x}_1)$ parameterized by a neural network. The \textit{interpolant} $\psi_t(\mathbf{x}_1):[0,1]\times \Delta^{V-1}\times \Delta ^{V-1}\to \Delta ^{V-1}$ is a function that defines the flow from a clean distribution $\mathbf{x}_1$ on a vertex of the simplex to the intermediate distribution $\mathbf{x}_t$ at time $t$, which satisfies the constraints $\psi_0(\mathbf{x}_0|\mathbf{x}_1)=\mathbf{x}_0$ and $\psi_1(\mathbf{x}_0|\mathbf{x}_1)=\mathbf{x}_1\sim p_t$. Therefore, the conditional velocity field is given by the time-derivative of $\psi_t(\mathbf{x}_1)$.
\begin{align}
    u_t(\mathbf{x}_t|\mathbf{x}_1)=\frac{d}{dt}\psi_t(\mathbf{x}_1)
\end{align}
where $u_t\in \mathcal{T}_{\mathbf{x}_t}\Delta^{V-1}$ and $\mathcal{T}_{\mathbf{x}_t}\Delta^V$ is the set of tangent vectors to the manifold at point $\mathbf{x}_t$. For a velocity field $u_t$ to \textit{generate} $p_t$, it must satisfy the \textit{continuity equation} given by
\begin{align}
    \frac{\partial}{\partial t}p_t(\mathbf{x}_t)=-\nabla\cdot(p_t(\mathbf{x}_t)u_t(\mathbf{x}_t))\label{eq:continuity equation}
\end{align}
where $\nabla\cdot$ is the divergence operator that describes the total outgoing flux at a point $\mathbf{x}_t$ along the flow trajectory. The flow matching (FM) objective is to train a parameterized model $u_t^{\theta}(\mathbf{x}_t)$ to approximate $u_t$ given a noisy sample $\mathbf{x}_t$ at time $t\in [0,1]$ by minimizing the squared norm
\begin{align}
    \mathcal{L}_{\text{FM}}=\mathbb{E}_{t, \mathbf{x}_t}\left \|u_t^{\theta}(\mathbf{x}_t)-u_t(\mathbf{x}_t)\right \|^2
\end{align}
But since computing $u_t(\mathbf{x}_t)$ requires marginalizing over all possible trajectories and is intractable, we condition the velocity field on each data point $\mathbf{x}_1$ and compute the conditional flow-matching (CFM) objective given by
\begin{align}
    \mathcal{L}_{\text{CFM}}=\mathbb{E}_{t, \mathbf{x}_t}\left \|u_t^{\theta}(\mathbf{x}_t)-u_t(\mathbf{x}_t|\mathbf{x}_1)\right\|^2\label{eq:CFM Loss}
\end{align}
which is tractable and has the same gradient as the unconditional flow-matching loss $\nabla_{\theta}\mathcal{L}_{\text{FM}}=\nabla_{\theta}\mathcal{L}_{\text{CFM}}$ \citep{Lipman2022, Optimal-Transport}. Among existing discrete flow matching methods, there are two methods of defining a discrete flow: defining the \textit{interpolant} $\psi_t(\mathbf{x}_1)$ that connects a noisy sample $\mathbf{x}_0$ to a clean one-hot sample $\mathbf{x}_1$ and defining the \textit{probability path} which pushes density from the prior distribution $p_0$ to the target data distribution $p_1$. In this work, we define a new temperature-dependent interpolant and derive the corresponding velocity field. 

\subsection{Score Matching Generative Models}
Score matching \citep{Yang2020} is another generative matching framework that learns the gradient of the conditional probability density path $\nabla_{\mathbf{x}_t}\log p_{t}(\mathbf{x}_t)$ (defined as the \textit{score}) of the interpolation between noisy and clean data. By parameterizing the score function with $s_{\theta}(\mathbf{x}_t, t)$, we can minimize the score matching loss given by
\begin{align}
    \mathcal{L}_{\text{score}}=\mathbb{E}_{p_t(\mathbf{x}_t)}\left \|\nabla_{\mathbf{x}_t}\log p_t(\mathbf{x}_t)-s_{\theta}(\mathbf{x}_t, t)\right \|^2
\end{align}
Similarly to flow-matching, directly learning $\nabla_{\mathbf{x}_t}\log p_t(\mathbf{x}_t)$ is intractable, so we learn the conditional probability path $\nabla_{\mathbf{x}_t}\log p_t(\mathbf{x}_t|\mathbf{x}_1)$ conditioned on $\mathbf{x}_1\sim p_1(\mathbf{x}_1)$ by minimizing 
\begin{align}
    \mathcal{L}_{\text{score}}=\mathbb{E}_{p_t(\mathbf{x}_t|\mathbf{x}_1), p_1(\mathbf{x}_1)}\left \|\nabla_{\mathbf{x}_t}\log p_t(\mathbf{x}_t|\mathbf{x}_1)-s_{\theta}(\mathbf{x}_t, t)\right \|^2
\end{align}
which we show in Appendix \ref{appendix:Conditional Score Function} equals the unconditional score function by expectation over $\mathbf{x}_1$. 

\begin{figure*}
    \centering
    \includegraphics[width=\linewidth]{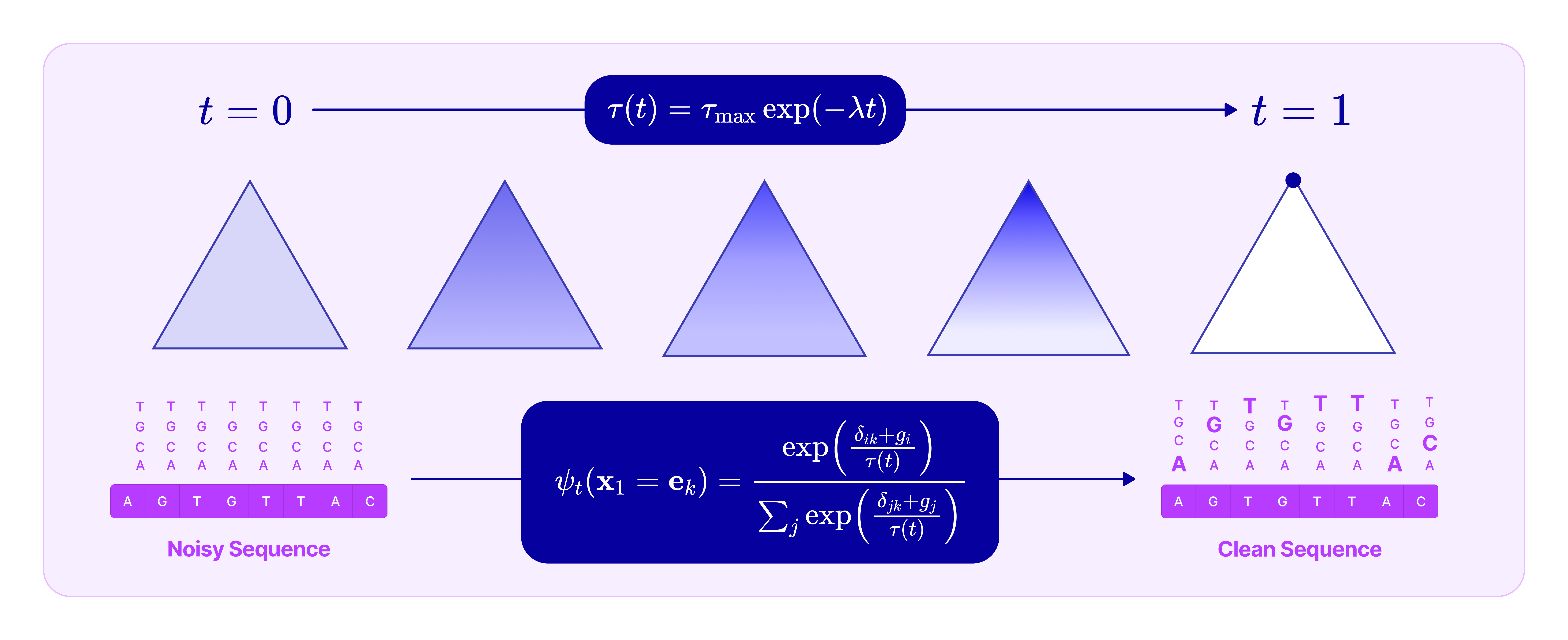}
    \caption{\textbf{Overview of Gumbel-Softmax Flow Matching.} Gumbel-softmax transformations are applied to clean one-hot sequences for varying temperatures dependent on time. The embedded noisy distributions are passed into a parameterized flow or score model and error prediction model to predict the conditional flow velocity and score function. }
    \label{fig:enter-label}
\end{figure*}

\section{Gumbel-Softmax Flow Matching}\label{section:Gumbel-Softmax Flow Matching}
In this work, we present \textbf{Gumbel-Softmax Flow Matching (FM)}, a novel simplex-based flow matching method that defines the noisy logits at each time step with the Gumbel-Softmax transformation, enabling smooth interpolation between noisy and clean data by modulating the temperature $\tau(t)$, which changes as a function of time. 

\subsection{Defining the Gumbel-Softmax Interpolant}
We propose a new definition of the discrete probability path by gradually decreasing the temperature of a Gumbel-Softmax categorical distribution as a function of time where the maximum probability corresponds to the target token. First, we define a monotonically decreasing function $\tau(t)\in (0, \infty)$ to prevent the Gumbel-Softmax distribution from being undefined at $\tau =0$.
\begin{align}
    \tau(t)=\tau_{\text{max}} \exp(-\lambda t)
\end{align}
where $\tau_{\max}$ is the initial temperature set to a large number so that the categorical distribution resembles a uniform distribution, $\lambda$ controls the decay rate, and $t$ is the time that goes from $t=0$ to $t=1$. 

Now, we define the conditional interpolant $\mathbf{x}_t=\psi_t(\mathbf{x}_1=\mathbf{e}_k)$ with $t\in [0, 1]$ and Gumbel-noise scaled by a factor $\beta$ as
\begin{align}
    \psi_t(\mathbf{x}_1=\mathbf{e}_k)=\frac{\exp\left(\frac{ \delta_{ik} +(g_i /\beta)}{\tau(t)}\right)}{\sum _{j=1}^V\exp\left(\frac{\delta_{jk}+(g_j /\beta)}{\tau(t)}\right)}\label{eq:gumbel softmax interpolant}
\end{align}
where $\tau(t)=\tau_{\text{max}} \exp(-\lambda t)$ and $\pi_i =\exp(\delta_{ik})$. $\delta_{ik}$ is the Kronecker delta function that returns 1 when $i=k$ and 0 otherwise. This decaying time-dependent temperature function $\tau (t)$ ensures that the distribution becomes more concentrated at the target vertex as $t\to 1$. Gumbel noise is applied during training to ensure that the model learns to reconstruct a clean sequence given contextual information. 

\begin{proposition}[Continuity]
The proposed conditional vector field and conditional probability path together satisfy the continuity equation (Equation \ref{eq:continuity equation}) and thus define a valid flow matching trajectory on the interior of the simplex.
\end{proposition}

We provide the proof of continuity in Appendix \ref{appendix:Continuity Equation}. This definition of the flow satisfies the boundary conditions. For $t= 0$, $\tau(t)=\tau_{\text{max}}$ which produces a near-uniform distribution $\psi_0(\mathbf{x}_0|\mathbf{x}_1)\approx\frac{\mathbf{1}}{V}$. For $t=1$, $\exp (-\lambda t )\to 0$ (faster decay for larger $\lambda$) and $\tau(t)\to 0$, meaning the flow trajectory converges to the vertex of the simplex corresponding to the one-hot vector $\psi_1(\mathbf{x}_0|\mathbf{x}_1)\approx\mathbf{x}_1$. 

\subsection{Reparameterizing the Velocity Field}
From our definition of the Gumbel-Softmax interpolant, we derive the conditional velocity field $u_t(\mathbf{x}_0|\mathbf{x}_1)$ by taking the derivative of the flow (Appendix \ref{appendix:Deriving the Conditional Velocity Field}).
\begin{align}
    u_{t, i}(\mathbf{x}|\mathbf{x}_1=\mathbf{e}_k)=\frac{ \lambda}{\tau(t)} x_{t, i}\sum_{j=1}^Vx_{t, j}\cdot \bigg((\delta_{ik}+g_i)-(\delta_{jk}+g_j)\bigg)
\end{align}
\begin{proposition} [Probability Mass Conservation]
The conditional velocity field preserves the probability mass and lies in the tangent bundle at point $\mathbf{x}_t$ on the simplex $\mathcal{T}_{\mathbf{x}_t}\Delta^{V-1}=\{u_t\in \mathbb{R}^V|\langle\mathbf{1}, u_t\rangle=0\}$. 
\end{proposition}

Proof in Appendix \ref{appendix:Probability Mass Conservation}. Instead of directly regressing $u_t(\mathbf{x}_t|\mathbf{x}_1)$ by minimizing $\mathcal{L}_{\text{CFM}}$ defined in Equation \ref{eq:CFM Loss}, we train a \textit{denoising} model that predicts the probability vector  $\mathbf{x}_{\theta}(\mathbf{x}_t, t)\in \Delta ^{V-1}$ given the noisy interpolant $\mathbf{x}_t$ by minimizing the negative log loss. 
\begin{align}
    \mathcal{L}_{\text{gumbel}}=\mathbb{E}_{p_t (\mathbf{x}_t|\mathbf{x}_1=\mathbf{e}_k), p_1(\mathbf{x}_1)}\left[-\log \langle \mathbf{x}_{\theta}(\mathbf{x}_t, t), \mathbf{x}_1 \rangle \right]
\end{align}

During inference, we compute the predicted marginal velocity field as the weighted sum of the conditional velocity fields scaled by the predicted token probabilities. 
\begin{align}
    u_t^{\theta}(\mathbf{x}_t)=\sum_{k=1}^Vu_{t}(\mathbf{x}|\mathbf{x}_1=\mathbf{e}_k) \langle \mathbf{x}_{\theta}(\mathbf{x}_t, t), \mathbf{e}_k \rangle\label{eq:Linear Combination Velocity Field}
\end{align}

\begin{proposition}[Valid Flow Matching Loss]
If $p_t(\mathbf{x}_t)>0$ for all $\mathbf{x}_t\in \mathbb{R}^d$ and $t\in [0, 1]$, then the gradients of the flow matching loss and the Gumbel-Softmax FM loss are equal up to a constant not dependent on $\theta$ such that $\nabla_{\theta}\mathcal{L}_{\text{FM}}=\nabla_{\theta}\mathcal{L}_{\text{gumbel}}$ 
\end{proposition}
 
Proof in Appendix \ref{appendix:Valid Flow Matching Loss}. By our definition of the Gumbel-Softmax interpolant, the intermediate distributions during inference represent a mixture of learned conditional interpolants $\psi _t(\mathbf{x}_1)$ from the training data. Since the denoising model is trained to predict the true clean distribution, we can set the Gumbel-noise random variable in the conditional velocity fields to 0 during inference as we want the velocity field to point toward the predicted denoised distribution. Therefore, the conditional velocity field becomes
\begin{align}
    u_{t}(\mathbf{x}_t|\mathbf{x}_1=\mathbf{e}_k)&= \frac{\lambda}{\tau(t)}x_{t, k}\left(\mathbf{e}_k-\mathbf{x}_t\right)
\end{align}
which points toward the target vertex $\mathbf{e}_k$ at a magnitude proportional to $x_{t, k}(1-x_{t, k})$ and away from all other vertices at a magnitude proportional to $-x_{t, i}x_{t, k}$. We observe that the velocity field vanishes both at the vertex and the $(V-2)$-dimensional face directly opposite to the vertex and increases as $t\to 1$ and $\tau(t)\to 0$, accelerating towards the target vertex at later time steps. 

\section{Gumbel-Softmax Score Matching}\label{section:Gumbel-Softmax Score Matching}
As an alternative to our flow matching framework, we propose \textbf{Gumbel-Softmax Score Matching} (Gumbel-Softmax SM), a score-matching method that learns the gradient of the probability density path $\nabla_{\mathbf{x}_t}\log p_{t}(\mathbf{x}_t)$ associated with the Gumbel-Softmax interpolant. 

\subsection{The Exponential Concrete Distribution}
When computing Gumbel-Softmax random variables, the exponentiation of small values associated with low-probability tokens can result in numerical underflow. Since the logarithm of 0 is undefined, this could result in numerical instabilities when computing the log probability density. To avoid instabilities, we take the logarithm of the Gumbel-Softmax probability distribution (known as the \textsc{ExpConcrete} distribution) \citep{Maddison2016} given by $x_i=\log \left(\text{SM}\left(\frac{\log \pi_i+g_i}{\tau}\right)\right)$. Expanding the logarithm, we get that the $i$th element \textsc{ExpConcrete} random variable is defined as
\begin{align}
    x_i=\frac{\log \pi_i+(g_i/\beta)}{\tau}-\log\sum_{j=1}^V \exp\left(\frac{\log \pi_j+(g_j/\beta)}{\tau}\right)\label{eq:ExpConcrete}
\end{align}
Translating this into our time-varying interpolant where $\pi_i=\exp(\delta_{ik})$, we define
\begin{align}
    \psi_t(\mathbf{x}_1=\mathbf{e}_k)=\frac{\delta_{ik}+(g_i/\beta)}{\tau(t)}-\log \sum_{j=1}^V\exp\left(\frac{\delta_{jk}+(g_j/\beta)}{\tau(t)}\right)
\end{align}
By our derivation in Appendix \ref{appendix:Conditional Score Function}, the \textit{score} defined as the gradient of the log-probability density of the \textsc{ExpConcrete} interpolant with respect to the $i$th element $x_{t, i}$ is given by
\begin{align}
    \nabla_{x_{t, i}}\log p_t(\mathbf{x}_t |\mathbf{x}_1)&=-\tau(t) +\tau(t)V\cdot \text{SM}\bigg(\delta_{ik}-\tau(t) x_{t, i}\bigg)
\end{align}

\subsection{Learning the Gumbel-Softmax Probability Density}
Given that the Gumbel-Softmax interpolant naturally converges towards the one-hot target token distribution, it follows that learning the evolution of probability density across training samples would enable generation in regions with high probability density. Our goal is to train a parameterized model to learn to estimate the gradient of the log-probability density of the Gumbel-Softmax interpolant such that the gradient converges at regions with high probability density. To achieve this, we define the score parameterization similar to \citep{Mahmood2024}, given by
\begin{align}
    s_{\theta}(\mathbf{x}_t, t)&=-\tau(t) +\tau(t)V\cdot \text{SM}\big(f_{\theta}(\mathbf{x}_t, t)\big)\;\;\;\;\text{where}\;\;\;s_{\theta}(\mathbf{x}_t, t)\approx\nabla_{x_{t, j}}\log p_t(\mathbf{x}_t)\label{eq:Score Matching Parameterization}
\end{align}
where $\theta $ minimizes the reparameterized score-matching loss given by
\begin{small}
\begin{align}
    \mathcal{L}_{\text{score}}&=\mathbb{E}_{p_t(\mathbf{x}_t|\mathbf{x}_1), p_1(\mathbf{x}_1)}\bigg\| \big[-\tau(t) +\tau(t)V\cdot \text{SM}(\delta_{ik}-\tau(t) x_{t,i})\big]-\big[-\tau(t) +\tau(t)V\cdot \text{SM}(f_{\theta}(\mathbf{x}_t, t)\big] \bigg\|^2\nonumber\\
    &=\tau(t)^2V^2\mathbb{E}_{p_t(\mathbf{x}_t|\mathbf{x}_1), p_1(\mathbf{x}_1)}  \|\text{SM}\big(\delta_{ik}-\tau(t)x_{t, i}\big)-\text{SM}(f_{\theta}(\mathbf{x}_t, t)\big)\|^2
\end{align}
\end{small}

The \texttt{softmax} function applied after parameterization ensures dependencies are preserved across the predicted output vector which defines the rate of probability flow towards each vertex. Since $\tau(t)\to 0$ when $t\to 1$, we remove the scaling term to ensure the losses are evenly scaled over time.
\begin{align}
    \mathcal{L}_{\text{score}}=\mathbb{E}_{p_t(\mathbf{x}_t|\mathbf{x}_1), p_1(\mathbf{x}_1)} \|\text{SM}\big(\delta_{ik}-\tau(t)x_{t, i}\big)-\text{SM}(f_{\theta}(\mathbf{x}_t, t)\big)\|^2
\end{align}

\begin{proposition}
    The gradient of the \textsc{ExpConcrete} log-probability density is proportional to the gradient of the Gumbel-softmax log-probability density such that $\nabla_{x_j}^{\text{GS}}\log p_{\theta}(\mathbf{x}_t|\mathbf{x}_1)\propto \nabla_{x_j}^{\text{ExpConcrete}}\log p_{\theta}(\mathbf{x}_t|\mathbf{x}_1)$. 
\end{proposition}

Proof in Appendix \ref{appendix:Concrete and ExpConcrete Score Function}. Therefore, by minimizing $\mathcal{L}_{\text{score}}$, we obtain a model that effectively transports intermediate Gumbel-Softmax distributions towards clean distributions in high-probability regions of the discrete state space. 

\section{Straight-Through Guided Flows (STGFlow)}\label{section:Straight-Through Guided Flows}
In this section, we present \textbf{Straight-Through Guided Flows} (STGFlow) — a novel classifier-based guidance method that guides the pre-trained conditional flow velocities towards sequences with higher classifier probabilities $p^{\phi}(y|\mathbf{x}_t)$ which does not require training a time-dependent classifier or classifier-guided velocity field. STGFlow leverages straight-through gradient estimators to compute gradients of classifier scores from discrete sequence samples with respect to the continuous logits from which they were sampled. The unconditionally predicted logits are refined using the gradients in a temperature-dependent manner, sharpening the guidance as $t\to 1$. 

\subsection{Straight-Through Gradient Estimators}

\begin{wrapfigure}{r}{0.5\textwidth}
\includegraphics[width=\linewidth]{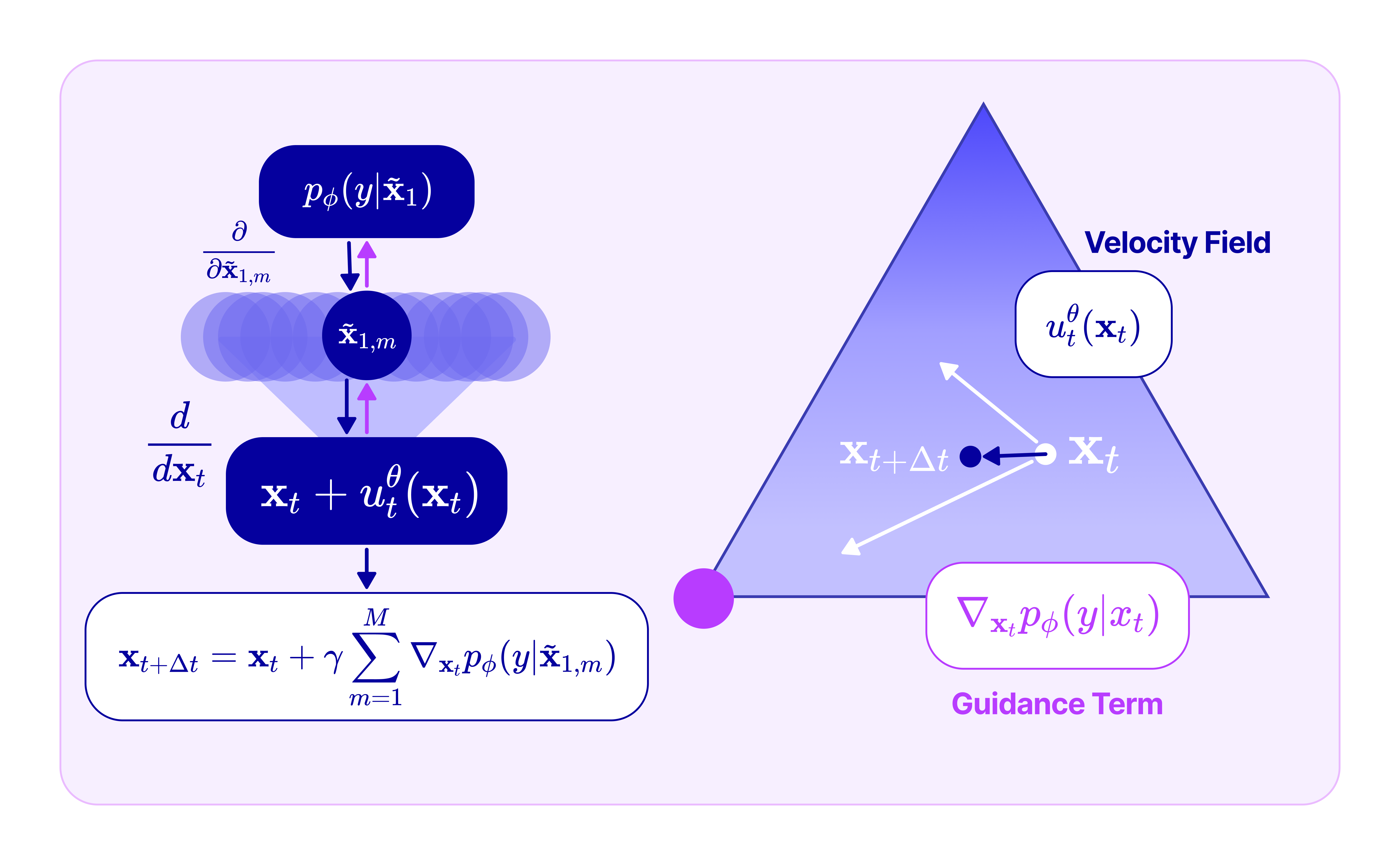} 
\caption{\textbf{Straight-Through Guided Flows (STGFlow).} We compute the gradients of the classifier function with respect to $M$ discrete sequences sampled from the intermediate token distribution $\mathbf{x}_t$, which act as a guided flow velocity that steers the unconditional trajectory towards sequences with optimal scores. }
\label{fig:Straight-Through Guided Flows}
\end{wrapfigure}

Straight-through gradient estimators aim to solve the problem of taking gradients with respect to \textit{discrete random variables}. Consider a reward function $\mathcal{R}(\mathbf{z})$ that takes a discrete sequence $\mathbf{z}$ of length $L$ sampled from a learned distribution $p_{\theta}(\mathbf{z})$, and our goal is to maximize the reward
\begin{align}
    \max_{\theta}\mathcal{R}=\min_{\theta}\mathbb{E}_{\mathbf{z}\sim p_{\theta}}\big[\mathcal{R}(\mathbf{z})\big]
\end{align}
Given the non-differentiability of $\mathcal{R}(\mathbf{z})$ with respect to the parameters $\theta$, the Straight-Through Gumbel-Softmax estimator (ST-GS) \cite{Jang2016} evaluates the gradient of the reward function through a \textit{surrogate} of the discrete random variable $\mathbf{z}$ defined as the tempered \texttt{softmax} distribution over the continuous logits from which $\mathbf{z}$ was sampled. 
\begin{align}
    \nabla_{\theta}\mathcal{R}=\frac{\partial \mathcal{R}(\mathbf{z})}{\partial \mathbf{z}}\frac{d}{d\theta}\text{SM}_{\tau }(p_{\theta}(\mathbf{z}))
\end{align}
ST-GS preserves the forward evaluation of the reward function while enabling low-variance gradient estimation for back-propagation of the gradient that does not need to be defined over continuous relaxations of discrete variables over the simplex. Instead, they only need to be defined for \textit{discrete} sequences, which is the case for most pre-trained classifier models. 

\subsection{Straight-Through Guided Flow Matching} 
We extend the idea of ST-GS to define a novel post-training guidance method. At each time step $t$, we compute the Gumbel-Softmax velocity field $u^{\theta}_t(\mathbf{x}_t)$ and take a step. Then, from the updated logits, we sample $M$ discrete sequences $\{\tilde{\mathbf{x}}_{1,1}, \dots, \tilde{\mathbf{x}}_{1,M}\}$ from the top $k$ logits in $\mathbf{x}_t$ re-normalized with the \texttt{softmax} function. For each sequence $\tilde{\mathbf{x}}_{1,m}$, we compute a classifier score using our pre-trained classifier $p_{\phi}(y|\tilde{\mathbf{x}}_{1,m})$. Since the gradient through the \texttt{argmax} function is either 0 or undefined, we compute the gradient of the classifier model with respect to the surrogate \texttt{softmax} distribution. 
\begin{align}
    \nabla_{\mathbf{x}_t}p_{\phi}(y|\tilde{\mathbf{x}}_{1, m})= \frac{\partial p_{\phi}(y|\tilde{\mathbf{x}}_{1, m})}{\tilde{\mathbf{x}}_{1, m}}\frac{d}{d\mathbf{x}_t}\text{SM}\big(\mathbf{x}_t\big)
\end{align}
Evaluating the straight-through gradient with respect to the probability of each token, we have
\begin{align}
    \nabla_{x_{t, i}}p_{\phi}(y|\tilde{\mathbf{x}}_{1, m})=\begin{cases}
        \frac{\partial p_{\phi}(y|\tilde{\mathbf{x}}_{1, m})}{\tilde{\mathbf{x}}_1}\cdot \big[\text{SM}(x_{t, i})\left(1-\text{SM}(x_{t, k})\right)\big]&i=k\\
        \frac{\partial p_{\phi}(y|\tilde{\mathbf{x}}_{1, m})}{\tilde{\mathbf{x}}_1}\cdot\big[-\text{SM}(x_{t, i})\text{SM}(x_{t, k})\big]&i\neq k
    \end{cases}
\end{align}
where $k$ denotes the index of the sampled token such that $\tilde{\mathbf{x}}_{1, m}=\mathbf{e}_k$. During inference, the partial derivative term $\frac{\partial p_{\phi}(y|\tilde{\mathbf{x}}_{1, m})}{\tilde{\mathbf{x}}_{1, m}}$ is computed with automatic differentiation with respect to each sequence position, enabling position-specific guidance. Finally, we guide the flow trajectory by adding the aggregate gradient across all $M$ sequences scaled by a constant $\gamma$ to get
\begin{align}
    \mathbf{x}_t=\mathbf{x}_t+\gamma\sum_{m=1}^M\nabla_{\mathbf{x}_t}p_{\phi}(y|\tilde{\mathbf{x}}_{1, m})
\end{align}

\begin{proposition} [Conservation of Probability Mass of Straight-Through Gradient]
The straight through gradient $\nabla_{\mathbf{x}_t}p_{\phi}(y|\tilde{\mathbf{x}}_{1, m})$ preserves probability mass and lies on the tangent bundle at point $\mathbf{x}_t$ on the simplex. 
\end{proposition}

Proof in Appendix \ref{appendix:Probability Mass Conservation of Straight-Through Gradient}. Conceptually, the straight-through gradient acts as a guiding velocity that steers the unconditional velocity toward valid, optimal sequences. Pseudocode for STGFlow is provided in Algorithm \ref{alg:Straight-Through Guided Flow Matching}.

%%%%%%%%%%%%%%%%%%%%%%%%%%%%%%%%%%%%%%%%%%%%%%%%%%%%%%%%%%%%
%%%%%%%%%%%%%%%%%%%%%%%%%%%%%%%%%%%%%%%%%%%%%%%%%%%%%%%%%%%%
\section{Experiments}\label{section:Experiments}

\subsection{Simplex-Dimension Toy Exepriment}

\textbf{Setup.} Following \citet{Stark2024}, we conduct a toy experiment that evaluates the KL divergence between the empirically-generated distribution and a random distribution of sequence length 4 over the $(V-1)$-dimensional simplex $(\Delta^{V-1})^4$ for $K=\{20, 40, 60, 80, 100, 120, 140, 160, 512\}$. The sequence length is set to 4 and the number of integration steps was set to 100 across all experiments.

\textbf{Training.} We trained Linear FM \cite{Stark2024}, Dirichlet FM \cite{Stark2024}, Fisher FM \cite{Davis2024}, and Gumbel-Softmax FM each for $50$K steps on $100$K sequences from a randomly generated distribution. We evaluated the KL divergence $\text{KL}(\tilde{q} \|p_{\text{data}})$ where $\tilde{q}$ is the normalized distribution from $51.2$K sequences generated by the model and $p_{\text{data}}$ is the distribution from which the training data was sampled.  

\textbf{Results.} As shown in Table \ref{table:Toy Experiment}, Gumbel-Softmax FM achieves superior performance to Dirichlet FM when scaled to dimensions $K \geq 60$, with stable KL divergence in the range $0.02-0.05$ for all simplex dimensions up to $K=512$. Although Gumbel-Softmax FM achieves higher KL divergence than Fisher FM, we note that the use of optimal transport in Fisher FM results in learning straight, deterministic flows that can result in overfitting to the training data. This can be observed when comparing the curves of the validation mean-squared error loss between the predicted and true conditional velocity fields summed over the simplex and sequence length dimensions (Figure \ref{fig:Validation Loss}).

\subsection{Promoter DNA Sequence Design}
Following the procedures of previous works \citep{Avdeyev2023, Stark2024}, we evaluate Gumbel-Softmax FM for conditional DNA promoter design and show superior performance to discrete diffusion and flow-matching baselines. 

\textbf{Setup.} Promoter DNA is the strand of DNA adjacent to a gene that binds to RNA polymerase and transcription factors to promote gene transcription and expression. The objective is to train a conditional flow model with the regulatory signal concatenated to the noisy input sequence to minimize the mean squared error (MSE) between the predicted regulatory activity of the generated sequence with the true sequence, predicted with a pre-trained Sei model \citep{Chen2022}.

\begin{wraptable}{r}{7cm}
\begin{small}
\centering
\begin{tabular}{@{}lc@{}}
\toprule
\textbf{Model}& \textbf{MSE} ($\downarrow$) \\
\midrule
Bit Diffusion (Bit Encoding)* & 0.041\\
Bit Diffusion (One-Hot Encoding)* & 0.040\\
D3PM-Uniform* & 0.038\\
DDSM* & 0.033\\
Language Model* & 0.033\\
\midrule
Dirichlet Flow Matching & 0.029 \\
Fisher Flow Matching & 0.030\\
\textbf{Gumbel-Softmax Flow Matching} (Ours)  & 0.029 \\ 
\bottomrule
\end{tabular}
\caption{\textbf{Evaluation of promoter DNA generation conditioned on transcription profile.} MSE was evaluated across all validation batches between the predicted signal of a conditionally generated sequence and the true sequence. Regulatory signals were predicted with a pre-trained Sei model \cite{Chen2022}. Numbers with * are from \citet{Stark2024}}
\label{table:promoter design}
\end{small}
\end{wraptable}

\textbf{Training.}  Following \citet{Stark2024}, we trained on a train/test/validation split of $88,470/3,933/7,497$ promoter sequences that are 1,024 base pairs in length. For each batch of size $256$, we applied the Gumbel-Softmax transformation according to Equation \ref{eq:gumbel softmax interpolant} with $\tau_{\text{max}}=10.0$ and $\lambda = 3.0$ for uniformly distributed time steps $t\in [0, 1]$ over each training batch. The training objective was to minimize the negative log loss between the true one-hot tokens $\mathbf{x}_1$ and predicted logits $\mathbf{x}_{\theta}(\mathbf{x}_t, t)$ from varying temperatures dependent on uniformly sampled $t\sim \mathcal{U}(0,1)$. We trained Dirichlet FM \cite{Stark2024}, Fisher FM \citep{Davis2024}, and Gumbel-Softmax FM parameterized with a $20$-layer 1D CNN architecture for $150$K steps and evaluated the MSE across all validation batches. 

\textbf{Results. }The MSE values for the diffusion and autoregressive language model baselines \cite{Avdeyev2023, Chen2023, Austin2021} were taken from \cite{Stark2024, Davis2024}, but the simplex-based flow baselines were retrained. Gumbel-Softmax FM produces lower signal MSE compared to diffusion and language model baselines and similar MSE to Dirichlet and Fisher FM. 

\subsection{\textit{De Novo} Protein Sequence Design}
\begin{table*}
\caption{\textbf{Evaluation metrics for generative quality of protein sequences.} Metrics were calculated on 100 unconditionally generated sequences from each model, including EvoDiff and ProtGPT2. The arrow indicates whether $(\uparrow)$ or $(\downarrow)$ values are better. }
\label{table:Protein Design}
\begin{center}
\begin{small}
\resizebox{\linewidth}{!}{
\begin{tabular}{@{}lccccccc@{}}
\toprule
\textbf{Model} & \textbf{Params} ($\downarrow$) & \textbf{pLDDT} ($\uparrow$) & \textbf{pTM} ($\uparrow$)& \textbf{pAE} ($\downarrow$) & \textbf{Entropy} ($\uparrow$)& \textbf{Diversity} (\%) ($\uparrow$) \\
\midrule
Test Dataset (random 1000) & - & 74.00 & 0.63 & 12.99 & 4.0 & 71.8 \\
\midrule
EvoDiff & 640M & 31.84 & 0.21 & 24.76 & 4.05 & 93.2 \\
ProtGPT2 & 738M & 54.92 & 0.41 & 19.39 & 3.85 & 70.9 \\
ProGen2-small & 151M & 49.38 & 0.28 & 23.38 & 2.55 & 89.3 \\
\textbf{Gumbel-Softmax Flow Matching} (Ours) & 198M & 52.54 & 0.27 & 16.67 & 3.41 & 86.1  \\
\textbf{Gumbel-Softmax Score Matching} (Ours) & 198M & 49.40 & 0.29 & 15.71 & 3.37 & 82.5 \\

\bottomrule
\end{tabular}
}
\end{small}
\end{center}
\end{table*}

Next, we evaluate the quality of unconditionally-generated \textit{de novo} protein sequences with Gumbel-Softmax SM and Gumbel-Softmax FM. Despite operating in the continuous simplex space with a considerably smaller backbone model, we demonstrate competitive generative quality compared to discrete diffusion and autoregressive baselines. 

\begin{wrapfigure}{r}{0.7\textwidth}
\includegraphics[width=\linewidth]{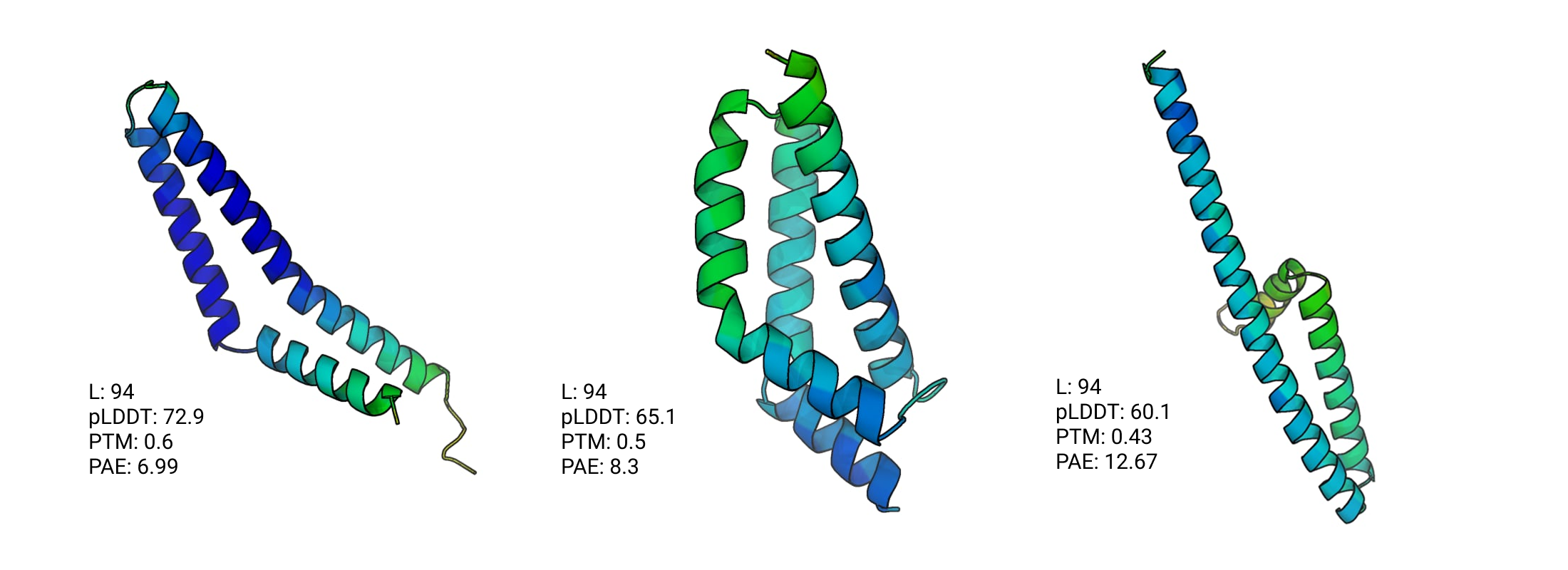} 
\caption{\textbf{Predicted structures of \textit{de novo} generated proteins from Gumbel-Softmax FM.} The structures, pLDDT, pAE, and pTM scores are predicted with ESMFold \citep{Lin2023-gh}}
\label{fig:Proteins}
\end{wrapfigure}

\textbf{Setup.} Given the larger vocabulary size of protein sequences, we compared both the performance of Gumbel-Softmax FM and Gumbel-Softmax SM for this task. For both models, we applied the Gumbel-Softmax transformation with varying temperatures $\tau(t)$ for time steps $t\sim \mathcal{U}(0, 1)$ and $\tau_{\text{max}}=10.0$. The decay rates were set to $\lambda = 3.0$ for both models and the noise scale was set to $\beta=2.0$. The models were trained following Algorithm \ref{alg:Training Gumbel-Softmax FM} for Gumbel-Softmax FM and \ref{alg:Training Gumbel-Softmax SM} for Gumbel-Softmax SM. Sampling was performed following Algorithm \ref{alg:Sampling Gumbel-Softmax FM} and Algorithm \ref{alg:Sampling Gumbel-Softmax SM}.

\textbf{Training.} We collected 68M Uniref50 and $207$M OMG\_PROT50 data \citep{suzek2007uniref, Cornman2024}. A total of $275$M protein sequences were first clustered to remove singletons using MMseqs2 linclust \citep{steinegger2018clustering} (parameters set to \texttt{--min-seq-id 0.5}  \texttt{-c 0.9} \texttt{--cov-mode 1}). We keep the sequences between lengths of $20$ to $2500$ and entries with only wild-type residues to avoid effects from outliers. The singleton sequences are removed. The resulting representative sequences undergo random 0.8/0.1/0.1 data splitting. We trained for $5$ epochs on $7$ NVIDIA $A100$ GPUs. 

\textbf{Results.} We compare the quality of our protein generation method against state-of-the-art \textit{de novo} protein sequence generation models including the discrete diffusion model EvoDiff \citep{Alamdari2023}, large language model ProtGPT2 \citep{Ferruz2022}, and the autoregressive model ProGen2-small \citep{Nijkamp2023}. For 100 unconditionally generated sequences per model, we compute the pLDDT, pTM, pAE scores using ESMFold \citep{Lin2023} as well as the token entropy and sequence diversity. Additional details on evaluation metrics are given in Appendix \ref{appendix:Protein Evaluation Metrics}. BLASTp runs for the proteins we generated indicate no homolog hits, highlighting again the novelty of the proteins we generated and indicating that our model is not sub-sampling from known homologous protein sequences. As summarized in Table \ref{table:Protein Design}, both Gumbel-Softmax SM and Gumbel-Softmax FM produce proteins with comparable pLDDT, pTM, and pAE scores to discrete baselines without significantly compromising sequence entropy and diversity. We believe further optimization of hyperparameters, leveraging informative priors, or functional/structural guidance would improve the generative quality of Gumbel-Softmax FM.

\subsection{Peptide Binder Design}\label{section:Peptide Binder Design}
\begin{figure*}
    \centering
    \includegraphics[width=\linewidth]{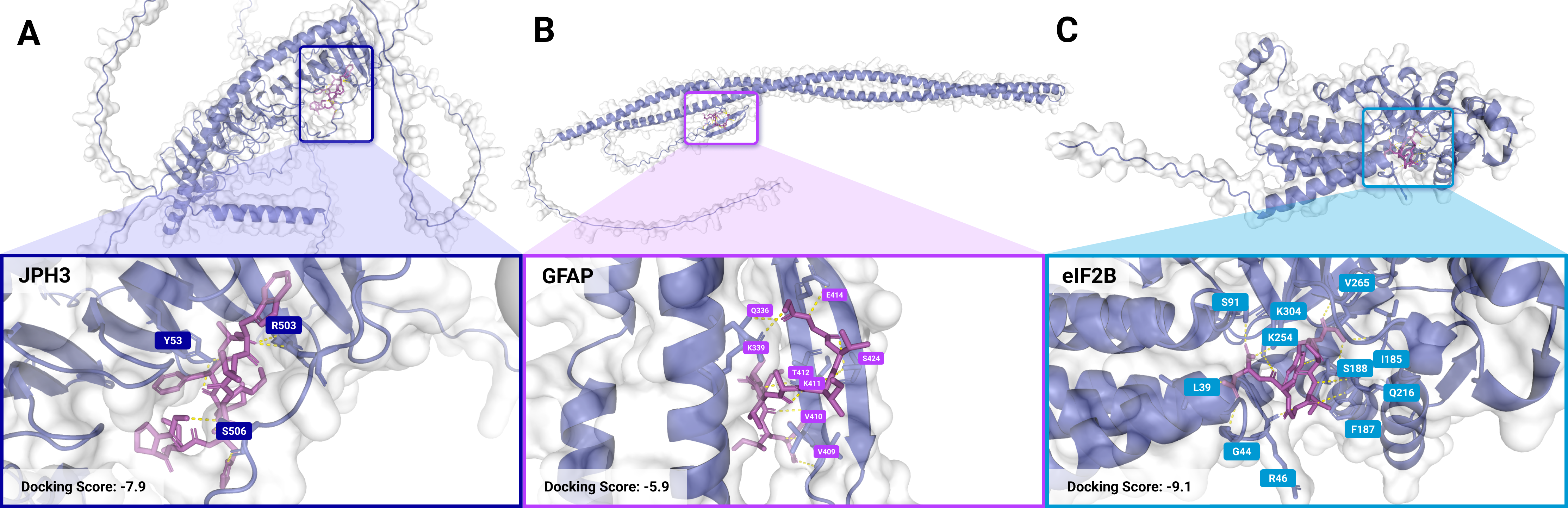}
    \caption{\textbf{Gumbel-Softmax FM generated peptide binders for three targets with no known binders.} (\textbf{A}) $10$ a.a. designed binder to JPH3 (structure generated with AlphaFold3) involved in Huntington’s Disease-Like 2. (\textbf{B}) $10$ a.a. designed binder to GFAP (PDB: 6A9P) involved in Alexander Disease. (\textbf{C}) $7$ a.a. designed binder to eIF2B (PDB: 6CAJ) involved in Vanishing White Matter Disease. Docked with AutoDock VINA and polar contacts within $3.5$ \AA\ are annotated. Additional targets are shown in Table \ref{table:Peptide No Existing Binder}.}
    \label{fig:Peptides No Existing Binder}
\end{figure*}

Finally, we integrate guidance into Gumbel-Softmax FM to generate \textit{de novo} peptides with high binding affinity to protein targets. We generate peptide binders with similar or higher binding affinity to proteins with known peptide binders and diverse, rare disease-associated proteins without known peptide binders.

\textbf{Setup.} First, we generated \textit{de novo} peptide binders for 10 structured targets with known peptide binders using our STGFlow algorithm (Algorithm \ref{alg:Straight-Through Guided Flow Matching}). To guide the flow paths, we train a target-binding cross-attention-based regression model (Appendix \ref{appendix:Peptide-Binding Affinity Classifier}) that takes an amino acid representation of a peptide binder and protein target and predicts the $K_d/K_i/IC50$ score, where scores $<6.0$ indicate weak binding, scores within $6.0-7.5$ indicate medium binding, and scores $>7.5$ indicate strong binding. Using a dataset of $1781$ experimentally validated peptides, our model achieved a strong Spearman correlation coefficient of $0.96$ on the training set and $0.64$ on the validation set. 

\textbf{Training.} We fine-tuned our Gumbel-Softmax FM protein generator for $600$ epochs on $17,479$ peptides ($0.8/0.2$ train/validation split) between $6-50$ amino acids in length curated from the PepNN \cite{Abdin2022}, BioLip2 \cite{BioLiP2}, and PPIRef \cite{Bushuiev2023} datasets.

\begin{wrapfigure}{r}{0.7\textwidth}
\includegraphics[width=\linewidth]{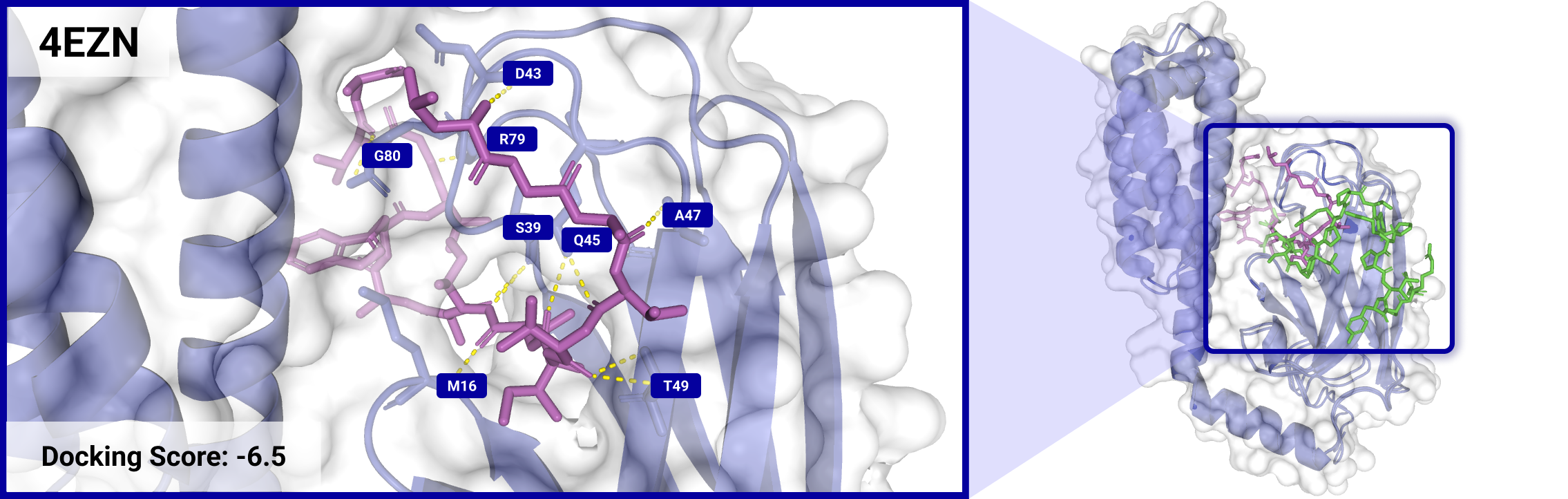} 
\caption{\textbf{Comparison of existing and Gumbel-Softmax FM designed binder to protein 4EZN.} AutoDock VINA docking score of the designed binder ($-6.5$ kcal/mol; magenta) is lower than that of the existing binder ($-4.1$ kcal/mol; green) indicating stronger binding affinity. Polar contacts within $3.5$ \AA\ are annotated. Additional comparisons of existing and designed binders are in Table \ref{table:Peptide Existing Binder}. }
\label{fig:EZN Peptide}
\end{wrapfigure}

\textbf{Results.} First, we compare peptide binders generated by Gumbel-Softmax FM coupled with STGFlow guidance to existing peptide binders to $13$ protein targets (Table \ref{table:Peptide Existing Binder}). After generating $20$ \textit{de novo} peptides of the same length as the existing binders, we computed the ipTM and pTM scores using AlphaFold3 to evaluate the predicted confidence of the peptide-protein complexes and the docking scores using AutoDock VINA to evaluate the free energy of the binding interaction (See Appendix \ref{appendix:Peptide Evaluation Metrics} for details on evaluation metrics). From the final \textit{de novo} generated peptides with optimized classifier scores against each target, we show that Gumbel-Softmax FM can consistently generate peptides with superior ipTM ($\uparrow$) and VINA docking scores ($\downarrow$) compared to experimentally-validated binders (Table \ref{table:Peptide Existing Binder}), indicating the efficacy of guided flow matching strategy in generating peptides with high binding affinity. 

To further validate the versatility of our framework, we evaluated peptide binders guided for six proteins involved in various diseases with no pre-existing peptide binders (Figure \ref{fig:Peptides No Existing Binder}; Table \ref{table:Peptide No Existing Binder}). We generated 20 peptide binders that are $5-15$ amino acids in length with Gumbel-Softmax FM and STGFlow guidance and randomly permuted the sequence to generate a scrambled negative control for comparison. Notably, our designed binders demonstrate strong ipTM higher than 0.62 and VINA docking scores below $-5.9$. Despite the short sequence length, we also show that scrambling the order of amino acids consistently decreases the binding affinity compared to the unscrambled binder, indicating that our guidance strategy effectively captures dependencies across tokens that lead to higher-affinity peptides (Table \ref{table:Peptide No Existing Binder}). Furthermore, the docked peptides show complementary structures to the target protein with several polar contacts within $3.5$ \AA\ (Figure \ref{fig:Peptides No Existing Binder}).

Since pTM ($\uparrow$) scores are dominated by the confidence in the protein target structure, there are no significant differences in the scores between the designed binders and control peptides; however, we still observe slightly higher scores indicating that our designed binders enhance the stability of the protein structure. Plotting the predicted binding affinity scores over the iteration or time step, we consistently see sharp upward curves, which proves the efficacy of STGFlow in optimizing classifier scores (Figure \ref{fig:Guidance Curves}). 

\begin{table*}
\caption{\textbf{Comparison of ipTM and VINA docking scores for existing and designed peptide binders to protein targets.} The ipTM scores are calculated by AlphaFold3 for peptide-protein complexes using both existing peptides and peptides designed by guided Gumbel-Softmax FM. *Contains unnatural amino acid X which cannot be processed by AlphaFold3.}
\label{table:Peptide Existing Binder}
\begin{center}
\begin{small}
\resizebox{\linewidth}{!}{
\begin{tabular}{@{}llccccccc@{}}
\toprule
\textbf{PDB ID} & \textbf{existing binder} &\multicolumn{2}{c}{\textbf{ipTM ($\uparrow$)}} & \multicolumn{2}{c}{\textbf{pTM ($\uparrow$)}} & \multicolumn{2}{c}{\textbf{VINA Docking Score (kcal/mol) ($\downarrow$)}} \\ \midrule
 & & existing & designed & existing & designed & existing & designed \\
\midrule
GLP-1R (3C5T) & HXEGTFTSDVSSYLEGQAAKEFIAWLVRGRG & * & 0.65 & * & 0.66 & -5.7 & -7.5  \\
1AYC & ARLIDDQLLKS & 0.68 & 0.67 & 0.88 & 0.88 & -5.3 & -4.6  \\
2Q8Y & ALRRELADW & 0.44 & 0.70 & 0.83 & 0.84 & -6.7 & -6.8 \\
3EQS & GDHARQGLLALG & 0.80 & 0.71 & 0.88 & 0.86 & -4.4 & -4.7 \\
3NIH & RIAAA & 0.85 & 0.86 & 0.91 & 0.90 & -6.2 & -5.7 \\
4EZN & VDKGSYLPRPTPPRPIYNRN & 0.54 & 0.59 & 0.85 & 0.87 & -4.1 & -6.5 \\
4GNE & ARTKQTA & 0.89 & 0.76 & 0.76 & 0.76 & -5.0 & -4.8 \\
4IU7 & HKILHRLLQD & 0.93 & 0.79 & 0.91 & 0.94 & -4.6 & -5.9  \\
5E1C & KHKILHRLLQDSSS & 0.83 & 0.80 & 0.91 & 0.91 & -4.3 & -5.1 \\
5EYZ & SWESHKSGRETEV & 0.73 & 0.81 & 0.77 & 0.78 & -2.9 & -6.9 \\
5KRI & KHKILHRLLQDSSS & 0.83 & 0.77 & 0.91 & 0.91 & -3.5 & -5.5 \\
7LUL & RWYERWV & 0.94 & 0.91 & 0.93 & 0.92 & -6.5& -7.6 \\
8CN1 & ETEV & 0.90 & 0.86 & 0.72 & 0.82 & -6.0 & -6.9  \\

\bottomrule
\end{tabular}
}
\end{small}
\end{center}
\end{table*}

\begin{table*}
\caption{\textbf{Comparison of ipTM and VINA docking scores for designed peptide binders and scrambled negative control to protein targets with no known binders.} The ipTM and pTM scores are calculated by AlphaFold3 and docking scores are calculated by AutoDock VINA for peptides designed by Gumbel-Softmax FM with STGFlow. Designed sequences are randomly permuted to generate a scrambled negative control for comparison. *No PDB structure available. Used AlphaFold3 predicted structure for docking.}
\label{table:Peptide No Existing Binder}
\begin{center}
\resizebox{\linewidth}{!}{
\begin{tabular}{@{}lllccccccc@{}}
\toprule
\textbf{PDB ID} & \textbf{Protein Name} & \textbf{Disease}  &\multicolumn{2}{c}{\textbf{ipTM ($\uparrow$)}} & \multicolumn{2}{c}{\textbf{pTM ($\uparrow$)}} & \multicolumn{2}{c}{\textbf{VINA Docking Score (kcal/mol) ($\downarrow$)}} \\
\midrule
& & & designed & scramble & designed & scramble & designed & scramble \\
\midrule
 6A9P & GFAP & Alexander Disease  & 0.62 & 0.38 & 0.31 & 0.29 & -5.9 & -3.7   \\
 6CAJ & eIF2B & Vanishing White Matter Disease & 0.61 & 0.39 & 0.77 & 0.76 & -9.1 & -9.0 \\
 3HVE & Gigaxonin & Giant Axonal Neuropathy  & 0.75 & 0.54 & 0.83 & 0.82 & -6.8 & -6.2 \\
 6W5V & NPC2 & Niemann-Pick Disease Type C  & 0.80 & 0.34 & 0.79 & 0.77 & -6.5 & -5.6 \\
 * & JPH3 & Huntington’s Disease-Like 2 (HDL2)  & 0.72 & 0.60 & 0.49 & 0.49 & -7.9 & -7.8  \\
 2CKL & BMI1 & Medulloblastoma & 0.71 & 0.43 & 0.81 & 0.73 & -6.8 & -6.2 \\
\bottomrule
\end{tabular}
}
\end{center}
\end{table*}

%%%%%%%%%%%%%%%%%%%%%%%%%%%%%%%%%%%%%%%%%%%%%%%%%%%%%%%%%%%%
%%%%%%%%%%%%%%%%%%%%%%%%%%%%%%%%%%%%%%%%%%%%%%%%%%%%%%%%%%%%

\section{Conclusion}

In this work, we introduce \textbf{Gumbel-Softmax Flow and Score Matching}, a novel discrete framework that learns interpolations between noisy and clean data by modulating the temperature of the Gumbel-Softmax distribution. By parameterizing a straight continuous-time interpolation with stochastic Gumbel noise, we overcome limitations of existing discrete generative models, such as computationally expensive iterative denoising in discrete diffusion \citep{Austin2021, Wang2024, Shi2024, Sahoo2024}, high variance training in Dirichlet Flow Matching \citep{Stark2024}, and restrictive probability constraints in Fisher Flow Matching \citep{Davis2024}.

We apply our model to three key biological sequence generation tasks: conditional DNA promoter design, \textit{de novo} protein sequence generation, and target-binding peptide design. For promoter design, Gumbel-Softmax FM generates functional DNA sequences with enhanced transcriptional activity, outperforming previous discrete generative approaches. For target-protein guided peptide binder design with STGFlow, our \textit{de novo} peptides show superior binding affinity against known binders for $10$ proteins and strong binding affinity to six rare neurological disease-associated proteins with no known peptide binders, opening up numerous therapeutic opportunities for these understudied diseases. For protein sequence generation, our method enables the design of structurally feasible proteins while maintaining sequence diversity and uniqueness against known proteins.

By bridging discrete flow matching with Gumbel-Softmax relaxations, our work provides a scalable and theoretically grounded framework for discrete sequence modeling on the simplex. Future directions include extending the approach to multi-objective sequence optimization, incorporating task-specific priors to enhance design constraints, and applying Gumbel-Softmax FM to other structured biological design problems, such as RNA sequence engineering and regulatory circuit design.

\section{Declarations}
\textbf{Acknowledgments.} We thank the Duke Compute Cluster, Pratt School of Engineering IT department, and Mark III Systems, for providing database and hardware support that has contributed to the research reported within this manuscript. 

\textbf{Author Contributions.} S.T. devised and developed model architectures and theoretical formulations, and trained and benchmarked models. Y.Z. advised on model design and theoretical framework, trained and benchmarked models, and performed molecular docking. S.T. drafted the manuscript and S.T. and Y.Z. designed the figures. A.T. reviewed mathematical formulations and provided advising. P.C. designed, supervised, and directed the study, and reviewed and finalized the manuscript.

\textbf{Data and Materials Availability.} The codebase will be freely accessible to the academic community at \url{https://huggingface.co/ChatterjeeLab/GumbelFlow}.

\textbf{Funding Statement.} This research was supported by NIH grant R35GM155282 as well as a grant from the EndAxD Foundation to the lab of P.C.

\textbf{Competing Interests.} P.C. is a co-founder of Gameto, Inc. and UbiquiTx, Inc. and advises companies involved in peptide therapeutics development. P.C.’s interests are reviewed and managed by Duke University in accordance with their conflict-of-interest policies. S.T., Y.Z., and A.T. have no conflicts of interest to declare.

%%%%%%%%%%%%%%%%%%%%%%%%%%%%%%%%%%%%%%%%%%%%%%%%%%%%%%%%%%%%
%%%%%%%%%%%%%%%%%%%%%%%%%%%%%%%%%%%%%%%%%%%%%%%%%%%%%%%%%%%%
%%%%%%%%%%%%%%%%%%%%%%%%%%%%%%%%%%%%%%%%%%%%%%%%%%%%%%%%%%%%
%%%%%%%%%%%%%%%%%%%%%%%%%%%%%%%%%%%%%%%%%%%%%%%%%%%%%%%%%%%%
\bibliographystyle{unsrtnat}
\bibliography{citation}  

\begin{thebibliography}{51}
\providecommand{\natexlab}[1]{#1}
\providecommand{\url}[1]{\texttt{#1}}
\expandafter\ifx\csname urlstyle\endcsname\relax
  \providecommand{\doi}[1]{doi: #1}\else
  \providecommand{\doi}{doi: \begingroup \urlstyle{rm}\Url}\fi

\bibitem[Madani et~al.(2023)Madani, Krause, Greene, Subramanian, Mohr, Holton, Olmos, Xiong, Sun, Socher, Fraser, and Naik]{Madani2023}
Ali Madani, Ben Krause, Eric~R. Greene, Subu Subramanian, Benjamin~P. Mohr, James~M. Holton, Jose~Luis Olmos, Caiming Xiong, Zachary~Z. Sun, Richard Socher, James~S. Fraser, and Nikhil Naik.
\newblock Large language models generate functional protein sequences across diverse families.
\newblock \emph{Nature Biotechnology}, 41\penalty0 (8):\penalty0 1099–1106, January 2023.
\newblock ISSN 1546-1696.
\newblock \doi{10.1038/s41587-022-01618-2}.
\newblock URL \url{http://dx.doi.org/10.1038/s41587-022-01618-2}.

\bibitem[Ferruz et~al.(2022)Ferruz, Schmidt, and H\"{o}cker]{Ferruz2022}
Noelia Ferruz, Steffen Schmidt, and Birte H\"{o}cker.
\newblock Protgpt2 is a deep unsupervised language model for protein design.
\newblock \emph{Nature Communications}, 13\penalty0 (1), July 2022.
\newblock ISSN 2041-1723.
\newblock \doi{10.1038/s41467-022-32007-7}.
\newblock URL \url{http://dx.doi.org/10.1038/s41467-022-32007-7}.

\bibitem[Nisonoff et~al.(2025)Nisonoff, Xiong, Allenspach, and Listgarten]{Nisinoff2024}
Hunter Nisonoff, Junhao Xiong, Stephan Allenspach, and Jennifer Listgarten.
\newblock Unlocking guidance for discrete state-space diffusion and flow models.
\newblock \emph{International Conference on Learning Representations}, 2025.
\newblock \doi{10.48550/ARXIV.2406.01572}.
\newblock URL \url{https://arxiv.org/abs/2406.01572}.

\bibitem[Stark et~al.(2024)Stark, Jing, Wang, Corso, Berger, Barzilay, and Jaakkola]{Stark2024}
Hannes Stark, Bowen Jing, Chenyu Wang, Gabriele Corso, Bonnie Berger, Regina Barzilay, and Tommi Jaakkola.
\newblock Dirichlet flow matching with applications to dna sequence design.
\newblock \emph{ICML}, 2024.

\bibitem[Bhat et~al.(2025)Bhat, Palepu, Hong, Mao, Ye, Iyer, Zhao, Chen, Vincoff, Watson, Wang, Srijay, Kavirayuni, Kholina, Goel, Vure, Deshpande, Soderling, DeLisa, and Chatterjee]{Bhat2025}
Suhaas Bhat, Kalyan Palepu, Lauren Hong, Joey Mao, Tianzheng Ye, Rema Iyer, Lin Zhao, Tianlai Chen, Sophia Vincoff, Rio Watson, Tian~Z. Wang, Divya Srijay, Venkata~Srikar Kavirayuni, Kseniia Kholina, Shrey Goel, Pranay Vure, Aniruddha~J. Deshpande, Scott~H. Soderling, Matthew~P. DeLisa, and Pranam Chatterjee.
\newblock De novo design of peptide binders to conformationally diverse targets with contrastive language modeling.
\newblock \emph{Science Advances}, 11\penalty0 (4), January 2025.
\newblock ISSN 2375-2548.
\newblock \doi{10.1126/sciadv.adr8638}.
\newblock URL \url{http://dx.doi.org/10.1126/sciadv.adr8638}.

\bibitem[Chen et~al.(2024)Chen, Dumas, Watson, Vincoff, Peng, Zhao, Hong, Pertsemlidis, Shaepers-Cheu, Wang, Srijay, Monticello, Vure, Pulugurta, Kholina, Goel, DeLisa, Truant, Aguilar, and Chatterjee]{Chen2024pepmlm}
Tianlai Chen, Madeleine Dumas, Rio Watson, Sophia Vincoff, Christina Peng, Lin Zhao, Lauren Hong, Sarah Pertsemlidis, Mayumi Shaepers-Cheu, Tian~Zi Wang, Divya Srijay, Connor Monticello, Pranay Vure, Rishab Pulugurta, Kseniia Kholina, Shrey Goel, Matthew~P. DeLisa, Ray Truant, Hector~C. Aguilar, and Pranam Chatterjee.
\newblock Pepmlm: Target sequence-conditioned generation of therapeutic peptide binders via span masked language modeling.
\newblock \emph{arXiv}, 2024.
\newblock \doi{10.48550/ARXIV.2310.03842}.
\newblock URL \url{https://arxiv.org/abs/2310.03842}.

\bibitem[Tang et~al.(2024)Tang, Zhang, and Chatterjee]{Tang2024}
Sophia Tang, Yinuo Zhang, and Pranam Chatterjee.
\newblock Peptune: De novo generation of therapeutic peptides with multi-objective-guided discrete diffusion.
\newblock \emph{arXiv}, 2024.
\newblock \doi{10.48550/ARXIV.2412.17780}.
\newblock URL \url{https://arxiv.org/abs/2412.17780}.

\bibitem[Austin et~al.(2021)Austin, Johnson, Ho, Tarlow, and Berg]{Austin2021}
Jacob Austin, Daniel~D. Johnson, Jonathan Ho, Daniel Tarlow, and Rianne van~den Berg.
\newblock Structured denoising diffusion models in discrete state-spaces.
\newblock \emph{Advances in Neural Information Processing Systems}, 2021.
\newblock \doi{10.48550/ARXIV.2107.03006}.
\newblock URL \url{https://arxiv.org/abs/2107.03006}.

\bibitem[Wang et~al.(2024)Wang, Li, Wang, Zhang, Du, Jiang, Wu, Deng, Kang, Pan, Li, Wang, Yao, Hou, and Hsieh]{Wang2024}
Mingyang Wang, Shuai Li, Jike Wang, Odin Zhang, Hongyan Du, Dejun Jiang, Zhenxing Wu, Yafeng Deng, Yu~Kang, Peichen Pan, Dan Li, Xiaorui Wang, Xiaojun Yao, Tingjun Hou, and Chang-Yu Hsieh.
\newblock Clickgen: Directed exploration of synthesizable chemical space via modular reactions and reinforcement learning.
\newblock \emph{Nature Communications}, 15\penalty0 (1), November 2024.
\newblock ISSN 2041-1723.
\newblock \doi{10.1038/s41467-024-54456-y}.
\newblock URL \url{http://dx.doi.org/10.1038/s41467-024-54456-y}.

\bibitem[Shi et~al.(2024)Shi, Han, Wang, Doucet, and Titsias]{Shi2024}
Jiaxin Shi, Kehang Han, Zhe Wang, Arnaud Doucet, and Michalis~K. Titsias.
\newblock Simplified and generalized masked diffusion for discrete data.
\newblock \emph{Advances in Neural Information Processing Systems}, 2024.
\newblock \doi{10.48550/ARXIV.2406.04329}.
\newblock URL \url{https://arxiv.org/abs/2406.04329}.

\bibitem[Sahoo et~al.(2024)Sahoo, Arriola, Schiff, Gokaslan, Marroquin, Chiu, Rush, and Kuleshov]{Sahoo2024}
Subham~Sekhar Sahoo, Marianne Arriola, Yair Schiff, Aaron Gokaslan, Edgar Marroquin, Justin~T Chiu, Alexander Rush, and Volodymyr Kuleshov.
\newblock Simple and effective masked diffusion language models.
\newblock \emph{Advances in Neural Information Processing Systems}, 2024.
\newblock \doi{10.48550/ARXIV.2406.07524}.
\newblock URL \url{https://arxiv.org/abs/2406.07524}.

\bibitem[Gat et~al.(2024)Gat, Remez, Shaul, Kreuk, Chen, Synnaeve, Adi, and Lipman]{Gat2024}
Itai Gat, Tal Remez, Neta Shaul, Felix Kreuk, Ricky T.~Q. Chen, Gabriel Synnaeve, Yossi Adi, and Yaron Lipman.
\newblock Discrete flow matching.
\newblock \emph{Advances in Neural Information Processing Systems}, 2024.
\newblock \doi{10.48550/ARXIV.2407.15595}.
\newblock URL \url{https://arxiv.org/abs/2407.15595}.

\bibitem[Davis et~al.(2024)Davis, Kessler, Petrache, Ceylan, Bronstein, and Bose]{Davis2024}
Oscar Davis, Samuel Kessler, Mircea Petrache, Ismail~Ilkan Ceylan, Michael Bronstein, and Avishek~Joey Bose.
\newblock Fisher flow matching for generative modeling over discrete data.
\newblock \emph{Advances in Neural Information Processing Systems}, 2024.
\newblock \doi{10.48550/ARXIV.2405.14664}.
\newblock URL \url{https://arxiv.org/abs/2405.14664}.

\bibitem[Goel et~al.(2024)Goel, Thoutam, Marroquin, Gokaslan, Firouzbakht, Vincoff, Kuleshov, Kratochvil, and Chatterjee]{Goel2024}
Shrey Goel, Vishrut Thoutam, Edgar~Mariano Marroquin, Aaron Gokaslan, Arash Firouzbakht, Sophia Vincoff, Volodymyr Kuleshov, Huong~T. Kratochvil, and Pranam Chatterjee.
\newblock Memdlm: De novo membrane protein design with masked discrete diffusion protein language models.
\newblock \emph{arXiv}, 2024.
\newblock \doi{10.48550/ARXIV.2410.16735}.
\newblock URL \url{https://arxiv.org/abs/2410.16735}.

\bibitem[Jang et~al.(2017)Jang, Gu, and Poole]{Jang2016}
Eric Jang, Shixiang Gu, and Ben Poole.
\newblock Categorical reparameterization with gumbel-softmax.
\newblock \emph{International Conference on Learned Representations}, 2017.
\newblock \doi{10.48550/ARXIV.1611.01144}.
\newblock URL \url{https://arxiv.org/abs/1611.01144}.

\bibitem[Maddison et~al.(2016)Maddison, Mnih, and Teh]{Maddison2016}
Chris~J. Maddison, Andriy Mnih, and Yee~Whye Teh.
\newblock The concrete distribution: A continuous relaxation of discrete random variables, 2016.
\newblock URL \url{https://arxiv.org/abs/1611.00712}.

\bibitem[Peluchetti(2022)]{peluchetti2022nondenoising}
Stefano Peluchetti.
\newblock Non-denoising forward-time diffusions, 2022.
\newblock URL \url{https://openreview.net/forum?id=oVfIKuhqfC}.

\bibitem[Liu(2022)]{liu_rectified_2022}
Qiang Liu.
\newblock Rectified flow: A marginal preserving approach to optimal transport.
\newblock \emph{arXiv preprint 2209.14577}, 2022.

\bibitem[Albergo et~al.(2023)Albergo, Boffi, and {Vanden-Eijnden}]{albergo_stochastic_2023}
Michael~S. Albergo, Nicholas~M. Boffi, and Eric {Vanden-Eijnden}.
\newblock Stochastic interpolants: A unifying framework for flows and diffusions.
\newblock \emph{arXiv preprint 2303.08797}, 2023.

\bibitem[Lipman et~al.(2023)Lipman, Chen, Ben-Hamu, Nickel, and Le]{Lipman2022}
Yaron Lipman, Ricky T.~Q. Chen, Heli Ben-Hamu, Maximilian Nickel, and Matt Le.
\newblock Flow matching for generative modeling.
\newblock \emph{International Conference on Learning Representations}, 2023.
\newblock \doi{10.48550/ARXIV.2210.02747}.
\newblock URL \url{https://arxiv.org/abs/2210.02747}.

\bibitem[Tong et~al.(2024)Tong, Fatras, Malkin, Huguet, Zhang, Rector-Brooks, Wolf, and Bengio]{Optimal-Transport}
Alexander Tong, Kilian Fatras, Nikolay Malkin, Guillaume Huguet, Yanlei Zhang, Jarrid Rector-Brooks, Guy Wolf, and Yoshua Bengio.
\newblock Improving and generalizing flow-based generative models with minibatch optimal transport.
\newblock \emph{Transactions on Machine Learning Research}, 2024.
\newblock \doi{10.48550/ARXIV.2302.00482}.
\newblock URL \url{https://arxiv.org/abs/2302.00482}.

\bibitem[Song and Ermon(2019)]{Yang2020}
Yang Song and Stefano Ermon.
\newblock Generative modeling by estimating gradients of the data distribution.
\newblock \emph{Advances in Neural Information Processing Systems}, 2019.
\newblock \doi{10.48550/ARXIV.1907.05600}.
\newblock URL \url{https://arxiv.org/abs/1907.05600}.

\bibitem[Mahmood et~al.(2024)Mahmood, Oliva, and Styner]{Mahmood2024}
Ahsan Mahmood, Junier Oliva, and Martin~Andreas Styner.
\newblock Anomaly detection via gumbel noise score matching.
\newblock \emph{Frontiers in Artificial Intelligence}, 7, September 2024.
\newblock ISSN 2624-8212.
\newblock \doi{10.3389/frai.2024.1441205}.
\newblock URL \url{http://dx.doi.org/10.3389/frai.2024.1441205}.

\bibitem[Avdeyev et~al.(2023)Avdeyev, Shi, Tan, Dudnyk, and Zhou]{Avdeyev2023}
Pavel Avdeyev, Chenlai Shi, Yuhao Tan, Kseniia Dudnyk, and Jian Zhou.
\newblock Dirichlet diffusion score model for biological sequence generation, 2023.
\newblock URL \url{https://arxiv.org/abs/2305.10699}.

\bibitem[Chen et~al.(2022{\natexlab{a}})Chen, Wong, Troyanskaya, and Zhou]{Chen2022}
Kathleen~M. Chen, Aaron~K. Wong, Olga~G. Troyanskaya, and Jian Zhou.
\newblock A sequence-based global map of regulatory activity for deciphering human genetics.
\newblock \emph{Nature Genetics}, 54\penalty0 (7):\penalty0 940–949, July 2022{\natexlab{a}}.
\newblock ISSN 1546-1718.
\newblock \doi{10.1038/s41588-022-01102-2}.
\newblock URL \url{http://dx.doi.org/10.1038/s41588-022-01102-2}.

\bibitem[Chen et~al.(2022{\natexlab{b}})Chen, Zhang, and Hinton]{Chen2023}
Ting Chen, Ruixiang Zhang, and Geoffrey Hinton.
\newblock Analog bits: Generating discrete data using diffusion models with self-conditioning, 2022{\natexlab{b}}.
\newblock URL \url{https://arxiv.org/abs/2208.04202}.

\bibitem[Lin et~al.(2023{\natexlab{a}})Lin, Akin, Rao, Hie, Zhu, Lu, Smetanin, Verkuil, Kabeli, Shmueli, Dos Santos~Costa, Fazel-Zarandi, Sercu, Candido, and Rives]{Lin2023-gh}
Zeming Lin, Halil Akin, Roshan Rao, Brian Hie, Zhongkai Zhu, Wenting Lu, Nikita Smetanin, Robert Verkuil, Ori Kabeli, Yaniv Shmueli, Allan Dos Santos~Costa, Maryam Fazel-Zarandi, Tom Sercu, Salvatore Candido, and Alexander Rives.
\newblock Evolutionary-scale prediction of atomic-level protein structure with a language model.
\newblock \emph{Science}, 379\penalty0 (6637):\penalty0 1123--1130, March 2023{\natexlab{a}}.

\bibitem[Suzek et~al.(2007)Suzek, Huang, McGarvey, Mazumder, and Wu]{suzek2007uniref}
Baris~E Suzek, Hongzhan Huang, Peter McGarvey, Raja Mazumder, and Cathy~H Wu.
\newblock Uniref: comprehensive and non-redundant uniprot reference clusters.
\newblock \emph{Bioinformatics}, 23\penalty0 (10):\penalty0 1282--1288, 2007.

\bibitem[Cornman et~al.(2024)Cornman, West-Roberts, Camargo, Roux, Beracochea, Mirdita, Ovchinnikov, and Hwang]{Cornman2024}
Andre Cornman, Jacob West-Roberts, Antonio~Pedro Camargo, Simon Roux, Martin Beracochea, Milot Mirdita, Sergey Ovchinnikov, and Yunha Hwang.
\newblock The omg dataset: An open metagenomic corpus for mixed-modality genomic language modeling.
\newblock 2024.
\newblock \doi{10.1101/2024.08.14.607850}.
\newblock URL \url{https://www.biorxiv.org/content/early/2024/08/17/2024.08.14.607850}.

\bibitem[Steinegger and S{\"o}ding(2018)]{steinegger2018clustering}
Martin Steinegger and Johannes S{\"o}ding.
\newblock Clustering huge protein sequence sets in linear time.
\newblock \emph{Nature communications}, 9\penalty0 (1):\penalty0 2542, 2018.

\bibitem[Alamdari et~al.(2023)Alamdari, Thakkar, van~den Berg, Tenenholtz, Strome, Moses, Lu, Fusi, Amini, and Yang]{Alamdari2023}
Sarah Alamdari, Nitya Thakkar, Rianne van~den Berg, Neil Tenenholtz, Robert Strome, Alan~M. Moses, Alex~X. Lu, Nicolò Fusi, Ava~P. Amini, and Kevin~K. Yang.
\newblock Protein generation with evolutionary diffusion: sequence is all you need.
\newblock September 2023.
\newblock \doi{10.1101/2023.09.11.556673}.
\newblock URL \url{http://dx.doi.org/10.1101/2023.09.11.556673}.

\bibitem[Nijkamp et~al.(2023)Nijkamp, Ruffolo, Weinstein, Naik, and Madani]{Nijkamp2023}
Erik Nijkamp, Jeffrey~A. Ruffolo, Eli~N. Weinstein, Nikhil Naik, and Ali Madani.
\newblock Progen2: Exploring the boundaries of protein language models.
\newblock \emph{Cell Systems}, 14\penalty0 (11):\penalty0 968--978.e3, November 2023.
\newblock ISSN 2405-4712.
\newblock \doi{10.1016/j.cels.2023.10.002}.
\newblock URL \url{http://dx.doi.org/10.1016/j.cels.2023.10.002}.

\bibitem[Lin et~al.(2023{\natexlab{b}})Lin, Akin, Rao, Hie, Zhu, Lu, Smetanin, Verkuil, Kabeli, Shmueli, dos Santos~Costa, Fazel-Zarandi, Sercu, Candido, and Rives]{Lin2023}
Zeming Lin, Halil Akin, Roshan Rao, Brian Hie, Zhongkai Zhu, Wenting Lu, Nikita Smetanin, Robert Verkuil, Ori Kabeli, Yaniv Shmueli, Allan dos Santos~Costa, Maryam Fazel-Zarandi, Tom Sercu, Salvatore Candido, and Alexander Rives.
\newblock Evolutionary-scale prediction of atomic-level protein structure with a language model.
\newblock \emph{Science}, 379\penalty0 (6637):\penalty0 1123–1130, March 2023{\natexlab{b}}.
\newblock ISSN 1095-9203.
\newblock \doi{10.1126/science.ade2574}.
\newblock URL \url{http://dx.doi.org/10.1126/science.ade2574}.

\bibitem[Abdin et~al.(2022)Abdin, Nim, Wen, and Kim]{Abdin2022}
Osama Abdin, Satra Nim, Han Wen, and Philip~M. Kim.
\newblock Pepnn: a deep attention model for the identification of peptide binding sites.
\newblock \emph{Communications Biology}, 5\penalty0 (1), May 2022.
\newblock ISSN 2399-3642.
\newblock \doi{10.1038/s42003-022-03445-2}.
\newblock URL \url{http://dx.doi.org/10.1038/s42003-022-03445-2}.

\bibitem[Zhang et~al.(2023{\natexlab{a}})Zhang, Zhang, Freddolino, and Zhang]{BioLiP2}
Chengxin Zhang, Xi~Zhang, Lydia Freddolino, and Yang Zhang.
\newblock Biolip2: an updated structure database for biologically relevant ligand–protein interactions.
\newblock \emph{Nucleic Acids Research}, 52\penalty0 (D1):\penalty0 D404–D412, July 2023{\natexlab{a}}.
\newblock ISSN 1362-4962.
\newblock \doi{10.1093/nar/gkad630}.
\newblock URL \url{http://dx.doi.org/10.1093/nar/gkad630}.

\bibitem[Bushuiev et~al.(2023)Bushuiev, Bushuiev, Kouba, Filkin, Gabrielova, Gabriel, Sedlar, Pluskal, Damborsky, Mazurenko, and Sivic]{Bushuiev2023}
Anton Bushuiev, Roman Bushuiev, Petr Kouba, Anatolii Filkin, Marketa Gabrielova, Michal Gabriel, Jiri Sedlar, Tomas Pluskal, Jiri Damborsky, Stanislav Mazurenko, and Josef Sivic.
\newblock Learning to design protein-protein interactions with enhanced generalization, 2023.
\newblock URL \url{https://arxiv.org/abs/2310.18515}.

\bibitem[Campbell et~al.(2024)Campbell, Yim, Barzilay, Rainforth, and Jaakkola]{Campbell2024}
Andrew Campbell, Jason Yim, Regina Barzilay, Tom Rainforth, and Tommi Jaakkola.
\newblock Generative flows on discrete state-spaces: Enabling multimodal flows with applications to protein co-design.
\newblock \emph{arXiv}, 2024.
\newblock \doi{10.48550/ARXIV.2402.04997}.
\newblock URL \url{https://arxiv.org/abs/2402.04997}.

\bibitem[Lou et~al.(2024)Lou, Meng, and Ermon]{Lou2024}
Aaron Lou, Chenlin Meng, and Stefano Ermon.
\newblock Discrete diffusion modeling by estimating the ratios of the data distribution.
\newblock \emph{International Conference on Machine Learning}, 2024.
\newblock \doi{10.48550/ARXIV.2310.16834}.
\newblock URL \url{https://arxiv.org/abs/2310.16834}.

\bibitem[Campbell et~al.(2022)Campbell, Benton, Bortoli, Rainforth, Deligiannidis, and Doucet]{campbell_continuous_2022}
Andrew Campbell, Joe Benton, Valentin~De Bortoli, Tom Rainforth, George Deligiannidis, and Arnaud Doucet.
\newblock A {Continuous} {Time} {Framework} for {Discrete} {Denoising} {Models}.
\newblock October 2022.
\newblock URL \url{https://openreview.net/forum?id=DmT862YAieY}.

\bibitem[Pooladian et~al.(2023)Pooladian, Ben-Hamu, Domingo-Enrich, Amos, Lipman, and Chen]{Pooladian2023}
Aram-Alexandre Pooladian, Heli Ben-Hamu, Carles Domingo-Enrich, Brandon Amos, Yaron Lipman, and Ricky T.~Q. Chen.
\newblock Multisample flow matching: Straightening flows with minibatch couplings.
\newblock \emph{International Conference on Machine Learning}, 2023.
\newblock \doi{10.48550/ARXIV.2304.14772}.
\newblock URL \url{https://arxiv.org/abs/2304.14772}.

\bibitem[Zhang et~al.(2024)Zhang, Pu, Kawamura, Loza, Bengio, Shung, and Tong]{Trajectory-Flow}
Xi~Zhang, Yuan Pu, Yuki Kawamura, Andrew Loza, Yoshua Bengio, Dennis~L. Shung, and Alexander Tong.
\newblock Trajectory flow matching with applications to clinical time series modeling, 2024.
\newblock URL \url{https://arxiv.org/abs/2410.21154}.

\bibitem[Zheng et~al.(2023)Zheng, Le, Shaul, Lipman, Grover, and Chen]{Zheng2023}
Qinqing Zheng, Matt Le, Neta Shaul, Yaron Lipman, Aditya Grover, and Ricky T.~Q. Chen.
\newblock Guided flows for generative modeling and decision making, 2023.
\newblock URL \url{https://arxiv.org/abs/2311.13443}.

\bibitem[Ho and Salimans(2022)]{HoandSaliman2022}
Jonathan Ho and Tim Salimans.
\newblock Classifier-free diffusion guidance.
\newblock \emph{NeurIPS 2021 Workshop on Deep Generative Models and Downstream Applications}, 2022.
\newblock \doi{10.48550/ARXIV.2207.12598}.
\newblock URL \url{https://arxiv.org/abs/2207.12598}.

\bibitem[Song et~al.(2021)Song, Sohl-Dickstein, Kingma, Kumar, Ermon, and Poole]{Song2021}
Yang Song, Jascha Sohl-Dickstein, Diederik~P. Kingma, Abhishek Kumar, Stefano Ermon, and Ben Poole.
\newblock Score-based generative modeling through stochastic differential equations.
\newblock \emph{International Conference on Learning Representations}, 2021.
\newblock \doi{10.48550/ARXIV.2011.13456}.
\newblock URL \url{https://arxiv.org/abs/2011.13456}.

\bibitem[Peebles and Xie(2023)]{DiT}
William Peebles and Saining Xie.
\newblock Scalable diffusion models with transformers.
\newblock \emph{IEEE/CVF International Conference on Computer Vision (ICCV)}, 2023.
\newblock \doi{10.48550/ARXIV.2212.09748}.
\newblock URL \url{https://arxiv.org/abs/2212.09748}.

\bibitem[Su et~al.(2021)Su, Lu, Pan, Murtadha, Wen, and Liu]{RoPE}
Jianlin Su, Yu~Lu, Shengfeng Pan, Ahmed Murtadha, Bo~Wen, and Yunfeng Liu.
\newblock Roformer: Enhanced transformer with rotary position embedding, 2021.
\newblock URL \url{https://arxiv.org/abs/2104.09864}.

\bibitem[Zhang et~al.(2023{\natexlab{b}})Zhang, Wu, Xiu, Li, Chen, Wang, Wang, Gao, and Zhou]{https://doi.org/10.48550/arxiv.2311.04419}
Ruochi Zhang, Haoran Wu, Yuting Xiu, Kewei Li, Ningning Chen, Yu~Wang, Yan Wang, Xin Gao, and Fengfeng Zhou.
\newblock Pepland: a large-scale pre-trained peptide representation model for a comprehensive landscape of both canonical and non-canonical amino acids.
\newblock \emph{arXiv}, 2023{\natexlab{b}}.
\newblock \doi{10.48550/ARXIV.2311.04419}.
\newblock URL \url{https://arxiv.org/abs/2311.04419}.

\bibitem[Akiba et~al.(2019)Akiba, Sano, Yanase, Ohta, and Koyama]{akiba2019optuna}
Takuya Akiba, Shotaro Sano, Toshihiko Yanase, Takeru Ohta, and Masanori Koyama.
\newblock Optuna: A next-generation hyperparameter optimization framework.
\newblock In \emph{Proceedings of the 25th ACM SIGKDD international conference on knowledge discovery \& data mining}, pages 2623--2631, 2019.

\bibitem[Abramson et~al.(2024)Abramson, Adler, Dunger, Evans, Green, Pritzel, Ronneberger, Willmore, Ballard, Bambrick, Bodenstein, Evans, Hung, O’Neill, Reiman, Tunyasuvunakool, Wu, Žemgulytė, Arvaniti, Beattie, Bertolli, Bridgland, Cherepanov, Congreve, Cowen-Rivers, Cowie, Figurnov, Fuchs, Gladman, Jain, Khan, Low, Perlin, Potapenko, Savy, Singh, Stecula, Thillaisundaram, Tong, Yakneen, Zhong, Zielinski, Žídek, Bapst, Kohli, Jaderberg, Hassabis, and Jumper]{Abramson2024}
Josh Abramson, Jonas Adler, Jack Dunger, Richard Evans, Tim Green, Alexander Pritzel, Olaf Ronneberger, Lindsay Willmore, Andrew~J. Ballard, Joshua Bambrick, Sebastian~W. Bodenstein, David~A. Evans, Chia-Chun Hung, Michael O’Neill, David Reiman, Kathryn Tunyasuvunakool, Zachary Wu, Akvilė Žemgulytė, Eirini Arvaniti, Charles Beattie, Ottavia Bertolli, Alex Bridgland, Alexey Cherepanov, Miles Congreve, Alexander~I. Cowen-Rivers, Andrew Cowie, Michael Figurnov, Fabian~B. Fuchs, Hannah Gladman, Rishub Jain, Yousuf~A. Khan, Caroline M.~R. Low, Kuba Perlin, Anna Potapenko, Pascal Savy, Sukhdeep Singh, Adrian Stecula, Ashok Thillaisundaram, Catherine Tong, Sergei Yakneen, Ellen~D. Zhong, Michal Zielinski, Augustin Žídek, Victor Bapst, Pushmeet Kohli, Max Jaderberg, Demis Hassabis, and John~M. Jumper.
\newblock Accurate structure prediction of biomolecular interactions with alphafold 3.
\newblock \emph{Nature}, 630\penalty0 (8016):\penalty0 493–500, May 2024.
\newblock ISSN 1476-4687.
\newblock \doi{10.1038/s41586-024-07487-w}.
\newblock URL \url{http://dx.doi.org/10.1038/s41586-024-07487-w}.

\bibitem[Eberhardt et~al.(2021)Eberhardt, Santos-Martins, Tillack, and Forli]{eberhardt2021autodock}
Jerome Eberhardt, Diogo Santos-Martins, Andreas~F Tillack, and Stefano Forli.
\newblock Autodock vina 1.2. 0: New docking methods, expanded force field, and python bindings.
\newblock \emph{Journal of chemical information and modeling}, 61\penalty0 (8):\penalty0 3891--3898, 2021.

\bibitem[{Schr\"odinger, LLC}(2015)]{PyMOL}
{Schr\"odinger, LLC}.
\newblock The {PyMOL} molecular graphics system, version~1.8.
\newblock November 2015.

\end{thebibliography}
%%%%%%%%%%%%%%%%%%%%%%%%%%%%%%%%%%%%%%%%%%%%%%%%%%%%%%%%%%%%
%%%%%%%%%%%%%%%%%%%%%%%%%%%%%%%%%%%%%%%%%%%%%%%%%%%%%%%%%%%%
%%%%%%%%%%%%%%%%%%%%%%%%%%%%%%%%%%%%%%%%%%%%%%%%%%%%%%%%%%%%
%%%%%%%%%%%%%%%%%%%%%%%%%%%%%%%%%%%%%%%%%%%%%%%%%%%%%%%%%%%%
\appendix
\newpage
\section{Extended Background}
\subsection{Flow Matching on the Simplex}
Here, we discuss the motivation behind discrete flow matching \cite{Campbell2024, Gat2024}, and specifically on the interior of the simplex \cite{Stark2024, Davis2024}. This discussion will help motivate the contribution of our work from past iterations. 

Discrete diffusion models \cite{Austin2021, Lou2024, campbell_continuous_2022} operate by applying categorical noise in the form of $\mathbf{x}_t\sim \text{Cat}(\cdot |\mathbf{Q}_t^{\top }\mathbf{x}_0)$ that convert the clean sequence of one-hot categorical distributions $\mathbf{x}_0$ to a noisy sequence $\mathbf{z}_t$. Then, a parameterized model learns to iteratively reconstruct the clean sequence $\mathbf{x}_0$ from the noisy sequence $\mathbf{z}_t$ by taking $t$ discrete backward transitions given by $\mathbf{z}_s\sim \text{Cat}\left(\cdot |\frac{\mathbf{Q}_{s|t}\mathbf{z}_t\odot \mathbf{Q}_s^{\top}\mathbf{x}_{\theta}(\mathbf{z}_t, t) }{\mathbf{z}_t^{\top}\mathbf{Q}_t^{\top}\mathbf{x}_{\theta}(\mathbf{z}_t, t)}\right)$. However, this method operates in the fully \textit{discrete} state space, meaning that the noisy sequence at each time step is a fully discrete sequence of one-hot vectors sampled from continuous categorical distributions. This can result in discretization errors during sampling when abruptly restricting continuous distributions to a single token. This presents the question: \textit{Can we generate discrete sequences by iteratively fine-tuning continuous probability distributions?}

This is the motivation behind discrete flow matching models on the simplex \cite{Stark2024, Davis2024}, which defines a smooth interpolation $\psi_t(\mathbf{x}_1)$ from a prior uniform distribution over the simplex $\mathbf{x}_0$ to a unitary distribution concentrated at a single vertex $\mathbf{x}_1$ over the time interval $t\in[0, 1]$. To ensure that noisy can be transformed into valid clean sequences at inference, the interpolant must satisfy the boundary conditions given by $\psi_0(\mathbf{x}_1)\approx\frac{\mathbf{1}}{V}$ where $V$ is the size of the token vocabulary. The advantage of this approach over fully discrete methods is the ability to refine probability distributions given the neighboring distributions rather than noisy discrete tokens that accumulate discretization errors at each time step. 

\subsection{Deterministic vs. Stochastic Interpolants}
The linear interpolant \cite{Lipman2022, Pooladian2023} defines a a deterministic flow $\psi_t(\mathbf{x}_t|\mathbf{x}_0, \mathbf{x}_1)=t\mathbf{x}_0+(1-t)\mathbf{x}_1$ between a pair of fixed endpoints $(\mathbf{x}_0, \mathbf{x}_1)$. Optimal transport \cite{Optimal-Transport} further defines an optimal mapping $\pi(\mathbf{x}_0, \mathbf{x}_1)$ that minimizes a cost function $c(\mathbf{x}_0, \mathbf{x}_1)$— often a squared distance cost $c(\mathbf{x}_0, \mathbf{x}_1)=d^2(\mathbf{x}_0, \mathbf{x}_1)$—between paired endpoints. Although the deterministic perspective is optimal for tasks like matching trajectories \cite{Trajectory-Flow}, it lacks expressivity and diversity for \textit{de novo} design tasks like protein or peptide-binder design. This approach also prevents the flow model from effectively learning to \textit{redirect} specific token trajectories that do not reflect the data distribution during inference given the sequence context. 

By defining a \textit{stochastic interpolant} with Gumbel-noise where each token has a small probability of being transformed into a distribution where the token with the highest probability does not match the true token during training, the model still needs to predict the clean distribution $\mathbf{x}_{\theta}(\mathbf{x}_t, t)$ or the target generating velocity field $u^{\theta}_t(\mathbf{x}_t)$ but with more ambiguity given that not all distributions are on the deterministically biased towards the target token. This pushes the model to place a greater weight on the global context of each token and learn dependencies across tokens to generate a valid clean sequence despite the increased ambiguity. Furthermore, this approach injects path variability to improve generalization and exploration of diverse flows for \textit{de novo} design tasks. 

\subsection{Guided Flow Matching}
A key limitation of current discrete flow matching techniques is the lack of training-free guidance strategies. Flow matching guidance \cite{Zheng2023, HoandSaliman2022} is performed either with \textit{classifier-based} or \textit{classifier-free} guidance. 

\textbf{Classifier-Free Guidance.} In classifier-free guided flow matching \cite{Zheng2023}, the guided velocity field is obtained by training a guided flow model $u^{\phi}_t(\mathbf{x}|y)$ and an unconditional flow model $u^{\theta}_t(\mathbf{x})$ and taking the linear combination of the guided and unconditional velocities scaled by a parameter $\gamma$.
\begin{align}
    \tilde{u}^{\theta}_t(\mathbf{x}| y)=(1-\gamma)u^{\theta}_t(\mathbf{x})+\gamma u^{\theta}_t(\mathbf{x}|y)
\end{align}
This strategy requires training an additional guided flow model on quality-labeled data, which is often scarce. Given that flow models require more training data than simple regression and classification models, classifier-based guidance is preferred for scalability. 

\textbf{Classifier-Based Guidance.} In classifier-based guided flow matching \cite{Song2021}, a \textit{time-dependent} classifier $p_t^{\phi}(y |\mathbf{x}_t)$ that predicts a classifier score given noisy samples $\mathbf{x}_t$ separately from the unconditional generator. Then, we sample with a guided velocity field given by
\begin{align}
    u^{\theta, \phi}_t (\mathbf{x}_t)=u^{\theta}_t (\mathbf{x}_t)+\gamma\nabla_{\mathbf{x}_t}\log p_t^{\phi}(y |\mathbf{x}_t)
\end{align}
which requires projection back to the simplex for guided discrete flows. For simplex-based flows, this approach typically involves additional training of \textit{noisy classifiers} that predict the classifier score given intermediate distributions over the simplex at each time step. Not only are these noisy classifiers less accurate than large pre-trained classifiers on clean sequences, but they also require extensive training as all noise levels need to be included in the training task. 

STGFlow overcomes these limitations by defining a guided flow velocity using the straight-through gradients of the scoring model on discrete sequences sampled with respect to the relaxed Gumbel-softmax probabilities. To ensure that the scores of sampled sequences are representative of the relaxed distribution, we sample $M$ sequences and take the aggregate gradient as the guided velocity. This provides a modular training-free strategy for discrete flow matching guidance that conserves the probability mass constraint (Proof in Appendix \ref{appendix:Probability Mass Conservation of Straight-Through Gradient}).  

\section{Relation to Prior Simplex-Based Flow Matching Models}
In this section, we discuss and compare Gumbel-Softmax FM with two related methods for discrete flow matching on the simplex: \textbf{Dirichlet Flow Matching} \cite{Stark2024} and \textbf{Fisher Flow Matching} \cite{Davis2024}. 

\subsection{Dirichlet Flow Matching}
The Dirichlet distribution is an extension of the Beta distribution $\mathcal{B}$ for multiple variables and models the probability of the next variable $x$ being in one of $V$ discrete categories given a parameter vector $\vec{\alpha}=(\alpha_1, \dots, \alpha_V)$. Intuitively, it acts as a distribution of smooth categorical vectors $\mathbf{x}\in \Delta ^{V-1}$ that lie on the probability simplex given that each category $i\in [1\dots V]$ was observed with \textit{frequency} $\alpha_i$. Increasing $\alpha_i$ for a given category $i$ would increase the probability of sampling $\mathbf{x}$ near the $i$th vertex of the simplex. Dirichlet FM \cite{Stark2024} defines the conditional probability path as
\begin{align}
    p_t(\mathbf{x}|\mathbf{x}_1=\mathbf{e}_k)=\text{Dir}(\mathbf{x};\vec{\alpha}=\mathbf{1}+t\cdot \mathbf{e}_k)= \frac{1}{\mathcal{B}(\alpha_1, \dots, \alpha_V)}\prod_{i=1}^Vx_{t,i}^{\alpha_i-1}
\end{align}
At $t=0$, the distribution reduces to a uniform prior over $\Delta^{V-1}$, with an equal probability of sampling $\mathbf{x}$ near any vertex. As $t\to \infty$, $\alpha_k$ increases while $\alpha_j$ for all $j\neq k$ remain constant, so the probability density converges to the $k$th vertex. As shown in \cite{Stark2024}, this distribution satisfies the boundary constraints. 

To compute the target vector field, we start with the following equation
\begin{align}
    u_t(\mathbf{x}_t|\mathbf{x}_1=\mathbf{e}_k)=-\tilde{I}_{x_{t, k}(t+1, V-1)}\frac{\mathcal{B}(t+1, V-1)}{(1-x_{t, k})^{V-1}\cdot x_{t, k}}(\mathbf{e}_k-\mathbf{x}_t) 
\end{align}
Similar to our approach, Dirichlet FM trains a denoising model by minimizing a negative log loss and computes the velocity field as the linear combination of the conditional velocity fields as in Equation \ref{eq:Linear Combination Velocity Field}. 

Although the Dirichlet probability path provides support over the entire simplex at all time steps, it suffers from high variance during training due to the stochastic nature of sampling from the Dirichlet distribution. Since flow matching learns a mixture of conditional velocity fields, there exists inherent variability during inference. Our definition of a Gumbel-Softmax interpolant ensures straighter flow paths and lower variance during training as Gumbel noise largely preserves the relative probabilities between categories. 

\subsection{Fisher Flow Matching}
Fisher FM \cite{Davis2024} overcomes the instability of the Fisher-Rao metric at the vertices of the simplex via a sphere map $\varphi: {\Delta}^{V-1}\to \mathbb{S}^{V-1}_+$ where $\varphi(x)=\sqrt{x}$ that maps a point in the interior of the $(V-1)$-dimensional simplex to a point on the positive orthant of the $(V-1)$-dimensional hypersphere. The conditional velocity field $u_t(\mathbf{x}_t|\mathbf{x}_1)$ of the linear interpolant on the sphere is given by
\begin{align}
    \psi_t (\mathbf{x}_1)&=\exp_{\mathbf{x}_0}\left(t\log_{\mathbf{x}_0}(\mathbf{x}_1)\right)\nonumber\\
    u_t(\mathbf{x}_t|\mathbf{x}_1)&=\frac{\log_{\mathbf{x}_t}(\mathbf{x}_1)}{1-t}
\end{align}
During inference, the parameterized velocity field $\tilde{u}_{\theta}(\mathbf{x}_t)\in \mathbb{R}^V$ is projected onto the tangent bundle of the hypersphere $\mathcal{T}_{\mathbf{x}_t}\mathbb{S}_+^V$ via the following mapping
\begin{align}
   u^{\theta}_t(\mathbf{x}_t)= \tilde{u}_{\theta}(\mathbf{x}_t)-\langle \mathbf{x}_t, \tilde{u}_{\theta}(\mathbf{x}_t)\rangle_2\mathbf{x}_t
\end{align}
which minimizes the mean-squared error with the true conditional velocity field given by
\begin{align}
    \mathcal{L}_{\text{fisher}}=\mathbb{E}_{t\sim \mathcal{U}(0, 1), p_t(\mathbf{x}_t| \mathbf{x}_1), p_1(\mathbf{x}_1)}\left\|u_{\theta}(t, \mathbf{x}_t)-\frac{\log_{\mathbf{x}_t}(\mathbf{x}_1)}{1-t}\right\|^2_{\mathbb{S}_+^V}
\end{align}
%Given that the interior of the simplex $\Delta^{V-1}$ and the hypersphere $\mathbb{S}_+^{V-1}$ are isometric, the Optimal-Transport (OT) map $\pi(\mathbf{x}_0, \mathbf{x}_1)$ is equal under both manifolds. This enables training with optimal-transport via the Sinkhorn algorithm.

Fisher FM addresses the high training variance of Dirichlet FM without the pathological properties of linear flows on the simplex by projecting the linear interpolant to the positive orthant of the $V$-dimensional hypersphere, which is isometric to the $(V-1)$-dimensional simplex. However, projecting velocity fields to and from the tangent space of the hypersphere can lead to inconsistencies when applying guidance methods. Empirically, we found that the Fisher FM exhibits significantly high validation MSE loss during training, especially for increasing simplex dimensions, suggesting that the parameterization easily overfits to training data and is not optimal for \textit{de novo} design tasks such as protein generation or peptide design. 

\section{Flow Matching Derivations}

\subsection{Deriving the Conditional Velocity Field}\label{appendix:Deriving the Conditional Velocity Field}
We derive the conditional velocity field at a point $\mathbf{x}_t$ denoted as $u_t(\mathbf{x}|\mathbf{x}_1=\mathbf{e}_i)$ by taking the derivative of the interpolant $\psi_t(\mathbf{x}_1=\mathbf{e}_i)$ with respect to time $t$.
\begin{align}
    u_{t, i}(\mathbf{x}_t|\mathbf{x}_1=\mathbf{e}_k)&=\frac{d}{dt}\psi_{t, i}(\mathbf{x}_0|\mathbf{x}_1=\mathbf{e}_k)\nonumber\\
    &=\frac{d}{dt}\frac{\exp\left(\frac{\log \pi_i+g_i}{\tau_{\max}\exp(-\lambda t)}\right)}{\sum _{j=1}^V\exp\left(\frac{\log \pi_j+g_j}{\tau_{\max}\exp(-\lambda t)}\right)} 
\end{align}
Letting $z_i=\exp\left(\frac{\log \pi_i+g_i}{\tau_{\max}\exp(-\lambda t)}\right)$, we have
\begin{align}
    u_t(\mathbf{x}_t|\mathbf{x}_1=\mathbf{e}_k)&=\frac{d}{dt}\frac{\exp(z_i)}{\sum _{j=1}^V\exp(z_j)}\nonumber\\
    &=\frac{\left(\frac{d}{dt}\exp(z_i)\right)\left(\sum _{j=1}^V\exp(z_j)\right)-\exp(z_i)\left(\frac{d}{dt}\sum_{j=1}^V\exp(z_j)\right)}{\left(\sum_{j=1}^V\exp(z_j)\right)^2}
\end{align}
First, we compute $\frac{d}{dt}\exp(z_i)$
\begin{align}
    \frac{d}{dt}\exp\left(\frac{\log\pi_i+g_i}{\tau_{\max}\exp(-\lambda t)}\right)&=\exp(z_i)\cdot \frac{d}{dt}\left(\frac{\log\pi_i+g_i}{\tau_{\max}\exp(-\lambda t)}\right)\nonumber\\
    &=\exp(z_i)\cdot\frac{\log\pi_i+g_i}{\tau_{\max}} \cdot \frac{d}{dt}\exp(\lambda t)\nonumber\\
    &=\exp(z_i)\cdot\frac{\log\pi_i+g_i}{\tau_{\max}} \cdot \lambda\exp(\lambda t)\nonumber\\
\end{align}
Then, we compute $\frac{d}{dt}\sum _j\exp(z_j)$
\begin{align}
    \frac{d}{dt}\sum_{j=1}^V\exp\left(\frac{\log\pi_j+g_j}{\tau_{\max}\exp(-\lambda t)}\right)&=\sum_{j=1}^V\frac{d}{dt}\exp\left(\frac{\log\pi_j+g_j}{\tau_{\max}\exp(-\lambda t)}\right)\nonumber\\
    &=\sum_{j=1}^V\left(\exp(z_j)\cdot\frac{\log\pi_j+g_j}{\tau_{\max}} \cdot \lambda\exp(\lambda t)\right)
\end{align}
Then, substituting these terms back into the expression for $u_t$, we get
\begin{small}
\begin{align}
    &u_{t, i}(\mathbf{x}_t|\mathbf{x}_1=\mathbf{e}_k)\nonumber\\
    &=\frac{\left(\sum _{j=1}^V\exp\left(z_j\right)\right)\cdot\exp(z_i)\cdot\frac{\log\pi_i+g_i}{\tau_{\max}} \cdot \lambda\exp(\lambda t) -\exp\left(z_i\right)\cdot \sum_{j=1}^V\left(\exp(z_j)\cdot\frac{\log\pi_j+g_j}{\tau_{\max}} \cdot \lambda\exp(\lambda t)\right)}{\left(\sum _{j=1}^V\exp\left(z_j\right)\right)^2}\nonumber\\
    &=\frac{\exp(z_i)\cdot \lambda \exp(\lambda t)}{\tau_{\text{max}}\left(\sum _{j=1}^V\exp\left(z_j\right)\right)^2} \left[\left(\sum_{j=1}^V \exp(z_j )\right)\cdot (\log \pi_i+g_i)-\sum_{j=1}^V\bigg(\exp(z_j)\cdot(\log \pi_j+g_j) \bigg)\right]\nonumber\\
    &=\frac{\exp(z_i)\cdot  \lambda\exp(\lambda t)}{\tau_{\text{max}}\left(\sum_{j=1}^V\exp\left(z_j\right)\right)^2} \left[\sum_{j=1}^V\exp(z_j) \bigg((\log \pi_i+g_i)-(\log \pi_j+g_j)\bigg)\right]\nonumber\\
    &=\frac{\exp(z_i)}{\sum _{j=1}^V\exp\left(z_j\right)}\frac{\lambda\exp(\lambda t)}{\tau_{\text{max}}} \left[\sum_{j=1}^V\bigg(\frac{\exp(z_j )}{\sum _{j'}\exp\left(z_j\right)}\cdot \bigg((\log \pi_i+g_i)-(\log \pi_j+g_j)\bigg)\bigg)\right]\nonumber\\
    &=\psi_{t, i}(\mathbf{x}_1)\cdot \frac{\lambda \exp(\lambda t)}{\tau_{\text{max}}} \left[\sum_{j=1}^V\bigg(\psi_{t, j}(\mathbf{x}_1)\cdot \bigg((\log \pi_i+g_i)-(\log \pi_j+g_j)\bigg)\bigg)\right]\nonumber\\
    &=\frac{ \lambda}{\tau(t)} x_{t, i}\sum_{j=1}^Vx_{t, j}\cdot \bigg((\log \pi_i+g_i)-(\log \pi_j+g_j)\bigg)
\end{align}
\end{small}

By our definition of the Gumbel-Softmax interpolant, the intermediate distributions during inference represent a mixture of learned conditional interpolants $\psi _t(\mathbf{x}_1)$ from the training data. Since the denoising model is trained to predict the true clean distribution, we can set the Gumbel-noise random variable in the conditional velocity fields to 0 during inference as we want the velocity field to point toward the predicted denoised distribution.

Substituting in $\pi_i=\exp(\delta_{ik})$, we have
\begin{align}
    u_{t, i}(\mathbf{x}_t|\mathbf{x}_1=\mathbf{e}_k)&=\frac{ \lambda}{\tau(t)} x_{t, i}\sum_{j=1}^Vx_{t, j}\cdot (\delta_{ik}-\delta_{jk})\nonumber
\end{align}
Since $\delta_{ij}=1$ only when $i$ is the index of the target token $i=k$ and 0 otherwise, the velocity field can be rewritten as
\begin{small}
\begin{align}
    u_{t, i}(\mathbf{x}_0|\mathbf{x}_1=\mathbf{e}_k)&=\begin{cases}
        \frac{ \lambda}{\tau(t)} x_{t, i}\sum_{j=1}^V\big(x_{t, j}\cdot (1-\delta_{jk})\big)&i=k\\
        \frac{\lambda}{\tau(t)}x_{t, i} \sum_{j=1}^V\big(x_{t, j}\cdot (-\delta_{jk})\big)&i\neq k
    \end{cases}\nonumber\\
    &=\begin{cases}
        \frac{ \lambda \exp(\lambda t)}{\tau_{\text{max}}}x_{t, i}\left(\sum_{j=1}^Vx_{t, j}-\sum_{j=1}^V x _{t, j}\delta_{jk}\right)&i=k\\
        \frac{\lambda \exp(\lambda t)}{\tau_{\text{max}}}x_{t, i}\left(-\sum_{j=1}^V x _{t, j}\delta_{jk}\right)&i\neq k
    \end{cases}\nonumber\\
    &=\begin{cases}
        \frac{ \lambda \exp(\lambda t)}{\tau_{\text{max}}}x_{t, i}\left(1-x _{t, k}\right)&i=k\\
        \frac{\lambda \exp(\lambda t)}{\tau_{\text{max}}}x_{t, i}\left(-x_{t, k}\right)&i\neq k
    \end{cases}
\end{align}
\end{small}
Rewriting in vector form, we get 
\begin{align}
    u_{t}(\mathbf{x}_t|\mathbf{x}_1=\mathbf{e}_k)&=\frac{\lambda}{\tau(t)}x_{t, k}\left(\mathbf{e}_k-\mathbf{x}_t\right)\label{appendix-eq:velocity field}
\end{align}
which points toward the target vertex $\mathbf{e}_k$. 

\subsection{Proof of Continuity}\label{appendix:Continuity Equation}
\textbf{Proposition 1.} The proposed conditional vector field and conditional probability path satisfy the continuity equation and thus define a valid flow-matching trajectory in the interior of the simplex.
\begin{align}
    \frac{\partial}{\partial t}p_t(\mathbf{x})=-\nabla\cdot(p_t(\mathbf{x})u_t(\mathbf{x}_t))
\end{align}

\textit{Proof of Proposition 1.} During training, each clean sequence $\mathbf{x}_1$ is transformed into some noisy interpolant $\psi_t(\mathbf{x}_t)$ with a sampled Gumbel-noise vector $\mathbf{g}\sim \text{Gumbel}(0,1)$. Therefore, we can rewrite the interpolant as a deterministic path conditioned on the one-hot distribution $\mathbf{x}_1$ and Gumbel-noise vector $\mathbf{g}$
\begin{align}
    \psi_{t}(\mathbf{x}_1)=\psi_{t}(\mathbf{x}_1, \mathbf{g})=\text{SM}\left(\frac{\mathbf{x}_1+\mathbf{g}}{\tau (t)}\right)
\end{align}
With this definition, we can define a deterministic probability path as the Dirac delta function along the interpolant $\mathbf{x}_t=\psi_t(\mathbf{x}_1)$ as
\begin{align}
    p_t(\mathbf{x}|\mathbf{x}_1)=\delta(\mathbf{x}-\psi _t(\mathbf{x}_1))
\end{align}
So, we can rewrite the continuity equation as
\begin{align}
     \frac{\partial}{\partial t}p_t(\mathbf{x}|\mathbf{x}_1)&=-\nabla\cdot\left(\delta(\mathbf{x}-\psi _t(\mathbf{x}_1))\frac{\partial}{\partial t}\psi_t(\mathbf{x}_1)\right)\nonumber\\
     &=-\nabla \delta(\mathbf{x}-\psi _t(\mathbf{x}_1))\cdot\frac{\partial}{\partial t}\psi_t(\mathbf{x}_1)
\end{align}

First, we will simplify the right-hand side (RHS) of the continuity equation. Taking the derivative with respect to $t$, we get
\begin{align}
    \frac{\partial}{\partial t}p_t(\mathbf{x}|\mathbf{x}_1)&=\frac{\partial}{\partial t}\delta(\mathbf{x}-\psi _t(\mathbf{x}_t))\nonumber
\end{align}
Taking the distributional derivative with an arbitrary test function $f(\mathbf{x})$ independent of $t$, we have
\begin{align}
    &=\int f(\mathbf{x})\frac{\partial}{\partial t}\delta(\mathbf{x}-\psi _t(\mathbf{x}_t, \mathbf{g}))d\mathbf{x}\nonumber\\
    &= \frac{\partial}{\partial t}\int f(\mathbf{x})\delta(\mathbf{x}-\psi _t(\mathbf{x}_t))d\mathbf{x}\nonumber\\
    &=\frac{\partial}{\partial t} f(\psi _t(\mathbf{x}_1))
\end{align}
Since $\psi _t(\mathbf{x}_1)\in \mathbb{R}^V$, we apply the multivariable chain rule to get
\begin{align}
    \frac{\partial}{\partial t}p_t(\mathbf{x}|\mathbf{x}_1)&=\nabla f(\psi _t(\mathbf{x}_1))\cdot \frac{\partial}{\partial t}\psi_t(\mathbf{x}_1)
\end{align}
Now, we integrate the left-hand side (LHS) of the continuity equation with an arbitrary test function. 
\begin{align}
    \int f(\mathbf{x})\left[-\nabla \delta(\mathbf{x}-\psi _t(\mathbf{x}_1))\cdot\frac{\partial}{\partial t}\psi_t(\mathbf{x}_1)\right]d\mathbf{x}&=-\left[\int f(\mathbf{x})\nabla \delta(\mathbf{x}-\psi _t(\mathbf{x}_1))d\mathbf{x}\right]\cdot \frac{\partial}{\partial t}\psi_t(\mathbf{x}_1)
\end{align}
Using integration by parts, we can write the term inside the bracket as
\begin{align}
    \int_{-\infty}^{\infty} f(\mathbf{x})\nabla \delta(\mathbf{x}-\psi _t(\mathbf{x}_1))d\mathbf{x}&=\underbrace{f(\mathbf{x})\delta(\mathbf{x}-\psi _t(\mathbf{x}_1))\bigg\vert_{-\infty}^{\infty}}_{=0}-\int_{-\infty}^{\infty} \delta(\mathbf{x}-\psi _t(\mathbf{x}_1))\nabla f(\mathbf{x})\nonumber\\
    &=-\int_{-\infty}^{\infty} \delta(\mathbf{x}-\psi _t(\mathbf{x}_1))\nabla f(\mathbf{x})
\end{align}
Substituting this back into the LHS, we get
\begin{align}
    \int f(\mathbf{x})\left[-\nabla \delta(\mathbf{x}-\psi _t(\mathbf{x}_1))\cdot\frac{\partial}{\partial t}\psi_t(\mathbf{x}_)\right]d\mathbf{x}&=-\left[-\int_{-\infty}^{\infty} \delta(\mathbf{x}-\psi _t(\mathbf{x}_1))\nabla f(\mathbf{x})\right]\cdot \frac{\partial}{\partial t}\psi_t(\mathbf{x}_1)\nonumber\\
    &=\nabla f(\psi_t(\mathbf{x}_1))\cdot \frac{\partial}{\partial t}\psi_t(\mathbf{x}_1)
\end{align}
We have shown that both sides of the continuity equation produce the same expression when integrated against any arbitrary test function $f(\mathbf{x})$. So, we can conclude
\begin{align}
    \frac{\partial}{\partial t}p_t(\mathbf{x}|\mathbf{x}_1)&=-\nabla \delta(\mathbf{x}-\psi _t(\mathbf{x}_1))\cdot\frac{\partial}{\partial t}\psi_t(\mathbf{x}_1)\nonumber\\
    &=\nabla \cdot (p_t(\mathbf{x}|\mathbf{x}_1)u_t(\mathbf{x}_t|\mathbf{x}_1))
\end{align}

Now that we have shown the continuity equation holds for the conditional probability density and flow velocities, it follows that the continuity equation holds for the unconditional flow. Following the proof in \cite{Optimal-Transport}, we have
\begin{small}
\begin{align}
    \frac{d}{dt}p_t(\mathbf{x}_t)&=\frac{d}{dt}\int_{\mathbf{x}_1}p_t(\mathbf{x}_t|\mathbf{x}_1)p_1(\mathbf{x}_1)d\mathbf{x}_1\nonumber\\
    &=\int_{\mathbf{x}_1}\frac{d}{dt}p_t(\mathbf{x}_t|\mathbf{x}_1)p_1(\mathbf{x}_1)d\mathbf{x}_1\nonumber\\
    &=\int_{\mathbf{x}_1}-\nabla \cdot \bigg(p_t(\mathbf{x}_t|\mathbf{x}_1) u_t(\mathbf{x}_t|\mathbf{x}_1)p_1(\mathbf{x}_1)\bigg)d\mathbf{x}_1\tag{substitute conditional continuity}\\
    &=-\nabla \cdot \left(\int_{\mathbf{x}_1}p_t(\mathbf{x}_t|\mathbf{x}_1) u_t(\mathbf{x}_t|\mathbf{x}_1)p_1(\mathbf{x}_1)d\mathbf{x}_1\right)\nonumber\\
    &=-\nabla \cdot \left(p_t(\mathbf{x}_t) u_t(\mathbf{x}_t)\right)
\end{align}
\end{small}
which concludes the proof.

\subsection{Proof of Flow Matching Propositions}
\textbf{Proposition 1} (Probability Mass Conservation) \label{appendix:Probability Mass Conservation} The conditional velocity field preserves probability mass and lies on the tangent bundle at point $\mathbf{x}_t$ on the simplex $\mathcal{T}_{\mathbf{x}_t}\Delta^{V-1}=\{u_t\in \mathbb{R}^V|\langle\mathbf{1}, u_t\rangle=0\}$.

\textit{Proof of Proposition 1.} We show that the conditional velocity field derived from the Gumbel-Softmax interpolant preserves probability mass such that
\begin{align}
    \sum_{i=1}^Vu_{t, i}(\mathbf{x}_t|\mathbf{x}_1=\mathbf{e}_k)=0
\end{align}
Summing up the velocities for all $i\in [1\dots V]$, we have 
\begin{small}
    \begin{align}
        \sum_{i=1}^Vu_{t}(\mathbf{x}_0|\mathbf{x}_1=\mathbf{e}_k)
        &=\sum_{i=1}^V\left[\frac{ \lambda}{\tau(t)} x_{t, i}\sum_{j=1}^Vx_{t, j}\cdot \bigg((\log \pi_i+g_i)-(\log \pi_j+g_j)\bigg)\right]\nonumber\\
        &=\frac{ \lambda}{\tau(t)}\sum_{i=1}^V\left[ x_{t, i}\left[\sum_{j=1}^Vx_{t, j}(\log \pi_i+g_i)-\sum_{j=1}^Vx_{t, j}(\log \pi_j+g_j)\right]\right]\nonumber\\
        &=\frac{ \lambda}{\tau(t)}\sum_{i=1}^V\left[ x_{t, i}\left[(\log \pi_i+g_i)\sum_{j=1}^Vx_{t, j}-\sum_{j=1}^Vx_{t, j}(\log \pi_j+g_j)\right]\right]\nonumber\\
        &=\frac{ \lambda}{\tau(t)}\left[ \sum_{i=1}^Vx_{t, i}(\log \pi_i+g_i)-\sum_{i=1}^Vx_{t, i}\sum_{j=1}^Vx_{t, j}(\log \pi_j+g_j)\right]\nonumber\\
        &=\frac{ \lambda}{\tau(t)}\left[ \sum_{i=1}^Vx_{t, i}(\log \pi_i+g_i)-\sum_{j=1}^Vx_{t, j}(\log \pi_j+g_j)\right]\nonumber\\
        &= 0
    \end{align}
\end{small}
which proves that our velocity field always preserves the probability mass $t$. 

\textbf{Proposition 3.} (Valid Flow Matching Loss)\label{appendix:Valid Flow Matching Loss}
\textit{If $p_t(\mathbf{x}_t)>0$ for all $\mathbf{x}_t\in \mathbb{R}^d$ and $t\in [0, 1]$, then the gradients of the flow matching loss and the Gumbel-Softmax FM loss are equal up to a constant not dependent on $\theta$ such that $\nabla_{\theta}\mathcal{L}_{\text{FM}}=\nabla_{\theta}\mathcal{L}_{\text{gumbel}}$ }

\textit{Proof of Proposition 3.} We can rewrite the conditional velocity field derived in Appendix \ref{appendix:Deriving the Conditional Velocity Field} as
\begin{align}
    u_{t}(\mathbf{x}_t|\mathbf{x}_1=\mathbf{e}_k)&=\frac{\lambda}{\tau(t)}x_{t, k}\left(\mathbf{e}_k-\mathbf{x}_t\right)\nonumber\\
    &=\frac{\lambda}{\tau(t)}\sum_{i=1}^Vx_{t, i}\left(\mathbf{e}_i-\mathbf{x}_t\right)\langle\mathbf{e}_i, \mathbf{x}_1\rangle
\end{align}
Furthermore, the predicted velocity field is given by
\begin{align}
    u^{\theta}_{t}(\mathbf{x}_t)&=\sum_{i=1}^Vu_{t}(\mathbf{x}_t|\mathbf{x}_1=\mathbf{e}_i)\left\langle \mathbf{e}_i, \mathbf{x}_{\theta}\right\rangle\nonumber\\
    &=\frac{\lambda}{\tau(t)}\sum_{i=1}^Vx_{t, i}(\mathbf{e}_i-\mathbf{x}_t)\left\langle \mathbf{e}_i, \mathbf{x}_{\theta}\right\rangle
\end{align}

Substituting the velocity field expressions into the flow-matching loss, we obtain
\begin{small}
\begin{align}
    &\mathbb{E}_{p_t(\mathbf{x}_t)}\|u_t(\mathbf{x}_t|\mathbf{x}_1) -u_t^{\theta}(\mathbf{x}_t)\|^2\nonumber\\
    &=\mathbb{E}_{p_t(\mathbf{x}_t)}\left\|\frac{\lambda}{\tau(t)}\sum_{i=1}^Vx_{t, i}\left(\mathbf{e}_i-\mathbf{x}_t\right)\langle\mathbf{e}_i, \mathbf{x}_1\rangle-\frac{\lambda}{\tau(t)}\sum_{i=1}^Vx_{t, i}(\mathbf{e}_i-\mathbf{x}_t)\left\langle \mathbf{e}_i, \mathbf{x}_{\theta}\right\rangle \right\|^2\nonumber\\
    &=\frac{\lambda^2}{\tau(t)^2}\mathbb{E}_{p_t(\mathbf{x}_t)}\left\|\sum_{i=1}^V\bigg[x_{t, i}\left(\mathbf{e}_i-\mathbf{x}_t\right)\langle\mathbf{e}_i, \mathbf{x}_1\rangle-x_{t, i}(\mathbf{e}_i-\mathbf{x}_t)\left\langle \mathbf{e}_i, \mathbf{x}_{\theta}\right\rangle \bigg] \right\|^2\nonumber\\
    &=\frac{\lambda^2}{\tau(t)^2}\mathbb{E}_{p_t(\mathbf{x}_t)}\left\|\sum_{i=1}^Vx_{t, i}\left(\mathbf{e}_i-\mathbf{x}_t\right)\bigg[\langle\mathbf{e}_i, \mathbf{x}_1\rangle-\langle \mathbf{e}_i, \mathbf{x}_{\theta}\rangle\bigg]  \right\|^2\nonumber\\
    &=\frac{\lambda^2}{\tau(t)^2}\mathbb{E}_{p_t(\mathbf{x}_t)}\left\|\sum_{i=1}^Vx_{t, i}\left(\mathbf{e}_i-\mathbf{x}_t\right)\langle\mathbf{e}_i, \mathbf{x}_1-\mathbf{x}_{\theta}\rangle \right\|^2\nonumber\\
    &=\frac{\lambda^2}{\tau(t)^2}\mathbb{E}_{p_t(\mathbf{x}_t)}\left\|\sum_{i=1}^Vx_{t, i}\left(\mathbf{e}_i-\mathbf{x}_t\right)(\mathbf{x}_1-\mathbf{x}_{\theta})_i \right\|^2\nonumber\\
    &=\frac{\lambda^2}{\tau(t)^2}\mathbb{E}_{p_t(\mathbf{x}_t)}\left\|\sum_{i=1}^Vx_{t, i}(\mathbf{x}_1-\mathbf{x}_{\theta})_i\mathbf{e}_i-\sum_{i=1}^Vx_{t, i}(\mathbf{x}_1-\mathbf{x}_{\theta})_i\mathbf{x}_t\right\|^2\nonumber\\
    &=\frac{\lambda^2}{\tau(t)^2}\mathbb{E}_{p_t(\mathbf{x}_t)}\left\|\mathbf{x}_t\odot (\mathbf{x}_1-\mathbf{x}_{\theta})-\mathbf{x}_t \langle \mathbf{x}_{t}, \mathbf{x}_1-\mathbf{x}_{\theta}\rangle \right\|^2
\end{align}
\end{small}

The remainder of the proof extends that of \citep{Lipman2022, Optimal-Transport}, which proved that the conditional flow matching loss $\nabla_{\theta}\mathcal{L}_{\text{CFM}}=\nabla_{\theta}\mathcal{L}_{\text{FM}}$ under similar constraints. 

First, we further expand the conditional flow-matching loss as follows
\begin{small}
\begin{align}
    &\mathbb{E}_{p_t(\mathbf{x}_t)}\|u_t(\mathbf{x}_t|\mathbf{x}_1)-u_t^{\theta}(\mathbf{x}_t, t)\|^2\nonumber\\
    &=\frac{\lambda^2}{\tau(t)^2}\mathbb{E}_{p_t(\mathbf{x}_t)}\left\|\mathbf{x}_t\odot (\mathbf{x}_1-\mathbf{x}_{\theta})-\mathbf{x}_t \langle \mathbf{x}_{t}, \mathbf{x}_1-\mathbf{x}_{\theta}\rangle \right\|^2\nonumber\\
    &=\frac{\lambda^2}{\tau(t)^2}\mathbb{E}_{p_t(\mathbf{x}_t)}\left\|\mathbf{x}_t\odot \mathbf{x}_1-\mathbf{x}_t\odot \mathbf{x}_{\theta}-\mathbf{x}_t \langle \mathbf{x}_{t}, \mathbf{x}_1-\mathbf{x}_{\theta}\rangle \right\|^2\nonumber\\
    &=\frac{\lambda^2}{\tau(t)^2}\mathbb{E}_{p_t(\mathbf{x}_t)}\left\|\mathbf{x}_t\odot \mathbf{x}_1-\mathbf{x}_t\odot (\mathbf{x}_{\theta}- \langle \mathbf{x}_{t}, \mathbf{x}_1-\mathbf{x}_{\theta}\rangle) \right\|^2\nonumber\\
    &=\frac{\lambda^2}{\tau(t)^2}\mathbb{E}_{p_t(\mathbf{x}_t)}\bigg[\|\mathbf{x}_t\odot \mathbf{x}_1\|^2-2\bigg\langle\mathbf{x}_t\odot \mathbf{x}_1,  \mathbf{x}_t\odot (\mathbf{x}_{\theta}- \langle \mathbf{x}_{t}, \mathbf{x}_1-\mathbf{x}_{\theta}\rangle)\bigg\rangle +\|\mathbf{x}_t\odot (\mathbf{x}_{\theta}- \langle \mathbf{x}_{t}, \mathbf{x}_1-\mathbf{x}_{\theta}\rangle)\|^2\bigg]\nonumber
\end{align}
\end{small}
Then, taking the gradient with respect to $\theta$, we have
\begin{align}
    &\nabla_{\theta}\mathbb{E}_{p_t(\mathbf{x}_t)}\|u_t(\mathbf{x}_t|\mathbf{x}_1)-u_t^{\theta}(\mathbf{x}_t, t)\|^2\nonumber\\
    &=\frac{\lambda^2}{\tau(t)^2}\nabla_{\theta}\mathbb{E}_{p_t(\mathbf{x}_t)}\bigg[\|\mathbf{x}_t\odot \mathbf{x}_1\|^2-2\bigg\langle\mathbf{x}_t\odot \mathbf{x}_1,  \mathbf{x}_t\odot (\mathbf{x}_{\theta}- \langle \mathbf{x}_{t}, \mathbf{x}_1-\mathbf{x}_{\theta}\rangle)\bigg\rangle +\|\mathbf{x}_t\odot (\mathbf{x}_{\theta}- \langle \mathbf{x}_{t}, \mathbf{x}_1-\mathbf{x}_{\theta}\rangle)\|^2\bigg]\nonumber\\
    &=\frac{\lambda^2}{\tau(t)^2}\bigg[-2\nabla_{\theta}\mathbb{E}_{p_t(\mathbf{x}_t)}\bigg\langle\mathbf{x}_t\odot \mathbf{x}_1,  \mathbf{x}_t\odot (\mathbf{x}_{\theta}- \langle \mathbf{x}_{t}, \mathbf{x}_1-\mathbf{x}_{\theta}\rangle)\bigg\rangle +\nabla_{\theta}\mathbb{E}_{p_t(\mathbf{x}_t)}\bigg\|\mathbf{x}_t\odot (\mathbf{x}_{\theta}- \langle \mathbf{x}_{t}\mathbf{x}_1, -\mathbf{x}_{t}\mathbf{x}_{\theta}\rangle)\bigg\|^2\bigg]
\end{align}
Now, we rewrite $\mathbf{x}_1$ as the expectation over noisy samples $\mathbf{x}_t$ learned by the model. By Bayes' theorem, we have
\begin{align}
    p(\mathbf{x}_1|\mathbf{x}_1)=\frac{p_t(\mathbf{x}_t|\mathbf{x}_1)p_1(\mathbf{x}_1)}{p_t(\mathbf{x}_t)}
\end{align}
Then, defining $\mathbf{x}_1$ as an expectation over $p_t(\mathbf{x}_t)$, we get
\begin{align}
    \mathbf{x}_1&=\mathbb{E}_{p(\mathbf{x}_1|\mathbf{x}_t)}\left[\mathbf{x}_1\right]\nonumber\\
    &=\int_{\mathbf{x}_1}\mathbf{x}_1p(\mathbf{x}_1|\mathbf{x}_t)d \mathbf{x}_1\nonumber\\
    &=\int_{\mathbf{x}_1}\mathbf{x}_1 \frac{p_t(\mathbf{x}_t|\mathbf{x}_1)p_1(\mathbf{x}_1)}{p_t(\mathbf{x}_t)}d \mathbf{x}_1
\end{align}
Now, we substitute this into the first expectation in the gradient to get
\begin{small}
\begin{align}
    &\mathbb{E}_{p_t(\mathbf{x}_t)}\bigg\langle\mathbf{x}_t\odot \mathbf{x}_1,  \mathbf{x}_t\odot \left(\mathbf{x}_{\theta}- \left\langle \mathbf{x}_{t}\mathbf{x}_1, -\mathbf{x}_{t}\mathbf{x}_{\theta}\right\rangle\right)\bigg\rangle\nonumber\\
    &=\int_{\mathbf{x}_t}\bigg\langle\mathbf{x}_t\odot \int_{\mathbf{x}_1}\mathbf{x}_1 \frac{p_t(\mathbf{x}_t|\mathbf{x}_1)p_1(\mathbf{x}_1)}{p_t(\mathbf{x}_t)}d \mathbf{x}_1,  \mathbf{x}_t\odot \left(\mathbf{x}_{\theta}- \left\langle \mathbf{x}_{t}\int_{\mathbf{x}_1}\mathbf{x}_1 \frac{p_t(\mathbf{x}_t|\mathbf{x}_1)p_1(\mathbf{x}_1)}{p_t(\mathbf{x}_t)}d \mathbf{x}_1, -\mathbf{x}_{t}\mathbf{x}_{\theta}\right\rangle\right)\bigg\rangle p_t(\mathbf{x}_t)d\mathbf{x}_t\nonumber\\
    &=\int_{\mathbf{x}_t}\bigg\langle\mathbf{x}_t\odot \int_{\mathbf{x}_1}\mathbf{x}_1 p_t(\mathbf{x}_t|\mathbf{x}_1)p_1(\mathbf{x}_1)d \mathbf{x}_1,  \mathbf{x}_t\odot \left(\mathbf{x}_{\theta}- \left\langle \mathbf{x}_{t}\int_{\mathbf{x}_1}\mathbf{x}_1 p_t(\mathbf{x}_t|\mathbf{x}_1)p_1(\mathbf{x}_1)d \mathbf{x}_1, -\mathbf{x}_{t}\mathbf{x}_{\theta}\right\rangle\right)\bigg\rangle d\mathbf{x}_t\nonumber\\
    &=\int_{\mathbf{x}_t}\bigg\langle\mathbf{x}_t\odot \int_{\mathbf{x}_1}\mathbf{x}_1 p_t(\mathbf{x}_t|\mathbf{x}_1)p_1(\mathbf{x}_1)d \mathbf{x}_1,  \mathbf{x}_t\odot \left(\mathbf{x}_{\theta}- \int_{\mathbf{x}_1}\left\langle \mathbf{x}_{t}\mathbf{x}_1, -\mathbf{x}_{t}\mathbf{x}_{\theta}\right\rangle p_t(\mathbf{x}_t|\mathbf{x}_1)p_1(\mathbf{x}_1)d \mathbf{x}_1\right)\bigg\rangle d\mathbf{x}_t\nonumber\\
    &=\int_{\mathbf{x}_t}\int_{\mathbf{x}_1}\bigg\langle\mathbf{x}_t\odot \mathbf{x}_1 ,  \mathbf{x}_t\odot \left(\mathbf{x}_{\theta}- \left\langle \mathbf{x}_{t}\mathbf{x}_1, -\mathbf{x}_{t}\mathbf{x}_{\theta}\right\rangle \right)\bigg\rangle p_t(\mathbf{x}_t|\mathbf{x}_1)p_1(\mathbf{x}_1)d \mathbf{x}_1 d\mathbf{x}_t\nonumber\\
    &=\int_{\mathbf{x}_1}\int_{\mathbf{x}_t}\bigg\langle\mathbf{x}_t\odot \mathbf{x}_1 ,  \mathbf{x}_t\odot \left(\mathbf{x}_{\theta}- \left\langle \mathbf{x}_{t}\mathbf{x}_1, -\mathbf{x}_{t}\mathbf{x}_{\theta}\right\rangle \right)\bigg\rangle p_t(\mathbf{x}_t|\mathbf{x}_1)p_1(\mathbf{x}_1)d \mathbf{x}_t d\mathbf{x}_1\nonumber\\
    &=\mathbb{E}_{p_t(\mathbf{x}_t|\mathbf{x}_1), p_1(\mathbf{x}_1) }\bigg\langle\mathbf{x}_t\odot \mathbf{x}_1,  \mathbf{x}_t\odot (\mathbf{x}_{\theta}- \langle \mathbf{x}_{t}, \mathbf{x}_1-\mathbf{x}_{\theta}\rangle)\bigg\rangle
\end{align}
\end{small}
where we use the linearity properties of integration.

Following similar logic, we have
\begin{small}
\begin{align}
    &\mathbb{E}_{p_t(\mathbf{x}_t)}\|\mathbf{x}_t\odot (\mathbf{x}_{\theta}- \langle \mathbf{x}_{t}\mathbf{x}_1, -\mathbf{x}_{t}\mathbf{x}_{\theta}\rangle)\|^2\nonumber\\
    &=\int_{\mathbf{x}_t} \|\mathbf{x}_t\odot (\mathbf{x}_{\theta}- \langle \mathbf{x}_{t}\mathbf{x}_1, -\mathbf{x}_{t}\mathbf{x}_{\theta}\rangle)\|^2p_t(\mathbf{x}_t)d\mathbf{x}_t\nonumber\\
    &=\int_{\mathbf{x}_t} \left\|\mathbf{x}_t\odot \left(\mathbf{x}_{\theta}- \left\langle \mathbf{x}_{t}\int_{\mathbf{x}_1}\mathbf{x}_1 \frac{p_t(\mathbf{x}_t|\mathbf{x}_1)p_1(\mathbf{x}_1)}{p_t(\mathbf{x}_t)}d \mathbf{x}_1, -\mathbf{x}_{t}\mathbf{x}_{\theta}\right\rangle\right)\right\|^2p_t(\mathbf{x}_t)d\mathbf{x}_t\nonumber\\
    &=\int_{\mathbf{x}_t} \int_{\mathbf{x}_1}\left\|\mathbf{x}_t\odot \left(\mathbf{x}_{\theta}- \left\langle \mathbf{x}_{t}\mathbf{x}_1 , -\mathbf{x}_{t}\mathbf{x}_{\theta}\right\rangle\right)\right\|^2p_t(\mathbf{x}_t|\mathbf{x}_1)p_1(\mathbf{x}_1)d \mathbf{x}_1d\mathbf{x}_t\nonumber\\
    &=\int_{\mathbf{x}_1} \int_{\mathbf{x}_t}\left\|\mathbf{x}_t\odot \left(\mathbf{x}_{\theta}- \left\langle \mathbf{x}_{t}\mathbf{x}_1 , -\mathbf{x}_{t}\mathbf{x}_{\theta}\right\rangle\right)\right\|^2p_t(\mathbf{x}_t|\mathbf{x}_1)p_1(\mathbf{x}_1)d \mathbf{x}_td\mathbf{x}_1\nonumber\\
    &=\mathbb{E}_{p_t(\mathbf{x}_t|\mathbf{x}_1), p_1(\mathbf{x}_1) }\|\mathbf{x}_t\odot (\mathbf{x}_{\theta}- \langle \mathbf{x}_{t}\mathbf{x}_1, -\mathbf{x}_{t}\mathbf{x}_{\theta}\rangle)\|^2
\end{align}
\end{small}
using the fact that the squared norm can be expressed as a bilinear inner product. 

Substituting these terms back into the gradient of the flow-matching loss, we get
\begin{small}
\begin{align}
    &\nabla_{\theta}\mathbb{E}_{p_t(\mathbf{x}_t)}\|u_t(\mathbf{x}_t|\mathbf{x}_1)-u_t^{\theta}(\mathbf{x}_t, t)\|^2\nonumber\\
    &=\frac{\lambda^2}{\tau(t)^2}\bigg[-2\nabla_{\theta}\mathbb{E}_{p_t(\mathbf{x}_t)}\bigg\langle\mathbf{x}_t\odot \mathbf{x}_1,  \mathbf{x}_t\odot (\mathbf{x}_{\theta}- \langle \mathbf{x}_{t}, \mathbf{x}_1-\mathbf{x}_{\theta}\rangle)\bigg\rangle +\nabla_{\theta}\mathbb{E}_{p_t(\mathbf{x}_t)}\bigg\|\mathbf{x}_t\odot (\mathbf{x}_{\theta}- \langle \mathbf{x}_{t}\mathbf{x}_1, -\mathbf{x}_{t}\mathbf{x}_{\theta}\rangle)\bigg\|^2\bigg]\nonumber\\
    &=\frac{\lambda^2}{\tau(t)^2}\bigg[-2\nabla_{\theta}\mathbb{E}_{p_t(\mathbf{x}_t|\mathbf{x}_1), p_1(\mathbf{x}_1) }\bigg\langle\mathbf{x}_t\odot \mathbf{x}_1,  \mathbf{x}_t\odot (\mathbf{x}_{\theta}- \langle \mathbf{x}_{t}, \mathbf{x}_1-\mathbf{x}_{\theta}\rangle)\bigg\rangle +\nabla_{\theta}\mathbb{E}_{p_t(\mathbf{x}_t|\mathbf{x}_1), p_1(\mathbf{x}_1) }\bigg\|\mathbf{x}_t\odot (\mathbf{x}_{\theta}- \langle \mathbf{x}_{t}\mathbf{x}_1, -\mathbf{x}_{t}\mathbf{x}_{\theta}\rangle)\bigg\|^2\bigg]\nonumber\\
    &=\nabla_{\theta}\mathbb{E}_{p_t(\mathbf{x}_t)}\left[\frac{\lambda^2}{\tau(t)^2}\left\|\mathbf{x}_t\odot (\mathbf{x}_1-\mathbf{x}_{\theta})-\mathbf{x}_t \langle \mathbf{x}_{t}, \mathbf{x}_1-\mathbf{x}_{\theta}\rangle \right\|^2\right]\nonumber\\
    &=\frac{\lambda^2}{\tau(t)^2}\nabla_{\theta}\mathbb{E}_{p_t(\mathbf{x}_t)}\left\|\mathbf{x}_t\odot (\mathbf{x}_1-\mathbf{x}_{\theta})-\mathbf{x}_t \langle \mathbf{x}_{t}, \mathbf{x}_1-\mathbf{x}_{\theta}\rangle \right\|^2
\end{align}
\end{small}
which concludes the proof that $\nabla_{\theta}\mathcal{L}_{\text{gumbel}}=\nabla_{\theta}\mathcal{L}_{\text{FM}}$. 

\section{Score Matching Derivations}
\subsection{Derivation of the Score Function} \label{appendix:Conditional Score Function}
We start by showing that the score function of the marginal probability density $\nabla_{\mathbf{x}_t}\log p_t(\mathbf{x}_t)$ is proportional to the conditional probability density $\nabla_{\mathbf{x}_t}\log p_t(\mathbf{x}_t|\mathbf{x}_1)$ given that $p_t(\mathbf{x}_t)=\mathbb{E}_{\mathbf{x}_1\sim p_1(\mathbf{x}_1)}\big[p_t(\mathbf{x}_t|\mathbf{x}_1)\big]$. 

Taking the gradient of the marginal log probability density and substituting in the definition of $p_t(\mathbf{x}_t)$, we have
\begin{align}
    \nabla_{\mathbf{x}_t}\log p_t(\mathbf{x}_t)&=\frac{\nabla_{\mathbf{x}_t}p_t(\mathbf{x}_t)}{p_t(\mathbf{x}_t)}\nonumber\\
    &=\frac{\nabla_{\mathbf{x}_t}\mathbb{E}_{\mathbf{x}_1\sim p_{\text{data}}}\big[p_t(\mathbf{x}_t|\mathbf{x}_1)\big]}{p_t(\mathbf{x}_t)}\nonumber\\
    &=\frac{\nabla_{\mathbf{x}_t}\int_{\mathbf{x}_1}\big[p(\mathbf{x}_1)p_t(\mathbf{x}_t|\mathbf{x}_1)\big]d\mathbf{x}_1}{p_t(\mathbf{x}_t)}\nonumber\\
    &= \frac{\int_{\mathbf{x}_1}p(\mathbf{x}_1)\nabla_{\mathbf{x}_t}p_t(\mathbf{x}_t|\mathbf{x}_1)d\mathbf{x}_1}{p_t(\mathbf{x}_t)}\nonumber\\
    &= \frac{\int_{\mathbf{x}_1}p(\mathbf{x}_1)p_t(\mathbf{x}_t|\mathbf{x}_1)\frac{\nabla_{\mathbf{x}_t}p_t(\mathbf{x}_t|\mathbf{x}_1)}{p_t(\mathbf{x}_t|\mathbf{x}_1)}d\mathbf{x}_1}{p_t(\mathbf{x}_t)}\nonumber\\
    &= \int_{\mathbf{x}_1}\underbrace{\frac{p_t(\mathbf{x}_t|\mathbf{x}_1)p(\mathbf{x}_1)}{p_t(\mathbf{x}_t)}}_{=p_t(\mathbf{x}_1|\mathbf{x}_t)}\nabla_{\mathbf{x}_t}\log p_t(\mathbf{x}_t|\mathbf{x}_1)d\mathbf{x}_1\nonumber\\
    &= \mathbb{E}_{\mathbf{x}_1\sim p_t(\mathbf{x}_1|\mathbf{x}_t)}\left[\nabla_{\mathbf{x}_t}\log p_t(\mathbf{x}_t|\mathbf{x}_1)\right]
\end{align}

which proves that with the perfect model such that $p_t(\mathbf{x}_1)=p(\mathbf{x}_1|\mathbf{x}_t)$, the gradient of the marginal log-probability density is exactly the expectation of the conditional log-probability density over the training data $\nabla_{\mathbf{x}_t}\log p_t(\mathbf{x}_t)=\mathbb{E}_{\mathbf{x}_1\sim p_1(\mathbf{x}_1)}\left[\nabla_{\mathbf{x}_t}\log p_t(\mathbf{x}_t|\mathbf{x}_1)\right]$.

\textbf{Theorem 2.} The gradient of the log-probability density of the \textsc{ExpConcrete} distribution is given by
\begin{align}
    \nabla_{x_{t, i}}\log p_t(\mathbf{x}_t|\mathbf{x}_1)&=\tau(t) -\tau(t) V\cdot \text{SM}\bigg(\delta_{ik}-\tau(t) x_{t, i}\bigg)
\end{align}

\textit{Proof of Theorem 2.} First, we start by defining the probability density of the \textsc{ExpConcrete} distribution. From \citep{Maddison2016}, integrating out the Gumbel-noise random variable we have 
\begin{align}
    p_t(\mathbf{x})=(V-1)!\tau^{V-1}\left(\sum_{i=1}^V\pi_j\exp(-\tau x_{t, j})\right)\left(\prod_{i=1}^V\pi_i\exp(-\tau x_{t, i})\right)
\end{align}
where $x_{t, i}$ is defined as a logit from the \textsc{ExpConcrete} distribution
\begin{align}
    x_{t, i}=\frac{\log\pi_i+g_i}{\tau}-\log\sum_{j=1}^V\exp \left(\frac{\log\pi_j+g_j   }{\tau}\right)
\end{align}
Taking the logarithm of the probability path, we have
\begin{small}
\begin{align}
    \log p_t(\mathbf{x}_t |\mathbf{x}_1)&=\log[ (V-1)!]+(V-1)\log \tau+\log \left(\prod_{i=1}^V\pi_i\exp(-\tau x_{t, i})\right)-V\log {\sum_{j=1}^V\pi_j\exp(-\tau x_{t, j})}\nonumber\\
    &=\log[ (V-1)!]+(V-1)\log \tau+\sum_{i=1}^V\log \left(\pi_i\exp(-\tau x_{t, i})\right)-V\log {\sum_{j=1}^V\exp\left(\log\left(\pi_j\exp(-\tau x_{t, j})\right)\right)}\nonumber\\
    &=\log[ (V-1)!]+(V-1)\log \tau+\sum_{i=1}^V\left(\log\pi_i-\tau x_{t, i}\right)-V\log {\sum_{j=1}^V\exp\bigg(\log \pi_j-\tau x_{t, j}\bigg)}\nonumber\\
    &=\log[ (V-1)!]+(V-1)\log \tau+\sum_{i=1}^V\log\pi_i-\sum_{i=1}^V\tau x_{t, i}-V\log {\sum_{j=1}^V\exp\bigg(\log \pi_j-\tau x_{t, j}\bigg)}
\end{align}
\end{small}
Then differentiating with respect to the logit of a single token $x_{t, j}$, we get
\begin{small}
\begin{align}
    \nabla_{x_{t, j}}\log p_t(\mathbf{x}_t |\mathbf{x}_1)&=-\nabla_{x_{t, i}}\sum_{i=1}^V\tau x_{t, i}-\nabla_{x_{t, i}}V\log {\sum_{j=1}^V\exp\bigg(\log \pi_j-\tau x_{t, j}\bigg)}\nonumber\\
    &=-\tau-V\left(\frac{1}{\sum_{j=1}^V\exp(\log \pi_j-\tau x_{t, j})}\right)\exp(\log \pi_i-\tau x_{t, i})(-\tau)\nonumber\\
    &=-\tau+\tau V\left(\frac{\exp(\log \pi_i-\tau x_{t, i})}{\sum_{i=1}^V\exp(\log \pi_j-\tau x_j)}\right)\nonumber\\
    &=-\tau +\tau V\cdot \text{SM}\bigg(\log \pi_i-\tau x_{t, i}\bigg)
\end{align}
\end{small}
Introducing time-dependence with $\tau(t)=\tau_{\text{max}}\exp(-\lambda t ) $ and target token dependence with $\pi_i=\exp (\delta_{ik})$, we have
\begin{align}
    \nabla_{x_{t, i}}\log p_t(\mathbf{x}_t|\mathbf{x}_1)&=\tau(t) -\tau(t) V\cdot \text{SM}\bigg(\delta_{ik}-\tau(t) x_{t, i}\bigg)
\end{align}

\subsection{Proof of Score Matching Propositions}

\textbf{Proposition 4.}\label{appendix:Concrete and ExpConcrete Score Function} The gradient of the \textsc{ExpConcrete} log-probability density is proportional to the gradient of the Gumbel-softmax log-probability density such that $\nabla_{x_{t, j}}^{\text{GS}}\log p_{\theta}(\mathbf{x}_t|\mathbf{x}_1)\propto \nabla_{x_{t, j}}^{\text{ExpConcrete}}\log p_{\theta}(\mathbf{x}_t|\mathbf{x}_1)$.

\textit{Proof of Proposition 4.} As derived in \citep{Maddison2016}, the explicit probability density of the Gumbel-Softmax distribution is defined as
\begin{align}
    p(\mathbf{x})=(V-1)!\tau^{V-1}\left(\sum_{i=1}^V\frac{\pi_i}{x_{t, i}^{\tau}}\right)^{-V}\prod_{i=1}^V\left(\frac{\pi_i}{x_{t, i}^{\tau+1}}\right)
\end{align}
We now derive the log-probability density of the Gumbel-Softmax distribution as
\begin{small}
\begin{align}
    \log p(\mathbf{x})&=\log[(V-1)!]+(V-1)\log\tau-V\log\sum_{i=1}^V\frac{\pi_i}{x_{t, i}^{\tau}}+\sum_{i=1}^V\log \left(\frac{\pi_i}{x_{t, i}^{\tau+1}}\right)\nonumber\\
    &=\log[(V-1)!]+(V-1)\log\tau-V\log\sum_{i=1}^V\frac{\pi_i}{x_{t, i}^{\tau}}+\sum_{i=1}^V\log \left(\pi_i\right)-(\tau+1)\sum_{i=1}^V\log(x_{t, i})
\end{align}
\end{small}

Taking the gradient with respect to a single token $x_{t, j}$, we have 
\begin{small}
\begin{align}
    \nabla_{x_{t, j}}^{\text{GS}}\log p_t(\mathbf{x}_t |\mathbf{x}_1)&=\nabla_{x_{t, j}}\left(-V\log\sum_{i=1}^V\frac{\pi_i}{x_{t, i}^{\tau}}\right)-\nabla_{x_{t, j}}\left((\tau+1)\sum_{i=1}^V\log(x_{t, i})\right)\nonumber\\
    &=-V\left(\frac{1}{\sum_{i=1}^V\frac{\pi_i}{x_{t, i} ^{\tau}}}\right)\left(\frac{-\pi_j\tau}{x_{t, j}^{\tau+1}}\right)-\frac{\tau+1}{x_{t, j}}\nonumber\\
    &= \frac{\tau V}{x_{t, j}}\left(\frac{\pi_j x_{t, j}^{-\tau}}{\sum_{i=1}^V\pi_ix_{t, i} ^{-\tau}}\right)-\frac{\tau+1}{x_{t, j}}\nonumber\\
    &= \frac{\tau V}{x_{t, j}}\left(\frac{\exp(\log(\pi_j x_{t, j}^{-\tau}))}{\sum_{i=1}^V\exp(\log(\pi_ix_{t, i} ^{-\tau}))}\right)-\frac{\tau+1}{x_{t, j}}\nonumber\\
    &= \frac{\tau V}{x_{t, j}}\left(\frac{\exp(\log\pi_j-\tau x_{t, j})}{\sum_{i=1}^V\exp(\log\pi_i-\tau x_{t, i} )}\right)-\frac{\tau+1}{x_{t, j}}\nonumber\\
    &=  \frac{\tau V}{x_{t, j}}\text{SM}\big(\log\pi_i-\tau x_{t, i}\big) -\frac{\tau +1}{x_{t, j}}\nonumber\\
    &= \frac{1}{x_{t, j}}\bigg(-\tau + \tau V\cdot \text{SM}\big(\log\pi_i-\tau x_{t, i}\big)\bigg)-\frac{1}{x_{t, j} }\nonumber\\
    &= \frac{1}{x_{t, j}}\bigg(\nabla_{x_{t, j}}^{\text{ExpConcrete}}\log p_t(\mathbf{x}_t |\mathbf{x}_1)\bigg)-\frac{1}{x_{t, j} }
\end{align}
\end{small}
Therefore, we show that the gradients of the Gumbel-Softmax and \textsc{ExpConcrete}  distributions are proportional to each other. Furthermore, we derive that the score of Gumbel-Softmax distribution further amplifies the scores for tokens with low probabilities by dividing by $x_{t, j}$ and subtracting $x_{t, j}^{-1}$. 

\section{Straight-Through Guided Flow Derivations}
\textbf{Proposition 5. } (Probability Mass Conservation of Straight-Through Gradient) 
\label{appendix:Probability Mass Conservation of Straight-Through Gradient} The straight through gradient $\nabla_{\mathbf{x}_t}p_{\phi}(y|\tilde{\mathbf{x}}_{1, m})$ preserves probability mass and lies on the tangent bundle at point $\mathbf{x}_t$ on the simplex $\mathcal{T}_{\mathbf{x}_t}\Delta^{V-1}=\{\nabla_{\mathbf{x}_t}p_{\phi}(y|\tilde{\mathbf{x}}_{1, m})\in \mathbb{R}^V|\langle\mathbf{1}, \nabla_{\mathbf{x}_t}p_{\phi}(y|\tilde{\mathbf{x}}_{1, m})\rangle=0\}$.

\textit{Proof of Proposition 5.} First, we recall our definition of the straight-through gradient of the classifier score $p_{\phi}(y|\tilde{\mathbf{x}}_{1, m})$ as
\begin{align}
    \nabla_{x_{t, i}}p_{\phi}(y|\tilde{\mathbf{x}}_{1, m})=\begin{cases}
        \frac{\partial p_{\phi}(y|\tilde{\mathbf{x}}_{1, m})}{\tilde{\mathbf{x}}_1}\cdot \big[\text{SM}(x_{t, i})\left(1-\text{SM}(x_{t, k})\right)\big]&i=k\\
        \frac{\partial p_{\phi}(y|\tilde{\mathbf{x}}_{1, m})}{\tilde{\mathbf{x}}_1}\cdot\big[-\text{SM}(x_{t, i})\text{SM}(x_{t, k})\big]&i\neq k
    \end{cases}\nonumber
\end{align}
Taking the sum over the simplex dimensions, we have
\begin{small}
\begin{align}
    \sum_{i=1}^V\nabla_{x_{t, i}}p_{\phi}(y|\tilde{\mathbf{x}}_{1, m})&=\frac{\partial p_{\phi}(y|\tilde{\mathbf{x}}_{1, m})}{\tilde{\mathbf{x}}_1}\left[\text{SM}(x_{t, k})\left(1-\text{SM}(x_{t, k})\right)-\sum_{i\neq k}\text{SM}(x_{t, i})\text{SM}(x_{t, k})\right]\nonumber\\
    &=\frac{\partial p_{\phi}(y|\tilde{\mathbf{x}}_{1, m})}{\tilde{\mathbf{x}}_1}\left[\text{SM}(x_{t, k})\left(1-\text{SM}(x_{t, k})\right)-\text{SM}(x_{t, k})\sum_{i\neq k}\text{SM}(x_{t, i})\right]\nonumber\\
    &=\frac{\partial p_{\phi}(y|\tilde{\mathbf{x}}_{1, m})}{\tilde{\mathbf{x}}_1}\bigg[\text{SM}(x_{t, k})\left(1-\text{SM}(x_{t, k})\right)-\text{SM}(x_{t, k})\left(1-\text{SM}(x_{t, k})\right)\bigg]\nonumber\\
    &=0\nonumber
\end{align}
\end{small}
which concludes the proof. In addition, it follows that the sum of straight-through gradients also preserves probability mass and lies on the tangent space of the simplex at any point. 

\begin{figure}[h!]
    \centering
    \includegraphics[width=\linewidth]{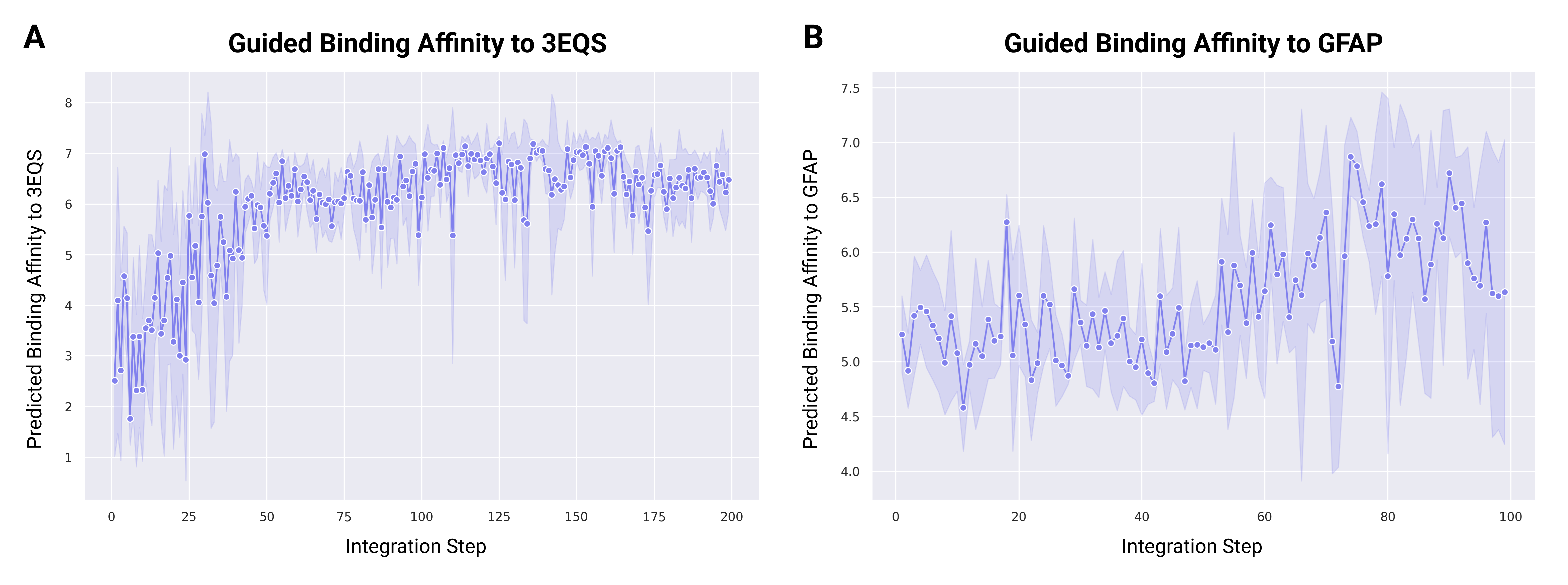}
    \caption{\textbf{Predicted binding-affinity scores over iteration of Gumbel-Softmax FM guided with STGFlow for target-binding peptide generation.} The predicted binding affinity is the mean regression scores of the $M$ discrete sequences sampled at each integration step. The gradients of the scores are used to compute the guided velocity. } 
    \label{fig:Guidance Curves}
\end{figure}

\section{Model Architecture}
\subsection{Diffusion Transformer}
To parameterize our flow and score matching models for the protein and peptide sequence generation tasks, we leverage the Diffusion Transformer (DiT) architecture \citep{DiT} which integrates time conditioning with adaptive layer norm (adaLN) and positional information with Rotary Positional Embeddings (RoPE) \citep{RoPE}. Our model consists of 32 DiT blocks, 16 attention heads, a hidden dimension of 1024, and dropout of 0.1. 

\begin{table}[H]
  \caption{Diffusion Transformer Architecture}
  \label{table:DiT Architecture}
  \centering
  \begin{tabular}{lcc}
    \toprule
     \textbf{Layers} & \textbf{Input Dimension}  & \textbf{Output Dimension}\\
    \midrule
    \textbf{Sequence Distribution Embedding Module}  & vocab size &  1024\\
    \hspace{5mm} Feed-Forward + GeLU & vocab size  & 1024\\
    \textbf{DiT Blocks} $\times 32$ & &\\
    \hspace{5mm} Adaptive Layer Norm (time conditioning) & 1024 & 1024\\
    \hspace{5mm} Multi-Head Self-Attention ($h=16$) \\\hspace{10mm}+ Rotary Positional Embeddings & 1024  & 1024\\
     \hspace{5mm} Dropout + Residual &  1024 & 1024\\
    \hspace{5mm} Adaptive Layer Norm (time conditioning) & 1024 & 1024\\
    \hspace{5mm} FFN + GeLU & 1024  & 1024\\

    \textbf{DiT Final Block} & &\\
    \hspace{5mm} Adaptive Layer Norm (time conditioning) & 1024 & 1024\\
    \hspace{5mm} Linear & 1024 & vocab size \\
    \bottomrule
  \end{tabular}
\end{table}

\subsection{Peptide-Binding Affinity Classifier} \label{appendix:Peptide-Binding Affinity Classifier} 
We trained a multi-head cross-attention network with ESM-2 650M \cite{Lin2023-gh} protein and peptide sequence embeddings to predict the binding affinity of a peptide to a protein sequence. We trained on 1781 sequences from the PepLand \cite{https://doi.org/10.48550/arxiv.2311.04419} protein-peptide binding dataset containing the protein-target sequence, peptide sequence, and the experimentally-validated $K_d/K_i/IC50$ binding affinity score, where higher values indicate stronger binding. 

In addition to the normalized binding affinity scores through regression, we also classified affinities into three categories: low ($< 6.0$), Medium ($6.0-7.5$), and Tight ($\geq 7.5$), with thresholds based on mean and Q3 quantile from the data distribution. The combined classification and regression approach helped the model better capture relationships between protein embeddings and binding affinities. Data was split in a 0.8/0.2 ratio with stratification preserving the score distribution.

We used OPTUNA \cite{akiba2019optuna} for hyperparameter optimization, tracking validation correlation, and F1 scores across 10 trials, resulting in an optimal learning rate of $3.84e-05$ and a dropout rate of 0.15. We retrain the whole classifier (Table \ref{table:affinity-predictor}) with the optimized set of parameters. After training for 50 epochs with early stopping based on validation Spearman correlation, the model achieved a Spearman correlation of 0.96 on training data and 0.64 on validation data, with F1 scores of 0.97 and 0.61 respectively. 
\begin{table}[H]
  \caption{Peptide-Binding Affinity Classifier}
  \label{table:affinity-predictor}
  \centering
  \begin{tabular}{lcc}
    \toprule
     \textbf{Layers} & \textbf{Protein Dimension} & \textbf{Peptide Dimension} \\
    \midrule
    Embedding Module  & $1280$ &  $1280$\\
    \textbf{CNN Layers} $\times 3$ (Kernel Sizes: 3,5,7) & $(1280, L)$ & $(64\times 3,  L)$ per kernel \\
    \quad ReLU Activation & $(64, L)$ per kernel & $(64, L)$ per kernel \\
    \textbf{Global Pooling} (Max + Avg) & $(64\times 3, L)$  & $64\times 3\times 2$ \\
    Linear Layer & $384$ & $384$ \\
    Layer Norm & $384$ & $384$ \\
    \textbf{Cross-Attention} $\times 4$ & &\\
    \hspace{5mm} Multi-Head Attention ($h=8$) & $384$  & $384$\\
     \hspace{5mm} Linear Layer &  $2048$& $2048$\\
    \hspace{5mm} ReLU & $2048$  & $2048$\\
    \hspace{5mm} Dropout & $2048$  & $2048$\\
    \hspace{5mm} Linear Layer &  $384$& $384$\\
    \textbf{Shared Prediction Head} & &\\
    \hspace{5mm} Linear Layer & \multicolumn{2}{c}{$1024$} \\
    \hspace{5mm} ReLU & \multicolumn{2}{c}{$1024$} \\
    \hspace{5mm} Dropout & \multicolumn{2}{c}{$1024$} \\
    \textbf{Regression Head} &  \multicolumn{2}{c}{$1$} \\
    \bottomrule
  \end{tabular}
\end{table}

\section{Experimental Details}
\subsection{Simplex-Dimension Toy Experiment}
We reproduce the experimental setup of the toy experiment in Davis et al. \cite{Davis2024}. We train $100,000$ sequences sampled from a randomly generated distribution over the $(K-1)$-dimensional simplex for $K=\{20, 40, 60, 80, 100, 120, 140, 160, 512\}$. We extend the experiment to dimension $512$ to evaluate performance in a higher simplex dimension. 

For the model architecture, we follow Stark et al. \cite{Stark2024} and parameterize all benchmark models with a 5-layer CNN with approximately 1M parameters that vary slightly with simplex dimension. After 50K steps, we evaluate the KL divergence $\text{KL}(\tilde{q} \|p_{\text{data}})$ where $\tilde{q}$ is the normalized distribution from 51.2K sequences generated by the model and $p_{\text{data}}$ is the distribution from which the training data was sampled. 

\begin{table*}[h!]
\caption{\textbf{KL divergences of toy experiment for increasing simplex dimensions compared to benchmark models.} The sequence length is set to a constant of 4 across all experiments. The toy models are trained on 100K sequences from a random distribution. KL divergence is evaluated for 51.2K sequences after 50K training steps.}
\label{table:Toy Experiment}
\begin{center}
\resizebox{\linewidth}{!}{
\begin{tabular}{@{}lcccccccccc@{}}
\toprule
\textbf{Simplex Dimension $K$} & 20 & 40 & 60 & 80 & 100 & 120 & 140 & 160 & 512 \\
\midrule
Linear FM & 0.013 & 0.046 & 0.070 & 0.100 & 0.114 & 0.112 & 0.156 & 0.146 & 0.479 \\
Dirichlet FM & 0.007 & 0.017 & 0.032 & 0.035 & 0.028 & 0.024 & 0.039 & 0.053 & 0.554 \\
Fisher FM (Optimal Transport) & 0.0004 & 0.007 & 0.007 & 0.007 & 0.008 & 0.043 & 0.013 & 0.013 & 0.036 \\
\textbf{Gumbel-Softmax FM (Ours)} & 0.029 & 0.027 & 0.025 & 0.027 & 0.030 & 0.029 & 0.035 & 0.038 & 0.048 \\
\bottomrule
\end{tabular}
}

\end{center}
\end{table*}

\begin{figure}
    \centering
    \includegraphics[width=\linewidth]{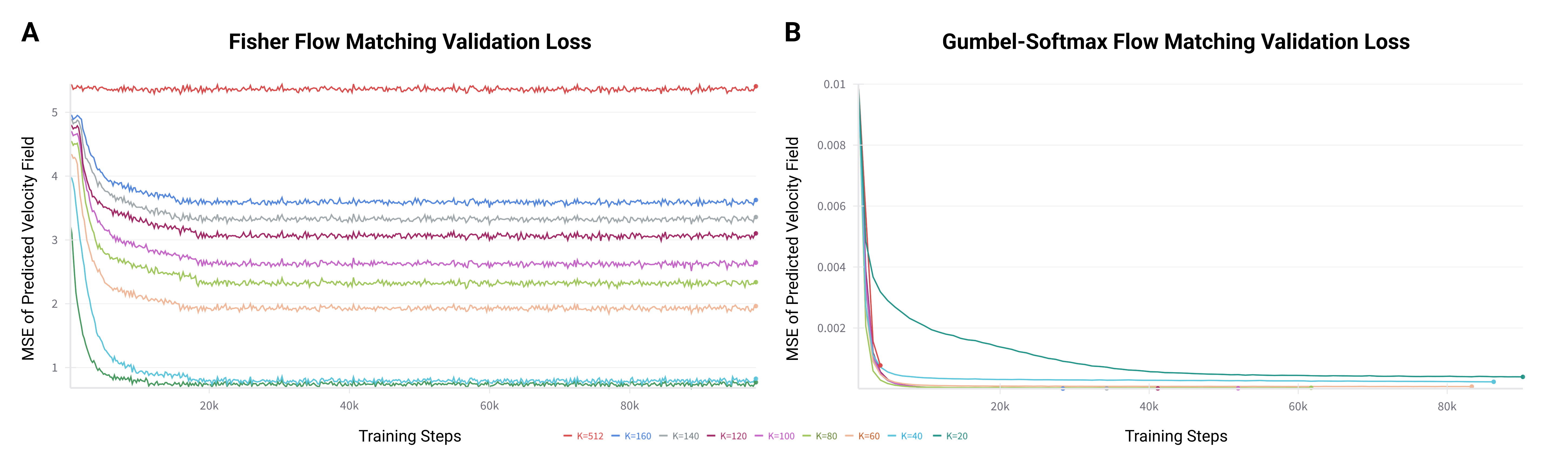}
    \caption{\textbf{Validation MSE loss over training step of simplex-dimension toy experiment.} Fisher FM exhibits significantly higher validation MSE loss during training than Gumbel-Softmax FM despite the same loss calculation, suggesting that the parameterization easily overfits to training data. }
    \label{fig:Validation Loss}
\end{figure}

\subsection{Hyperparameter Selection}
\textbf{Maximum Temperature $\tau_{\text{max}}$}. The maximum temperature controls the uniformity of the probability distribution at $t=0$ when $\exp(-\lambda t)=1$. Theoretically, the probability distribution is fully uniform $\psi_0(\mathbf{x}_t|\mathbf{x}_1)=\frac{\mathbf{1}}{V}$ when $\tau_{\text{max}}\to \infty$. Empirically, we find that setting $\tau_{\text{max}}=10.0$ ensures that the distribution is near uniform at $t=0$ even after applying Gumbel noise, satisfying the boundary condition $\psi_0(\mathbf{x}_t|\mathbf{x}_1)\approx \frac{\mathbf{1}}{V}$. 

\textbf{Decay Rate $\lambda$}. The decay rate determines how quickly the temperature drops as $t\to 1$. A decay rate of $\lambda = 1$ means that the function becomes $\exp(-t)$ which drops the temperature to $\approx 0.367$ at $t=1$. Since we want the temperature to approach 0 to increase the concentration of probability mass at the vertex, we set $\lambda = 3.0$ such that $\tau(t)=\tau_{\text{max}}\exp(-3.0t)$. For larger decay rates $\lambda =10.0$, the distribution converges too quickly to a vertex which may cause overfitting. 

\textbf{Stochasticity Factor $\beta$. }We can tune the effect of the Gumbel noise applied during training by scaling down by a factor $\beta \geq 1.0$ such that $g_i=\frac{-\log(-\log (\mathcal{U}_i +\epsilon ) +\epsilon ) }{\beta}$. For larger $\beta$, the stochasticity decreases and for smaller $\beta$, the stochasticity increases. For the toy experiment, we found similar performance for noise factors ranging between $\beta =2.0\to 10.0$. The remaining experiments were conducted with $\beta=2.0$. 

\textbf{Step Size $\eta$ and Integration Steps $N_{\text{steps}}$. } For Gumbel-Softmax FM, the step size is equal to $\Delta t =\frac{1}{N_{\text{steps}}}$ since we are integrating the velocity field from $t=0\to 1$. For Gumbel-Softmax SM, the step size determines the rate of convergence to high-probability density regions. Small step sizes $\eta \leq 0.1$ increase computation cost and number of steps needed to converge. In contrast, larger step sizes $0.1\leq \eta \leq 1.0$ increase the speed of convergence but may result in mode-collapse to the high-density regions. Empirically, we found that a step size of $\eta =0.5$ is optimal with the number of integration steps $N_{\text{steps}}=100$. 

\textbf{Guidance Scale $\gamma$.} Given that the \texttt{softmax} gradients tend to be small, especially for low-probability tokens, the guidance scale $\gamma$ amplifies the gradient value across all tokens to ensure effective guidance. For the target-guided peptide design experiments, we set $\gamma=10.0$ to scale the guidance term to be in the order $10^{-1}$ which produced increasing classifier scores over iterations. 

\textbf{Number of Guidance Samples $M$}. For STGFlow, the number of guidance samples $M$ determines the number of discrete sequences that are sampled from the distribution $\mathbf{x}_t$ at each time step to compute the aggregate straight-through gradient. Larger $M$ enables more informed and precise guidance based on the culmination of the classifier on various token combinations to determine tokens that lead to enhanced classifier scores, while smaller $M$ results in more spurious guidance that may not lead to truly optimal sequences. We found that $M=10$ maintained a good balance between effective guidance while minimizing computational costs. 

\subsection{Protein Evaluation Metrics}\label{appendix:Protein Evaluation Metrics}
We evaluate protein generation quality based on the following metrics computed by ESMFold \citep{Lin2023-gh}. 
\begin{enumerate}
    \item \textbf{pLDDT} (predicted Local Distance Difference Test) measures residue-wise local structural confidence on a scale of 0-100. Proteins with mean pLDDT $>70$ generally correspond to correct backbone prediction and more stable proteins.
    \item \textbf{pTM} (predicted Template Modeling) measures global structural plausibility. High pTM corresponds to a high similarity between a predicted structure and a hypothetical true structure. 
    \item \textbf{pAE} (predicted Alignment Error) measures the confidence in pair-wise positioning of residues. Low pAE scores correspond to low predicted pair-wise error. 
\end{enumerate}
In addition, we compute:
\begin{enumerate}
    %\item \textbf{Pseudo-perplexity} is calculated using ESM2 which computes the \textit{feasibility} of a protein sequence based on the predicted probability of each token based on the previous tokens via the equation:
    %\begin{align}
        %\text{PPL}(x)=\exp\left(-\frac{1}{L}\sum_{\ell=1}^L\log p(x_{\ell}|x_{<\ell})\right)\nonumber
    %\end{align}
    
    \item \textbf{Token entropy} measures the diversity of tokens within each sequence. It is defined as the Shannon entropy, where $p_{i}$ is the probability of $i$-th unique token divided by the total number of tokens $N$ in the sequence. 
    \begin{align}
        E = -\sum_{i=1}^{N}p_{i}\log_{2}(p_{i})\nonumber
    \end{align}
    \item \textbf{Diversity} is calculated as $1-$ pairwise sequence identity within a batch of generated sequences with equal length. 
\end{enumerate}

\subsection{Peptide Evaluation Metrics}\label{appendix:Peptide Evaluation Metrics}
We evaluate our \textit{de novo} peptide binders based on two metrics that measure their affinity to their target protein.

\textbf{ipTM Score.} We use AlphaFold3 \cite{Abramson2024} to compute the interface predicted template modeling (ipTM) score which is on the scale from 0-1 and measures the accuracy of the predicted relative positions between residues involved in the interaction between the two sequences. 

\textbf{pTM Score.} We use AlphaFold3 \cite{Abramson2024} to compute the predicted template modeling (pTM) score which is on a scale from 0-1 and measures the accuracy of the predicted structure of the whole peptide-protein complex. This score is less relevant when evaluating binding affinity since it can be dominated by the stability of the target protein. 

\textbf{VINA Docking Score.}\label{methods:docking} We use Autodock Vina \cite{eberhardt2021autodock} (v 1.1.2) for \textit{in silico} docking of the peptide binders to their target proteins (Table \ref{table:Peptide Existing Binder}) to confirm binding affinity. The complex was first docked with Alphafold3 for the starting conformation \cite{Abramson2024}. The final results were visualized in PyMol \cite{PyMOL} (v 3.1), where the residues in the protein targets with polar contacts to the peptide binder with distances closer than 3.5 \AA\ are annotated. 

\begin{figure*}[h!]
    \centering
    \includegraphics[width=\linewidth]{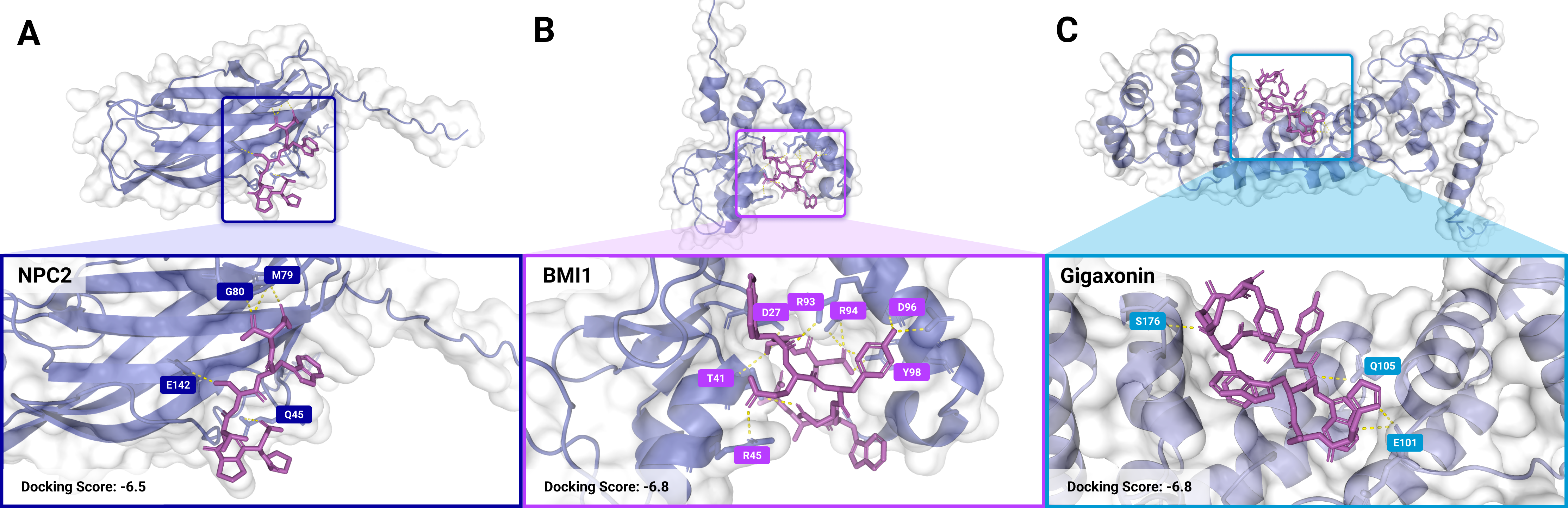}
    \caption{\textbf{Gumbel-Softmax FM generated peptide binders for three targets with no known binders.} (\textbf{A}) 7 a.a. designed binder to NPC2 (PDB: 6W5V) involved in Niemann-Pick Disease Type C. (\textbf{B}) 10 a.a. designed binder to BMI1 (PDB: 2CKL) involved in Medulloblastoma. (\textbf{C}) 10 a.a. designed binder to Gigaxonin (PDB: 3HVE) involved in Giant Axonal Neuropathy. Docked with AutoDock VINA and polar contacts within 3.5 \AA\ are annotated. Additional targets are shown in Table \ref{table:Peptide No Existing Binder}.}
    \label{fig:Peptides No Existing Binder}
\end{figure*}

\section{Algorithms}
In this section, we provide detailed procedures for the training and inference of the flow and score-matching models. Algorithm \ref{alg:Training Gumbel-Softmax FM} and \ref{alg:Sampling Gumbel-Softmax FM} describe training and sampling with Gumbel-Softmax FM, respectively. Algorithm \ref{alg:Training Gumbel-Softmax SM} and \ref{alg:Sampling Gumbel-Softmax SM} describe training and sampling with Gumbel-Softmax SM, respectively.  We consider $\mathbf{x}_1$ as a single token in a sequence for simplicity, but in practice, the training and sampling is conducted on a sequence of tokens of length $L$. 

\begin{algorithm}[h!]
\caption{Training Gumbel-Softmax Flow Matching}\label{alg:Training Gumbel-Softmax FM}
    \begin{algorithmic}
        \State\textbf{Inputs:} Training sequences of one-hot vectors $\mathbf{x}_1\in \mathcal{D}$, parameterized neural network $\text{NN}_{\theta}(\mathbf{x}_t, t)$, maximum temperature $\tau_{\text{max}}$, decay rate $\lambda$, and learning rate $\eta$.
        \Procedure{Training Gumbel-Softmax FM}{}
            \For {$\mathbf{x}_1$ in \texttt{batch}}
                \State Sample $t\sim \text{Uniform}(0,1)$
                \State Set $\tau(t)\gets \tau_{\text{max}}\exp (-\lambda t)$

                \State Sample $\mathcal{U}\sim \text{Uniform}(0, 1)^V$
                \State Sample Gumbel noise vector $\mathbf{g}=-\log(-\log(\mathcal{U}+\epsilon)+\epsilon)$
                \State Given the clean token $\mathbf{x}_1=\mathbf{e}_k$, sample noisy interpolant for time $t$
                    \[x_{t, i}\gets \frac{\exp\left(\frac{\delta_{ik}+(g_i/\beta)}{\tau(t)}\right)}{\sum _{j=1}^V\exp\left(\frac{\delta_{jk}+(g_j/\beta)}{\tau(t)}\right)}\]
                
                \If {\texttt{denoise}} 
                    \State Predict $\mathbf{x}_{\theta}(\mathbf{x}_t, t)\gets \text{NN}_{\theta}(\mathbf{x}_t, t)$
                    \State Minimize negative log loss $\mathcal{L}_{\text{denoise}}\gets \mathbb{E}_{\mathbf{x}_1\sim \mathcal{D}}\left[-\log(\mathbf{x}^{(k)}_{\theta}(\mathbf{x}_t, t))\right]$
                \Else
                    \State Predict $u_t^{\theta}(\mathbf{x}_t)\gets \text{NN}_{\theta}(\mathbf{x}_t, t)$
                    \State Calculate $u_t(\mathbf{x}_t|\mathbf{x}_1)\gets \frac{\lambda\exp(\lambda t)}{\tau_{\text{max}}}\left[(\mathbf{x}_t\odot \mathbf{x}_1) (1-x_{t, k})-(\mathbf{x}_t\odot (\mathbf{1}-\mathbf{x}_1))x_{t, k}\right]$
                    \State Optimize denoising loss $\mathcal{L}_{\text{mse}}\gets\mathbb{E}_{\mathbf{x}_1\sim \mathcal{D}}\|u_t^{\theta}(\mathbf{x}_t)-u_t(\mathbf{x}_t|\mathbf{x}_1)\|^2$
                \EndIf

                \State $\theta\gets \theta+\eta \nabla_{\theta} \mathcal{L}_{\text{denoise}}$
                %\State Predict error $\mathcal{E}_{\phi}(\mathbf{x}_t, t)\gets \text{MLP}_{\phi}(\mathbf{x}_t, t)$
                %\State Optimize error prediction loss $\mathcal{L}_{\text{error}}=\frac{1}{L}\sum_{\ell=1}^L\|\mathcal{E}_{\phi}(\mathbf{x}_t, t ) -\|\mathbf{x}_{\theta}(\mathbf{x}_t,t)-\mathbf{x}_1\|^2\|^2$
                %\State $\phi\gets \phi+\eta \nabla_{\phi} \mathcal{L}_{\text{error}}$
            \EndFor
        \EndProcedure
    \end{algorithmic}
\end{algorithm}

\begin{algorithm}[h!]
\caption{Unconditional Sampling with Gumbel-Softmax Flow Matching}\label{alg:Sampling Gumbel-Softmax FM}
    \begin{algorithmic}
        \State\textbf{Inputs:} Trained neural network $\text{NN}_{\theta}(\mathbf{x}_t, t)$, number of integration steps $N_{\text{step}}$
        \State \textbf{Output:} Clean sequence $\mathbf{x}$ from learned data distribution
        \Procedure{Sampling Gumbel-Softmax FM}{}
            \State Compute step size $\Delta t\gets \frac{1}{N_{\text{step}}}$
            \State Sample uniform distribution $\mathbf{x}_0 \gets\frac{\mathbf{1}}{V}$
            \State Set $\mathbf{x}_t\gets \mathbf{x}_0$
            \For {$t=0\to 1$}
                \State Compute $\tau (t)\gets \tau_{\text{max}}\exp(-\lambda t)$
                \If {\texttt{denoise}} 
                    \State Predict $\mathbf{x}_{\theta}(\mathbf{x}_t, t)\gets \text{NN}_{\theta}(\mathbf{x}_t, t)$
                    \For {all simplex dimensions $k\in [1, V]$} 
                        \begin{align}
                            u_{t}(\mathbf{x}_t|\mathbf{x}_1=\mathbf{e}_k)&=\frac{\lambda}{\tau(t)}x_{t, k}\left(\mathbf{e}_k-\mathbf{x}_t\right)\nonumber
                        \end{align}
                    \EndFor
                    \State Calculate conditional velocity field
                    \begin{align}
                        u^{\theta}_{t}(\mathbf{x}_t)\gets \sum_{k=1}^V u_{t}(\mathbf{x}|\mathbf{x}_1=\mathbf{e}_k)\cdot \langle \mathbf{x}_{\theta}(\mathbf{x}_t, t), \mathbf{e}_k\rangle \nonumber
                    \end{align}
                \Else
                    \State Directly predict conditional velocity field $u_t^{\theta}(\mathbf{x}_t)\gets \text{NN}_{\theta}(\mathbf{x}_t , t)$ 
                \EndIf
                \State Take step $\mathbf{x}_t\gets \mathbf{x}_t+\Delta t\cdot u^{\theta}_t(\mathbf{x}_t)$
                \State $\mathbf{x}_t\gets$ \textsc{SimplexProj}$(\mathbf{x}_t)$
            \EndFor
            \State Sample sequence $\mathbf{x}\gets \arg \max(\mathbf{x}_t) $
            \State \textbf{return} $\mathbf{x}$
        \EndProcedure
    \end{algorithmic}
\end{algorithm}

\begin{algorithm}[h!]
\caption{Training Gumbel-Softmax Score Matching}\label{alg:Training Gumbel-Softmax SM}
    \begin{algorithmic}
        \State\textbf{Inputs:} Training sequences of one-hot vectors $\mathbf{x}_1\in \mathcal{D}$, parameterized neural network $\text{NN}_{\theta}(\mathbf{x}_t, t)$, maximum temperature $\tau_{\text{max}}$, decay rate $\lambda$, and learning rate $\eta$.
        \Procedure{Training Gumbel-Softmax SM}{}
            \For {$\mathbf{x}_1$ in \texttt{batch}}
                \State Sample $t\sim \text{Uniform}(0,1)$
                \State Set $\tau(t)\gets \tau_{\text{max}}\exp (-\lambda t)$

                \State Sample $\mathcal{U}\sim \text{Uniform}(0, 1)^V$
                \State Sample Gumbel noise vector $\mathbf{g}=-\log(-\log(\mathcal{U}+\epsilon)+\epsilon)$
                \State Given the clean token $\mathbf{x}_1=\mathbf{e}_k$, sample noisy interpolant for time $t$
                    \[x_{t, i}\gets \frac{\exp\left(\frac{\delta_{ik}+(g_i/\beta)}{\tau(t)}\right)}{\sum _{j=1}^V\exp\left(\frac{\delta_{jk}+(g_j/\beta)}{\tau(t)}\right)}\]
                
                \State Predict $f_{\theta}(\mathbf{x}_t, t)\gets \text{NN}_{\theta}(\mathbf{x}_t, t)$
                \State Optimize loss given $\mathbf{x}_1=\mathbf{e}_k$ 
                \begin{align}
                    \mathcal{L}_{\text{score}}\gets \mathbb{E}_{\mathbf{x}_1\sim \mathcal{D}}\| f_{\theta}(\mathbf{x}_t, t)-(\delta_{ik} +\tau(t)x_{t, i})\|^2\nonumber
                \end{align}
                \State $\theta\gets \theta+\eta \nabla_{\theta} \mathcal{L}_{\text{score}}$
            \EndFor
        \EndProcedure
    \end{algorithmic}
\end{algorithm}

\begin{algorithm}[h!]
\caption{Unconditional Sampling with Gumbel-Softmax Score Matching}\label{alg:Sampling Gumbel-Softmax SM}
    \begin{algorithmic}
        \State\textbf{Inputs:} Trained score model $s_{\theta}(\mathbf{x}_t, t)$, step size $\Delta$, noise factor $\beta$
        \State \textbf{Output:} Clean sequence $\mathbf{x}$ from learned data distribution
        \Procedure{Sampling}{}
            \State $\mathbf{x}_0\gets\frac{\mathbf{1}}{V}$
            \State Set $\mathbf{x}_t\gets \mathbf{x}_0$
            \For {$t=0\to 1$}
                \State Compute $\tau(t)\gets \tau_{\text{max}}\exp (-\lambda t)$
                \State Predict $f_{\theta}(\mathbf{x}_t, t)\gets \text{NN}_{\theta}(\mathbf{x}_t, t)$
                \State Compute predicted score $s_{\theta}(\mathbf{x}_t, t)\gets -\tau (t)+\tau(t) V\cdot \text{SM}\big(f_{\theta}(\mathbf{x}_t, t)\big)$
                \State $\mathbf{x}_t\gets \mathbf{x}_t+\Delta \cdot  s_{\theta}(\mathbf{x}_t, t)$
                \State $\mathbf{x}_t\gets$ \textsc{SimplexProj}$(\mathbf{x}_t)$
            \EndFor
            \State Sample sequence $\mathbf{x}\gets \arg \max(\mathbf{x}_t) $
            \State \textbf{return} $\mathbf{x}$
        \EndProcedure
    \end{algorithmic}
\end{algorithm}

\begin{algorithm}[h!]
\caption{Straight-Through Guided Flow Matching (STGFlow)}\label{alg:Straight-Through Guided Flow Matching}
    \begin{algorithmic}
        \State\textbf{Inputs:} Trained simplex-based flow matching model $u^{\theta}_t(\mathbf{x}_t)$, trained classifier model $p_{\phi}(y|\mathbf{x}): \mathcal{V}^L\to \mathbb{R}$ that takes a sequence of length $L$ and returns a classifier score, number of integration steps $N_{\text{iter}}$
        \State \textbf{Output:} Clean sequence $\mathbf{x}$ from learned data distribution with optimized classifier score
        \Procedure{Guided Sampling with STGFlow}{}
            \State Compute step size $\Delta t\gets \frac{1}{N_{\text{step}}}$
            \State $\mathbf{x}_0\gets\frac{\mathbf{1}}{V}$
            \State Set $\mathbf{x}_t\gets \mathbf{x}_0$
            \For {$t=0\to 1$}
                \State Predict unguided conditional velocity field $u^{\theta}_{t}(\mathbf{x}_t)$ as in Algorithm \ref{alg:Sampling Gumbel-Softmax FM}
                \State Take step $\mathbf{x}_t\gets \mathbf{x}_t+\Delta t\cdot u_t^{\theta}(\mathbf{x}_t)$
                \State Compute top-$k$ distribution $\text{SM}\left(\text{top}k(\mathbf{x}_t)\right)$
                \State Sample $M$ sequences from top$k$ distribution $\tilde{\mathbf{x}}_{1, m}\sim \text{SM}\left(\text{top}k(\mathbf{x}_t)\right)$
                \State Initialize total guided velocity $u_t^{\phi}(\mathbf{x}_t|\mathbf{x}_1, y)\gets 0$
                \For {each $\tilde{\mathbf{x}}_{1, m}$}
                    \State Compute score $y\gets p_{\phi}(y|\tilde{\mathbf{x}}_{1, m})$
                    \State Compute straight-through gradient with respect to distribution $\mathbf{x}_t$
                    \begin{align}
                        \nabla_{\mathbf{x}_t}p_{\phi}(y|\tilde{\mathbf{x}}_{1, m})=\begin{cases}
                            \frac{\partial p_{\phi}(y|\tilde{\mathbf{x}}_{1, m})}{\tilde{\mathbf{x}}_1}\cdot \left[\text{SM}(x_{t, i})\left(1-\text{SM}(x_{t, k})\right)\right]&i=k\\
                            \frac{\partial p_{\phi}(y|\tilde{\mathbf{x}}_{1, m})}{\tilde{\mathbf{x}}_1}\cdot\left[-\text{SM}(x_{t, i})\text{SM}(x_{t, j})\right]&i\neq k
                        \end{cases}\nonumber
                    \end{align}
                    \State Add to total guidance $u_t^{\phi}(\mathbf{x}_t|\mathbf{x}_1, y)\gets u_t^{\phi}(\mathbf{x}_t|\mathbf{x}_1, y)+\nabla_{\mathbf{x}_t}p_{\phi}(y|\tilde{\mathbf{x}}_{1, m})$
                \EndFor
                \State Add total guided velocity $\mathbf{x}_t\gets \mathbf{x}_t+\gamma \cdot u_t^{\phi}(\mathbf{x}_t|\mathbf{x}_1, y)$
            \EndFor
            \State Sample sequence $\mathbf{x}\sim \mathbf{x}_t$
            \State \textbf{return} $\mathbf{x}$
        \EndProcedure
    \end{algorithmic}
\end{algorithm}

\begin{figure}[h!]
    \centering
    \includegraphics[width=\linewidth]{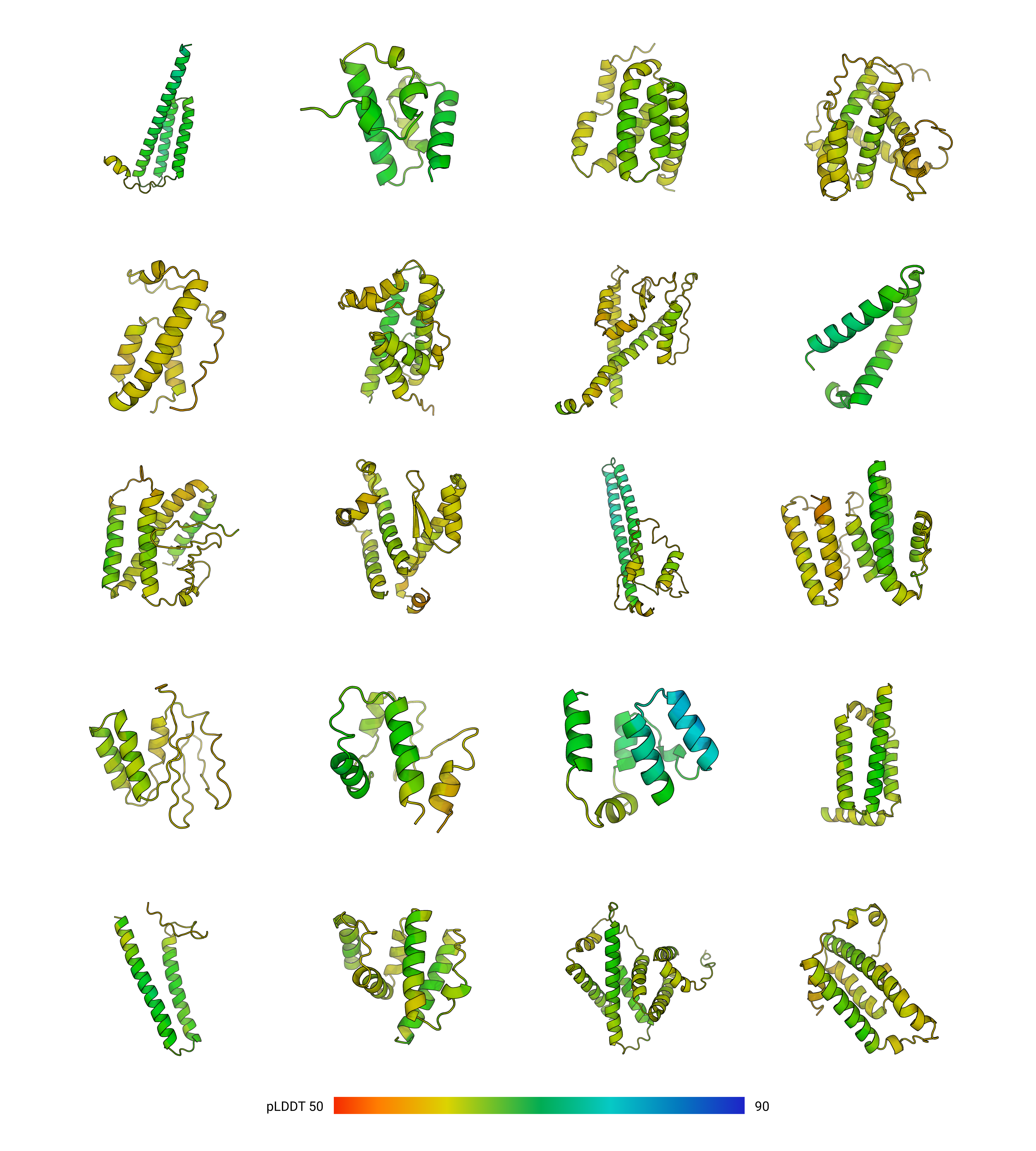}
    \caption{\textbf{Predicted structures of \textit{de novo} generated proteins with Gumbel-Softmax FM.} Generated proteins demonstrate diverse structural generation. }
    \label{fig:Appendix Proteins}
    
\end{figure}

\end{document}